\documentclass[preprint]{elsarticle}

\makeatletter
\def\ps@pprintTitle{%
 \let\@oddhead\@empty
 \let\@evenhead\@empty
 \def\@oddfoot{\centerline{\thepage}}%
 \let\@evenfoot\@oddfoot}
\makeatother

\usepackage[margin=2cm]{geometry}

\usepackage{hyperref}

\usepackage{graphicx}
\usepackage{amsmath,amssymb} %
\usepackage{color}
\usepackage{url}
\usepackage{scrextend}
\usepackage{hyperref}
\usepackage{cleveref}
\usepackage{diagbox} %

\usepackage{footnote}
\usepackage{multirow}
\usepackage{algorithm,algorithmic}
\usepackage{rotating}
\usepackage[section]{placeins} %

\graphicspath{{./im/}}
\newcommand\NB{IconArt} %
\newcommand\MIL{MI-max without score}
\newcommand\MILS{MI-max}

\newcommand\MAX{MAX}
\newcommand\MAXA{MAXA}

\newcommand\mimodels{mi-perceptron} 
 
\newcommand\mimaxaddlayerS{MI-max-HL} 
\newcommand\MaxOfMaxS{Polyhedral MI-max}
\newcommand\MaxOfMax{Polyhedral MI-max without score}

\newcommand{\heightimageCASPA}{4cm}

\newcommand{\heightimageCASPAzeroSept}{3cm}

\newcommand{\repeatfootnote}{\textsuperscript{\thefootnote}}

\definecolor{cyan}{rgb}{0.0, 1.0, 1.0}
\definecolor{darkpastelgreen}{rgb}{0.01, 0.75, 0.24}
\definecolor{electricyellow}{rgb}{1.0, 1.0, 0.0}
\definecolor{trueblue}{rgb}{0.0, 0.45, 0.81}
\definecolor{carrotorange}{rgb}{0.93, 0.57, 0.13}
\definecolor{darkpastelpurple}{rgb}{0.59, 0.44, 0.84}
\definecolor{fuchsiapink}{rgb}{1.0, 0.47, 1.0}
\definecolor{purpleheart}{rgb}{0.41, 0.21, 0.61}

\usepackage{newunicodechar}
\newunicodechar{ﬁ}{fi}
\newunicodechar{ﬀ}{ff}

\usepackage[colorinlistoftodos]{todonotes} %
\usepackage{breakcites}

\begin{document}

\setcounter{secnumdepth}{3} %

\begin{frontmatter}

\title{Multiple instance learning on deep features for weakly supervised object detection with extreme domain shifts}
\author[telecomadress,UPSaclay]{Nicolas Gonthier\corref{mycorrespondingauthor}}
\cortext[mycorrespondingauthor]{Corresponding author}
\ead{nicolas.gonthier@telecom-paris.fr}

\author[telecomadress]{Sa\"id Ladjal}
\author[telecomadress]{Yann Gousseau}
\address[telecomadress]{LTCI, T\'el\'ecom Paris, Institut Polytechnique de Paris, 19 Place Marguerite Perey, 91120 Palaiseau, France}
\address[UPSaclay]{Universit\'e Paris-Saclay, 91190, Saint-Aubin, France}

\begin{abstract}
Weakly supervised object detection (WSOD) using only image-level annotations has attracted a growing attention over the past few years. Whereas such task is typically addressed with a domain-specific solution focused on natural images, we show that a simple multiple instance approach applied on pre-trained deep features yields excellent performances on non-photographic datasets, possibly including new classes. The approach does not include any fine-tuning or cross-domain learning and is therefore efficient and possibly applicable to arbitrary datasets and classes. We investigate several flavors of the proposed approach, some including multi-layers perceptron and polyhedral classifiers. 
Despite its simplicity, our method shows competitive results on a range of publicly available datasets, including paintings (People-Art, IconArt), watercolors, cliparts and comics and allows to quickly learn unseen visual categories.
\end{abstract}


\end{frontmatter}

\section{Introduction}

The task of object detection has witnessed great progresses over the last few years, most notably through the development of clever and pragmatic combinations of region proposal methods and deep neural network architectures~\citep{ren_faster_2015}. Nevertheless, the training of such architectures is well known to necessitate huge databases of manually annotated images. In the case of object detection, these annotations are extremely costly. It requires around one minute for a non expert to draw a bounding box around an object \citep{su_crowdsourcing_2016}. For more specialized datasets, such as artworks databases for instance, experts are likely to be reluctant to such annotations. The usual way to annotate such databases is to rely on specialized micro-tasks platforms such as Amazon Mechanical Turk. This, by creating social exploitation and excessive precariousness, poses serious ethical concerns \citep{tubaro_microwork_2019}. For these reasons, reducing the annotation stage is of great importance. In particular, many Weakly Supervised Object Detection (WSOD) methods have been developed~\citep{bilen_weakly_2016,zhu_soft_2017,tang_weakly_2018} in order to train detection architectures using annotations only at image level, thus avoiding the precise localization of objects. 

On the other hand, many different image modality exist for which object detection is desirable. Such modality include photographs taken in difficult conditions, as it is common in the case of autonomous driving~\citep{vu_advent_2019}, different imaging modality as in medical~\citep{yang_unsupervised_2019} or satellite imaging~\citep{li_adaptive_2018} or even hand created images such as artworks, clipart, etc. In such cases, available databases may be small and it is essential to be able to reuse information gathered on existing large photographic databases, a strategy known as domain adaptation~\citep{saenko_adapting_2010}.

In particular, methods for the weakly supervised detection of objects have been developed to deal with domain adaptation. But while this problem has been extensively studied for photographic images, much less attention has been paid to WSOD in the case of strong domain shifts, as in the case of non-photographic images, possibly including domain-specific visual category. Some works focus on cross-domain weakly supervised object detection (i.e. where bounding boxes are available for the same visual category but in an other domain than the target one), as in \citep{inoue_crossdomain_2018,fu_deeply_2020}.

Methods that detect objects in photographs have been developed thanks to massive image databases on which several classes (such as cats, people, cars) have been manually localised with bounding boxes. The PASCAL VOC~\citep{everingham_pascal_2007} and MS COCO~\citep{lin_microsoft_2014} datasets have been crucial in the development of detection methods and the more recent Google Open Image Dataset (2M images, 15M boxes for 600 classes) is expected to push further the limits of detection. Even though large databases of artistic images have been  build by many cultural institutions or academic research teams, e.g.~\citep{rijksmuseum_online_2018,met_image_2018,wilber_bam_2017}, these databases include image-level annotations and, to the best of our knowledge, none includes location annotations. Besides, manually annotating such large databases is tedious and must be performed each time a new category is searched for.  There is therefore a strong need for methods permitting the weakly supervised detection of objects for non-photographic images. In particular, only a few studies have been dedicated to the case of painting or drawings.

Moreover, these studies are mostly dedicated to the cross depiction problem: they learn to detect the same objects in photographs and in paintings, in particular man-made objects (cars, bottles ...) or animals. While these may be useful in some contexts, it is obviously needed, e.g. for art historian, to detect more specific objects or attributes such as ruins or nudity, and characters of iconographic interest such as Mary, Jesus as a child or the crucifixion of Jesus, for instance. These last categories can hardly be directly inherited from photographic databases.

In this work, we take interest in weakly supervised object detection in the case of extreme domain shifts, namely non-photographic images, possibly addressing the detection of new, never seen classes.  We claim that an efficient way to perform this task is to rely on a simple Multiple Instance Learning (MIL) paradigm that is applied directly to the deep features of a pre-trained network. This approach does not involve any cross-domain learning step and can therefore be applied to arbitrary datasets and classes. Beside being efficient, as we will see in the experimental section, such a strategy also enables one to have relatively small training times. First, no fine-tuning is involved and second, we introduce a MIL strategy  that is much lighter than the classical SVM approaches~\citep{andrews_support_2003}. 

In order to illustrate the usefulness and efficiency of the approach, we focus on databases of man-made images, namely paintings, drawings, cliparts or comics. This poses a serious challenge because of both the lack or scarcity\footnote{Classical databases used for training networks are made of millions of natural images (Imagenet \citep{russakovsky_imagenet_2014}(millions of images), PASCAL VOC~\citep{everingham_pascal_2007}, MS COCO~\citep{lin_microsoft_2014} Google Open Image Dataset (9M images) \citep{kuznetsova_open_2020}). In contrast, datasets for recognition in non-photographic images are rare and usually only containing image-level annotations, as in the iMet dataset (375k) \citep{zhang_imet_2019} or BAM! (2.5M) \citep{wilber_bam_2017}. The very few datasets with bounding boxes such as PeopleArt \citep{westlake_detecting_2016}, used later in this paper, are very small.}  of annotated databases and the great variety of depicting styles. Being able to detect objects in such image modality has become an important issue, mostly because of the large digitization campaigns of fine arts. These include digital scans and photographs of artworks (mainly done by the museums and other public institutions) and scans of archive photographs (such as the Cini Foundation archive \citep{seguin_new_2018}).

In a previous conference paper~\citep{gonthier_weakly_2018} we have shown that the proposed method is a valid strategy when dealing with extreme domain shifts. In this paper, we fully develop the approach, exploring several extensions of the model such as a multi-layers version of the Multiple Instance perceptron and a polyhedral version obtained by aggregating several linear classifiers.
We also thoroughly evaluate the performances of the approach by comparing it to several state-of-the-art approaches on databases with challenging domain shifts, including paintings, drawings and cliparts. The experimental section shows that in such cases, the approach outperforms methods specially developed for the considered databases, as well as classical MIL approaches and some state-of-the-art WSOD approaches.

 The paper is organized as follows. In the next section we review WSOD algorithms and MIL methods as well as some deep learning applications to recognition tasks in non-photorealistic images. In section \ref{sec:PresentationModel}, we then present our algorithm as well as some of its variants. In section \ref{sec:experiments}, extensive experiments are presented, including comparisons to alternative algorithms and study of sensitivity of our method to its parameters.

\section{Related Work}
In this section we first review some state-of-the-art WSOD algorithms (an exhaustive review of this field is beyond the scope of the paper) and then explore MIL methods. Eventually, we make a brief survey of applications of deep learning for visual recognition in non-photographic images.

\subsection{Weakly Supervised Object Detection}
\label{sec:WSOD_relatedWork}

Computer vision methods often treat WSOD as a Multiple Instance Learning (MIL) problem \citep{dietterich_solving_1997}, especially in realistic cases where objects are not necessarily centered and with cluttered background~\citep{nguyen_weakly_2009,siva_weakly_2011,song_learning_2014,bilen_weakly_2016}.
In such cases, the image is viewed as a collection of potential instances of the object to be found (for example crops of various sizes and positions). 

A sketch of a typical weakly supervised detector is as follows:
\begin{enumerate}
\item Proposal generation: extract a certain number of regions of interest from the image. 
\item Feature extraction: compute a feature vector per region (off the shelf, handcrafted, CNN based\dots).
\item Classification: this is often done with a MIL algorithm to obtain an instance classifier.
\end{enumerate}

 These general steps can be alternated or entangled (for example to enhance the region proposition or feature extraction parts based on the performance of the final classifier).
In \citep{song_learning_2014} steps 1 and 2 are handled by extracting the features (and regions) proposed by RCNN \citep{girshick_rich_2014} . These features are passed to a smoothed version of SVM that serves as a MIL algorithm. Particular attention is paid to the initialization phase, which is crucial due to the fact that the MIL problem is essentially non-convex even if the SVM algorithm is. 

More recent methods tend to entangle all the mentioned steps in an end-to-end manner. For instance, some CNN based methods group feature extraction and classification \citep{bilen_weakly_2016,diba_weakly_2017,kantorov_contextlocnet_2016,tang_multiple_2017} whereas others group the three steps together \citep{zhu_soft_2017}. \cite{bilen_weakly_2016} propose a Weakly Supervised Deep Detection Network (WSDDN)  based on Fast RCNN \citep{girshick_fast_2015}. It consists in transforming a pre-trained network by replacing its classification part by a two streams network (a region ranking stream and a classification one) combined with a weighted MIL pooling strategy. 
 This work has been improved in many ways~\citep{wan_minentropy_2018,kantorov_contextlocnet_2016,zhang_zigzag_2018,zhang_w2f_2018,dong_dualnetwork_2017,wan_cmil_2019}. For instance, \cite{tang_weakly_2017} refine the prediction iteratively through multistage instance classifier. Later, this model was improved by adding a clustering of the region proposals \citep{tang_weakly_2018}. In \citep{wan_minentropy_2018}, the WSDDN model has been improved by adding two entropy term at the loss function to minimize the randomness of  object localization during learning, whereas in \cite{wan_cmil_2019}, the authors propose to tackle the non-convexity of the MIL pooling by using a series of smoothed loss functions.

In \citep{li_weakly_2016}, a two steps strategy is proposed, first collecting good regions by a mask-out classification, then selecting the best positive region in each image by a MIL formulation and then fine-tuning a detector with those propositions acting as ground truth bounding boxes. This pseudo-labeling step is often used in the weakly supervised pipeline. In \citep{zhu_soft_2017} a region proposal generator is built using weak supervision. The feature maps are transformed into a graph then into an objectness score map. This objectness score ponderates the feature maps that are subsequently fed to a classification layer. In \citep{arun_dissimilarity_2019} the authors proposed to train two collaborative networks one of it being a Conditional Network with noisy extra-channel. The goal is to jointly minimize the dissimilarity between the prediction distribution and the conditional distribution.

It is worth noting that although CNN feature maps contain some localization information \citep{oquab_is_2015}, the main difficulty for weakly supervised detection is the construction of an efficient box proposal model. Most works in the field use effective unsupervised methods for region proposals such as Selective Search \citep{uijlings_selective_2013} or EdgeBoxes \citep{zitnick_edge_2014}.

\subsection{Generic Multiple-Instance Learning}
\label{sec:GenericMIL}

 As stated above, the problem of weakly supervised object detection can be recast into a multiple instance learning (MIL)  problem \citep{dietterich_solving_1997}. More precisely, we are interested in instance classification as opposed to bag classification. We want to find an object among several candidate boxes in order to detect the object of interest. In \citep{andrews_support_2003}  a solution based on iterative applications of a Support Vector Machine (SVM) has been proposed to solve the MIL problem. Actually two flavors are considered, mi-SVM and MI-SVM. In the case of mi-SVM, each element of positive bags is assigned a label and the SVM margin is imposed at the instance level. In the case of MI-SVM, the SVM margin is imposed the most positive element of each positive bag and to the least negative element of each negative bag. In both cases, at test time, the learned classifier can be applied at the instance level. In \citep{felzenszwalb_object_2010} a reformulation of MI-SVM is proposed and called latent SVM (LSVM). But in this work, a bag of instance represents the set of parts of an object and the MIL formulation is used to train an object detector with a fully-supervised training. 
 
Several heuristics to solve the non convex-problem posed by the MIL have been proposed. For example, in  \citep{gehler_deterministic_2007} is introduced a new objective function that try to estimate the quantity of positive examples in a positive bag, before using deterministic annealing to optimize it. In contrast to the MI-SVM method, the algorithm can consider several elements as positive in the positive bag. In \citep{joulin_convex_2012}, the authors propose a convex relaxation of the softmax loss.  A comprehensive review of SVM based MIL methods can be found in   \citep{doran_theoretical_2014}. From this review it appears that mi-SVM and MI-SVM are still competitive on the tasks studied there.

\begin{figure} %
    \centering
    \includegraphics[height=3.5cm]{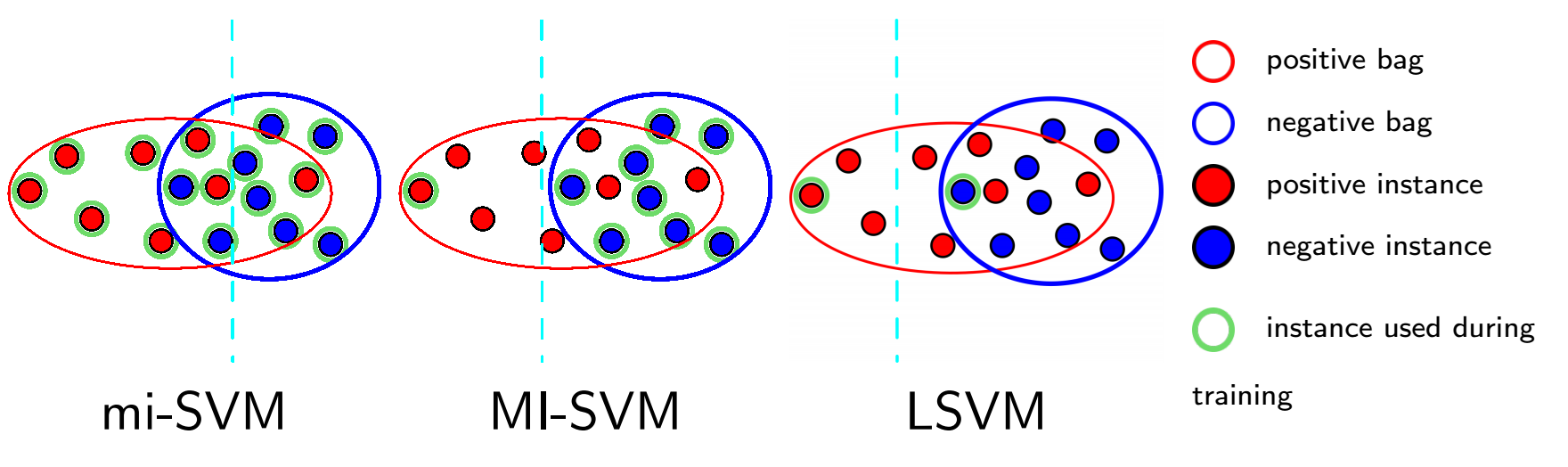}
    \caption{Comparison of standard SVM based MIL models. The blue dotted lines show the hyperplanes learned by the models, and the blue circles show the instances used during the SVM training. Figure must be seen in color.}
    \label{fig:MIL_difference}
\end{figure}

Figure \ref{fig:MIL_difference} summarizes the instances on which the SVM margins are imposed in the most popular SVM based MIL methods. 

 Another approach to the MIL problem is to use neural networks whose architecture treats each instance symmetrically, before an explicit aggregation (max, average) is performed. From this point a classical neural network performs a classification task  \citep{ramon_multi_2000,zhou_neural_2002}. An improvement  using more recent deep learning building blocks is proposed in \citep{wang_revisiting_2018}.  The aforementioned works did not focus on the instance classification performance. They all, by design, provide an instance classification network (present the network with a bag consisting of one item).

From a recent survey~\citep{carbonneau_multiple_2016} on multiple Instance Learning it appears that the most efficient algorithm for an instance level classification seems to be a clever variation of bagging and multiple classifiers to deal with multi-modal distributions \citep{carbonneau_robust_2016}.

 Based on these surveys, we are driven to propose a method that mimics an SVM within a neural network. The main difference between our approach and the SVM based MIL methods is that iterations are performed during the training of the neural network and the multi-modal nature of the objects to be found drives us to consider multiple linear classifiers of each considered class.

\subsection{Deep Learning for visual recognition in non-photographic images}

As almost all applications of computer vision, tasks dealing with hand-drawn or computer generated non-photographic images benefited from the resurgence of neural networks. One point in common between all works in the field is the reuse of architectures that where originally designed for photographs classification. Some works use the pre-final features of a network as the only features retained to represent an image and do not fine-tune the network for the task at hand. Other methods allow for a certain amount of fine-tuning and add a specific network after the original architecture.  Another significant difference between the papers we are going to cite is whether or not the considered classes where present in the training dataset of the original network. 
In the simplest setting, features from a pre-trained network are retained and used to train a linear SVM  \citep{crowley_search_2014,crowley_visual_2016}, the task being the recognition of classes already present in the original training set the network was pre-trained on.

Several works have also shown that pre-trained CNN architecture can be efficiently transferred for learning new semantic visual categories, those networks either being used as features extractors \citep{crowley_search_2014,crowley_visual_2016} or being fine-tuned \citep{yin_object_2016,strezoski_omniart_2018,wilber_bam_2017}. %

A large body of works investigate the fine-tuning of CNN for style recognition \citep{lecoutre_recognizing_2017,mao_deepart_2017,elgammal_shape_2018}, material \citep{sabatelli_deep_2018}, scene \citep{florea_domain_2017} or author classification \citep{vannoord_learning_2017}.
The use of CNN also opens the way to efficient artwork analysis tasks, such as visual links retrieval \citep{seguin_visual_2016}, posture estimation \citep{jenicek_linking_2019}, visual question answering \citep{bongini_visual_2020} and instance recognition \citep{shen_discovering_2019,delchiaro_weblysupervised_2019}.  
Some works try to tackle several of those tasks at the same time
\citep{garcia_contextaware_2019,bianco_multitask_2019}. A survey about machine learning for cultural heritage have been recently published \citep{fiorucci_machine_2020}.

The object detection problem (recognize and locate an object) in artworks has been less studied.
In \citep{westlake_detecting_2016} and \citep{strezoski_omniart_2018} it is proposed to fine-tuned a detection network in a fully supervised manner to detect people and classical Pascal VOC classes, respectively.
In \citep{inoue_crossdomain_2018}, an efficient pipeline is proposed to train a detector on new artistic modalities in a semi-supervised manner. This approach requires natural images with bounding boxes annotation of those classes and involves a relatively costly style transfer procedure. In particular, this method only allows the detection of object classes that are present and have been annotated in natural images. This specific problem have been recently studied by different research teams \citep{saito_strongweak_2019,fu_deeply_2020}. The same is true for many works focusing on recognizing the same object categories in different modalities \citep{li_deeper_2017,wilber_bam_2017,thomas_artistic_2018}.
Only very few work have focused on visual categories that are new and specific to artworks \citep{lang_finding_2019,gonthier_weakly_2018}.
In \citep{lang_finding_2019}, the authors proposed an interactive search engine to detect objects in artistic images for  object categories such as praying hands, cross or grape.
In \citep{gonthier_weakly_2018}, the authors proposed a simple MIL classifier coupled with Faster RCNN \citep{ren_faster_2015} to weakly learn to detect new visual categories such as Mary or Saint Sebastian. The present work extends the MIL model proposed in this paper by allowing polyhedral classification and evaluate its performances on various modality such as paintings, drawings or cliparts.

\section{Multiple instance perceptron for the weakly supervised detection of objects}
\label{sec:PresentationModel}
 
 In this section, we first give the general motivation behind this work, before recalling the classical MIL framework and then introducing our approach.

\subsection{Motivation}

\begin{figure}
    \centering
  \includegraphics[height=5.5cm]{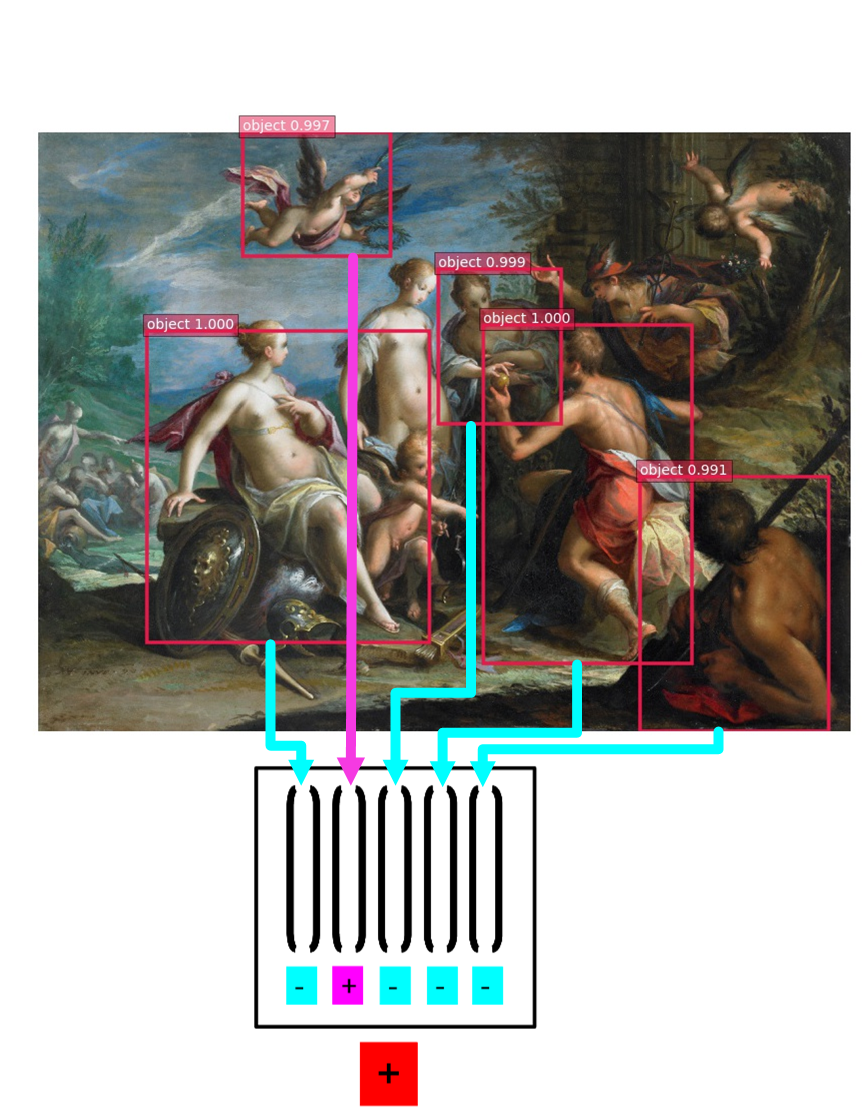} \hspace{2cm}
\includegraphics[height=5.5cm]{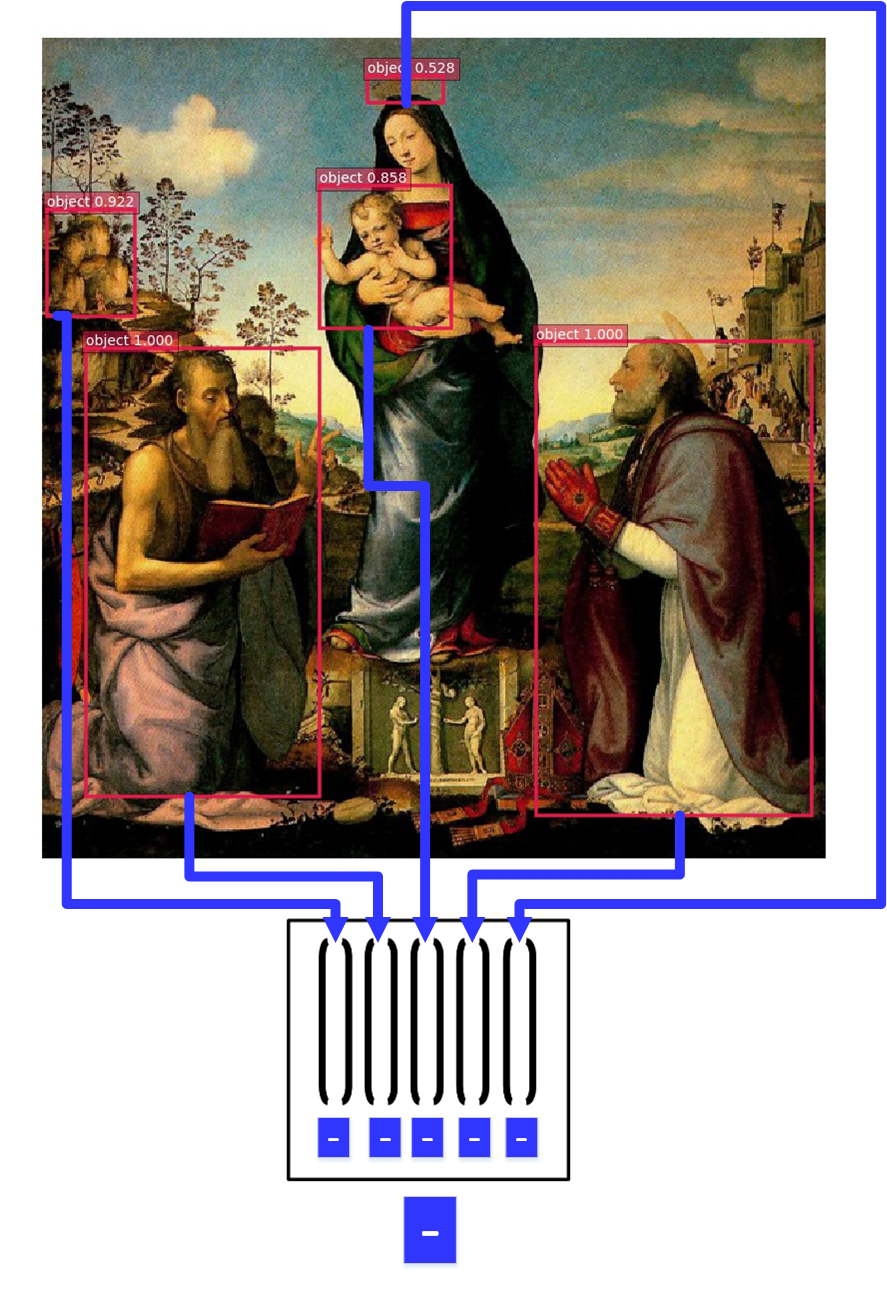}
    \caption{Illustration of positive and negative sets of detections (bounding boxes) for the {\it angel} category.}
    \label{fig:angel-Bag}
\end{figure}

As explained earlier, we tackle in this paper the problem of weakly supervised object detection (WSOD) in the following sense : we assume that for each image to be analyzed,  bounding boxes are available, together with a global classification information. Figure~\ref{fig:angel-Bag} illustrates the situation we face at training time. For each image and for a given category, we are given a set of bounding boxes and a global label, equal to $+1$ (the visual category of interest is present at least once in the image) or $-1$ (the category is not present in this image). 

Since we are especially interested by non-photographic images, for which databases may be limited, we wish to keep the learning step as light as possible. We therefore choose to combine a pre-trained detector with a classical MIL strategy. For the task of instance level classification, this approach can be used to weakly transfer an object detector to a new domain or to new visual category. 

Now, the MIL framework involves the minimisation  of a non-convex energy, which results in heavy computational costs. For this reason, efficient relaxation schemes have been proposed~\citep{joulin_convex_2012}. In this paper we propose a simple and fast heuristic to this problem, together with several variants. 
This, combined with the fact that we avoid fine-tuning by using features extracted from pre-trained CNNs, permits a flexible on-the-fly learning of new category in a few minutes.

\subsection{The MIL framework}

We give here some basic notations related to Multiple Instance Learning. 
Let $\mathcal{B} = \{B_1,B_2,\dots B_N \} $ denotes a set of N {\it bags}, each bag $B_i$ being a collection of feature vectors (instances) : $\{X_{i,1},X_{i,2},\dots X_{i,K_i} \}$ where $X_{i,k} \in \mathbb{R}^M$. To each feature $X_{i,k}$ is associated a label $y_{i,k}$. In the MIL framework, each bag is associated a label which is positive if at least one instance is positive, and negative if all instances are negative. That is, the bags labels $Y_{i}$ are defined as :
$$ Y_i=\left\{\begin{array}{ll}{+1} & {\text { if } \exists ~ k \in \{1,\dots,K_i\}: y_{i,k}=+1} \\ {-1} & {\text { if } \forall k \in \{1,\dots,K_i\}: y_{i,k}=-1}\end{array}\right.$$
In this paper we consider the task of instance level classification, that is the task of infering the unknown instance labels $y_{i,k}$ from the known bag labels. Another classical MIL problem is the one of bag-level classification. 

 In an object detection setting each feature vector will represent a region. As in a typical classification problem, the goal is to learn a prediction function $f_w$, parametrized by $w$, so that the predicted output $f_w(X) = \hat{Y}$ minimizes the empirical risk.
The typical way to do so is to minimize a loss function that measures the correctness of the prediction over the training examples. %

There are two main ways to tackle the fact that we only have bag level ground truth information.

First, one can aggregate all the predictions of one bag to a single prediction (at bag level) during training.
Hence we can write $\hat{y}_i = g ( \{ \hat{y}_{i,k} \}_{k \in \{1 \dots K_i\}}) $ with $g$ an aggregation function over the elements of a bag $i$. %
In this case, the loss function can be written as $L (Y_i, \hat{y}_i ) = l ( Y_i , g ( \{ \hat{y}_{i,k} \}_{k \in \{1 \dots K_i\}}) ) $.

Second, one can consider each instance of a bag individually (as in the mi-SVM case, see Figure \ref{fig:MIL_difference}) and the loss function can be written as $L (Y_i,\{ \hat{y}_{i,k} \}_{k \in \{1 \dots K_i\}}) =  g ( l ( h_{i,k} ( Y_i ) ,\{ \hat{y}_{i,k} \}_{k \in \{1 \dots K_i\}} ) )$ where $g$ is an aggregation function (usually an average), $l$ a penalty function and $h_{i,k}$ a modification function of the label associated to the instance  $k$ and depending on the bag label $Y_i$, usually named a latent label (see \citep{felzenszwalb_object_2010}). If we consider that the label of a bag is equal to the label of its instances, $h_{i,k}$ is the identity, otherwise it is a function from $ \{ -1,1 \} $ to $  \{ -1,1 \} $ depending on the bag and the instance.

\subsection{A multiple instance perceptron}
\label{sec:MainModel} 

In contrast with classical approaches to the MIL problem, such as~\citep{andrews_support_2003,carbonneau_robust_2016}, based on costly iterations of SVM or complex bagging methods, we propose a simple heuristic to solve the multiple instance problem.
It is a multiple instance extension of the perceptron  \citep{rosenblatt_perceptron_1958} with a maximum taken over the instances of a bag. Our model can be seen as a latent perceptron if we use the same designation as~\citep{felzenszwalb_object_2010}.

We denote our model {\bf \MILS{}} as introduced in \citep{gonthier_weakly_2018}. 
As we consider each class individually, we focus on the case of binary classification.

We build on a linear model $f_w(X_{i,k}) =   W^{T} X_{i,k} + b$ with $W \in \mathbf{R}^M$, $b\in\mathbf{R}$, which we combine with a maximum aggregation function $g = max_{k \in  \{1 \dots K_i\}}$ and a per example loss function equal to 
\begin{equation}
\label{eqn:perexampleloss}
l(y,\hat{y}) = 1 - y \mathop{Tanh}(\hat{y}) = 1 - \mathop{Tanh}(y\hat{y}).
\end{equation}
We also use a regularization term on the norm of $W$ and a weighting of the two classes, so that the complete loss function is: 

\begin{equation}
\mathcal{L} (W,b) = 2  - \sum_{i=1}^{N} \frac{Y_{i}}{n_{Y_i}} \mathop{Tanh} \left( \max_{k \in \{1 \dots  K_i \} }  \left(  W^{T} X_{i,k} + b \right) \right)+ C ||W||^2,
\label{eqn:LossReg}
\end{equation}
with $n_1$ the number of positive examples in the training set and $n_{-1}$ the number of negative examples.

As mentioned before, the intuition behind this formulation is that minimizing $\mathcal{L} (W,b) $ amounts to seek a hyperplane separating the most positive element of each positive image from the least negative element of the negative image (i.e. from all examples in the negative bags). Also this loss seeks to maximize the margin.

 If the hyperplane  $W^TX+b=0$ exactly separates the most positive examples of each positive bag from the set of all examples of all negative bags, then replacing $C,W$ and $b$ by $\lambda C,\frac{1}{\lambda} W$ and $\frac1{\lambda} b$ respectively and taking $\lambda$ to 0 will lead to a loss as close to 0 as desired. This implies that if the MIL problem admits an exact linear solution, then our loss accepts it provided $C$ is small enough. 
 In the worst case scenario, its value is 4 (plus the regularization term).

 One advantage of this formulation is that it can be tackled by a simple gradient descent, therefore avoiding the very costly iterative procedures of other MIL solutions such as~\citep{andrews_support_2003}. Taking the $\max$ over all instance of a bag is akin to what is done in MI-SVM (mentioned in section \ref{sec:GenericMIL}) when after each full training of an SVM, a new representative element of each bag is selected for the next SVM training.
We can switch to a stochastic gradient descent by iterating on random batches when the dataset is too big.  Of course, since our loss is not convex, we are not guaranteed to find the global minimizer of the function. To tackle this problem, we run $r$ times the model with a random initialization and pick the best one on the training set evaluation of the loss function. 

 If we refer to the simple description of the WSOD standard pipeline, we only focus on the multiple instance classification task and not on the boxes proposals algorithms, features extraction or refinement methods mentioned section \ref{sec:WSOD_relatedWork}.

\subsection{From multiple instance learning to weakly supervised object detection in images} 
\label{sec:objectTrick}

In the context of Weakly Supervised Object Detection (WSOD), each bag $i$ corresponds to an image and each instance $k$ corresponds to a candidate region to be labeled. We here assume that candidate regions are returned by a classical detection network, together with a high level semantic feature vector of size M $X_{i,k}$ and a class-agnostic objectness score $s_{i,k}$. We ignore the classification ability of the detection network: no classification label is used.

For simplicity, we consider only one class.
Assume we have $N$ images, with $K$ bounding boxes. When an image is a positive example (the visual category is present), it is given an image-level label $Y_i = +1$ when it is ); otherwise it is given the label  $Y_i = -1$. The number of positive examples in the training set is denoted by $n_1$, and the number of negative ones by $n_{-1}$.
Training a WSOD model from scratch, especially when the database is rather small and from another domain, is a very hard problem. Thus, reusing as much as possible models that have been trained on large datasets is advisable. In this paper, we will rely on the faster RCNN detection network but other networks could be used. We assume that features are associated to each box. We do not rely on any classification information, but we assume that an objectness score is associated to each box. The idea is to give more importance to the classification of boxes with the highest score. We observed that using the class-agnostic objectness score attached to each proposed box consistently gave better results (see section \ref{sec:ablationStudy}).  We chose to multiply each $W^TX_{i,k}+b$ by the objectness score of the region $k$ before taking the maximum:

\begin{equation}
\label{eqn:scoreTrick}
f_w(X_{i,k}) =  \left( s_{i,k} + \epsilon \right)  \left(W^{T} X_{i,k} + b \right),
\end{equation}
with $ \epsilon \geq 0$ and where $s_{i,k}$ is the class-agnostic objectness score of the region $k$, as returned by the detection network. 
The motivation behind this formulation is that the score $s_{i,k}$, roughly a  clue that there is an object in box $k$, provides a prioritization between boxes. The same idea is used in the WSDDN model \citep{bilen_weakly_2016} or in MELM \citep{wan_minentropy_2018}. %

At test time, the instance level decision is made as before according to the sign of $\left(W^{\star T} x + b^{\star} \right)$, since multiplication by a positive score does not change the sign. Indeed, the hyperplane $W^\star,b^\star$ is chosen to separate two classes and the loss $\mathcal{L}$ aims at maximizing the margin with respect to this hyperplane. It stands to reason that the instance level classification must be related to the relative position of the instance and the hyperplane. Nevertheless, we will propose in section \ref{sec:experiments} a non maximal suppression strategy that will once again use the objectness score to filter the boxes proposed for each class. More precisely the non maximal suppression algorithm will use the following score:

\begin{equation}
    S(x) = \mathop{Tanh}  \lbrace \left( s(x) + \epsilon \right) \left(W^{\star T} x + b^{\star} \right) \rbrace
    \label{eq:scoreMILS}
\end{equation}
which mixes the objectness score $s(x)$ and the signed distance from the hyperplane $W^{\star T} x + b^{\star}$.

We now present two natural extensions of our core model. We first make use a neural network to transform the bare features $X_{i,k}$, so that the transformed features can be more relevant to the task at hand. Then, we investigate the interest of a polyhedral separation instead of a hyperplane for classification. 

\subsection{Extensions of our model}

\subsubsection{One hidden layer network}

 In this extension, called \textbf{\mimaxaddlayerS{}}, the bare features $X_{i,k}$ are transformed by a hidden layer before the \MILS{} approach is applied. This can be summarized by modifying the function $f_w$ as follows:
  $$f_w(X_{i,k}) = \Omega^{T} \left( \mathop{Tanh} \left( W^{T} X_{i,k} + b \right) \right) + \beta, $$ with $W \in \mathbf{R}^{L \times M}$, $b\in\mathbf{R}^{L}$, $\Omega \in \mathbf{R}^{L}$, $\beta\in\mathbf{R}$ and $L$ the dimension of the hidden layer. When compared with \MILS{} the parameters to be learned are $\Omega, \beta,W,b$ for a total dimension of $L+1+L\times M+L=L\times(M+2)+1$ compared to the original $M+1$ scalars. 
We keep the function $\mathop{Tanh}$ as activation function to be coherent with the previous model; using a ReLU instead has little effect on the performance.

\subsubsection{Multiple linear classifier model}

As mentioned in the introduction, an improvement of  the linear model consists in learning several hyperplanes in parallel, so that the binary classification is performed in a collaborative manner instead of selecting the best hyperplane. The contributions of several hyperplanes are gathered with a maximum function, so that the model can be defined as: 

$$f_w(X_{i,k}) = max_{j \in \{1\dots r\} } \left( W_j^{T} X_{i,k} + b_j \right)$$

At each iteration of the gradient descent only one of the couple $(W_j, b_j)$ is updated. For the inference the $r$ hyperplanes are used. 

This model, named \textbf{\MaxOfMaxS{}} yields a concave polyhedral boundary between the two classes.  The concept of convex polyhedral separability has been introduced by \cite{megiddo_complexity_1988} and well studied in the framework of polyhedral and piece-wise linear classifier. In our case, this allows one to get more complex boundary at a modest extra-cost compared to a kernel SVM. 

These models will be experimentally compared in section \ref{sec:experiments}. %

\subsection{Discussions}

The MIL part of our model \mimaxaddlayerS{} is close in spirit to the multiple instance neural networks proposed by \cite{ramon_multi_2000} and \cite{zhou_neural_2002}\footnote{These models involve a sigmoid activation and they are trained with a quadratic loss $l(y,\hat{y}) = (y - \hat{y})^2$ and no re-initialization ($r=0$).} and further extended in~\citep{wang_revisiting_2018}. 
The best way to aggregate instance level predictions in order to find a classifier separating each of the individual vectors $X_{i,k}$ of each bag at test time is still an open-problem. Some works use the max operator \citep{zhou_neural_2002}, the average operator or the Log-Sum-Exponential \citep{ramon_multi_2000} for the pooling. Indeed, since the training is done with only bag level information, at test time the learned classifier must be able to handle each instance almost independently from the others because of the variety of objects that may appear in the test image. 

None of these works use such approach for instance level classification and even less for weakly supervised object detection. We include in the experimental comparisons some applications (that we will call MI\_net or mi\_net \citep{wang_revisiting_2018}) of this MIL methodology to the same deep features used in our method. These can be seen as variations on the general approach proposed in this paper.

\section{Experiments}
\label{sec:experiments}

\subsection{Experimental Setup}

{\bf Features extraction:}
We use the Faster RCNN  detection network~\citep{ren_faster_2015} as a feature extractor and region proposal algorithm. We extract 300 regions per image along with their high-level features\footnote{The output of layer fc7 often called 2048-D.} and the class-agnostic objectness score attached to each proposed box by the Region Proposal Network (RPN). Let us stress that, by using Faster R-CNN, our system uses a subpart that has been trained on databases with bounding boxes ground truth. In WSOD setups such as \citep{bilen_weakly_2016,zhu_soft_2017,tang_pcl_2018}, the models have not seen any bounding boxes, even on different modality. Observe nevertheless that, in contrast with domain adaptation methods such as \citep{inoue_crossdomain_2018}, our method allows the detection of new classes.

According to \cite{kornblith_better_2018}, the ResNet family of networks appears to be the best architecture for transfer learning by feature extraction. Among this family we chose ResNet 152 layers trained on MS COCO \citep{lin_microsoft_2014}. Therefore, the backbone we used has been trained on ImageNet, then fine-tuned on MS COCO. Remember that we chose not to fine-tune the backbone in order to provide a fast and flexible tool that can be used on small data sets. As a consequence, the backbone of our model only saw photographs for its two-phase training (ImageNet, MS COCO).

{\bf Parameters of the models:} For training our MIL models, we use a batch size of 1000 examples (for smaller sets, all features are loaded into the GPU), 300 iterations of gradient descent for the linear model, performed with a constant learning rate of 0.01 and  $\epsilon=0.01$ and $C=1$  (equations \eqref{eqn:scoreTrick} and \eqref{eqn:LossReg}). The complete training takes about 6 minutes for 7 classes on the IconArt dataset \citep{gonthier_weakly_2018} with 12 random starting points per class using a consumer GPU (GTX 1080Ti).  In the case of  \MaxOfMaxS{} and \mimaxaddlayerS{} we used 3000 iterations which increase the training time to 1 hour. For \mimaxaddlayerS{}, we use a maximum batch size of 500 elements.
Actually, the random restarts and classes are performed in parallel to take advantage of the presence of the features in the GPU memory, thus reducing the GPU-CPU transfer times. Typically, 20 classes can be learned in parallel on a standard GPU, due to the light weight of the model. One of other the advantage of not fine-tuning the network is that there is no need to store the heavy weights of the new trained model.

\subsection{Results and comparison to other methods}
\label{sec:ExpResults}

In this section, we perform weakly supervised object detection experiments on different databases.%
We compare our different models \MILS{}, \MaxOfMaxS{} and \mimaxaddlayerS{}, to the three types of methods. %

The first group of methods are those specifically targeted at WSOD using fine-tuned networks. We have included state-of-the-art methods for which a source code is available: Soft Proposal Network\footnote{ Trained with the following hyperparameters: batch size = 16, learning rate = 0.01, multi-scale strategy with image of sizes 112, 224 and 560, with 20 epochs. There is no regularization term in this method.} (SPN \citep{zhu_soft_2017}) and Proposal Cluster Learning\footnote{ Trained with the following hyperparameters: batch size = 2, learning rate = 0.001, decay=0.0005, step decay = 7, momemtum of 0.9 and defaut number of clusters (3), with 13 epochs. Those parameters correspond to the ones used by the authors for the Pascal VOC07 dataset. There is no regularization term in this method either.} (PCL \citep{tang_pcl_2018}). For some of the datasets, we also include results from the Weakly supervised detection network (WSDDN \citep{bilen_weakly_2016}) from \citep{inoue_crossdomain_2018}.  For those datasets we also show the performance obtained by the mixed supervised method with domain adaptation proposed by \citep{inoue_crossdomain_2018}, a method that assume that datasets with bounding boxes for the same classes on different modality are available.

The second family of methods are generic MIL-methods directly applied to the set of deep features vectors generated by Faster RCNN. Observe that these methods ignore the objectness scores returned by the detection network. The first ones are MI-SVM and mi-SVM\footnote{ We allow up to 50 iterations of the algorithm (i.e. the complete training of a SVM for each class). We experimentally observe that the re-initialization of the model does not improve the performance in our case.} from \citep{andrews_support_2003}. These two methods require to train several SVMs and are therefore costly. In some cases (for the datasets PeopleArt and IconArt) we performed a PCA on the training set to reduce the number of components from 2048 to around 650 dimensions by keeping 90\% of the variance (to fit the SVM in the CPU memory). We experimentally observed on the other datasets that this dimensionality reduction doesn't reduce the performances. Eventually, the computationally lighter MI\_Net, MI\_Net with Deep Supervision (DS) or Residual Connection (RC) and mi\_Net from \citep{wang_revisiting_2018} are also considered\footnote{For this method, we consider the following hyperparameters: three fully-connected layers with 256, 128 and 64 hidden units, a kernel l2 regularization with a weight equal to 0.005, an initial learning rate equal to 0.001 with a momentum of 0.9 and a decay of $10^{-4}$ for 20 epochs}. Although those models are designed for bag level classification, we used them for instance level prediction. Again, these can be seen as variants on the method we develop in this paper (the weakly detection of objects is not addressed in \citep{wang_revisiting_2018}). 

The last type of methods are those who (before any training) use the objectness score of the proposed regions to keep only one feature vector for each positive image. The method \MAX{} keeps one feature vector per image and learns a linear SVM classifier that separates the positive vectors from the negative one \citep{crowley_art_2016}. The variant \MAXA{} also keeps one vector per positive image but uses all vectors from the negative ones. Again, a linear SVM is learned. In both cases a 3-fold cross validation is performed for determining the main hyperparameter of the SVM.

At test time, the labels and the bounding boxes are used to evaluate the performance of the methods in term of Average Precision par class. The generated boxes are filtered by a NMS with an IoU threshold of 0.3 \citep{everingham_pascal_2007} and a confidence threshold of 0.05 for all methods.

\begin{savenotes}
\begin{table*}[h]
\centering
\caption{Overall information of the evaluated datasets.}
\label{tab:datasets_recap}
  \resizebox{\textwidth}{!}{
\begin{tabular}{|c|c|c|c|c|c|c|c|c|}
\hline
\multirow{2}{*}{Reference} & \multirow{2}{*}{Dataset} & \# Images & \# Images & \# Instances & \multirow{2}{*}{\# Classes} & Min \# Images & Classes from & Classes from \\ 
 &  & in train  & in test  & in test  &  & per class  & natural images  & Pascal VOC  \\
\hline
\citep{westlake_detecting_2016} & PeopleArt &  3007 & 1616  & 1137 & 1 & 968 & Yes & Yes \\
\citep{inoue_crossdomain_2018} & Watercolor2k & 1000 & 1000  & 3315  &  6 & 27 & Yes & Yes  \\
\citep{inoue_crossdomain_2018} & Clipart1k & 500 & 500 & 3615 & 20 & 21  & Yes & Yes \\
\citep{inoue_crossdomain_2018} & Comic2k & 1000 & 1000 &  6389 & 6 & 87 & Yes & Yes \\
\citep{thomas_artistic_2018} & CASPA paintings & 1045 & 1033  & 1486  & 36 & 8 & Yes & 6 out of 8  \\
\citep{gonthier_weakly_2018} & IconArt & 2978 & 1480 & 3009 & 7 & 75 & No & No \\
\hline
\end{tabular}
}
 \end{table*}  
 \end{savenotes}

\begin{savenotes}
\begin{table*}[h]
\centering
\caption{\textbf{People-Art (test set)} Average precision (\%). Comparison of the proposed \MILS{}, \MaxOfMaxS{} and \mimodels{} methods to alternative approaches. 
In red the best weakly supervised method.} %
\label{tab:People-Art_detection}
\begin{tabular}{|c|c|c|c|}
\hline
Network &  Method & Model & person \\
\hline \hline
 \multirow{2}{*}{VGG16-IM} &  Weakly supervised  & SPN \citep{zhu_soft_2017}  & 10.0 \\
  & fine tuning   &  PCL \citep{tang_pcl_2018} & 3.4 \\
\hline \hline
\multirow{11}{*}{\begin{tabular}{c}RES- \\ 152- \\ COCO\end{tabular}}  & \multirow{11}{*}{\begin{tabular}{c}Features \\  extraction\end{tabular}}  &  \MAX{} \citep{crowley_art_2016} & 25.9 \\
& & \MAXA{}  &   48.9  \\
\cline{3-4}
& & MI-SVM  \citep{andrews_support_2003} & 13.3  \\ %
& & mi-SVM \citep{andrews_support_2003} &  5.6 \\ %
& &  MI\_Net \citep{wang_revisiting_2018} &  33.0  $\pm$ 6.0 \\
 & & MI\_Net\_with\_DS \citep{wang_revisiting_2018} &19.5  $\pm$ 11.4  \\
& &   MI\_Net\_with\_RC \citep{wang_revisiting_2018}  & 12.5  $\pm$ 8.3  \\ 
 & & mi\_Net \citep{wang_revisiting_2018}   &  26.5  $\pm$ 8.5  \\ \cline{3-4}
 & & \MILS{}   &  55.5 $\pm$ 1.0 \\
 & & \MaxOfMaxS{} &  {\color{red} \bf 58.3}  $\pm$ 1.2  \\ 
& & \mimaxaddlayerS{}   & 57.3  $\pm$ 2.0 \\
\hline
\end{tabular}
 \end{table*}  
 \end{savenotes}

\begin{savenotes}
\begin{table*}[h]
\caption{\textbf{Watercolor2k (test set)} Average precision (\%).  Comparison of the proposed \MILS{}, \MaxOfMaxS{} and \mimodels{} methods to alternative approaches. 
In green the best mixed supervised method and in red the best weakly supervised one.}
\label{table:Watercolor2k_detection}
 \resizebox{\textwidth}{!}{
\begin{tabular}{|c|c|c|cccccc|c|}
\hline
 Net & Method & Model  & bike &bird &car &cat &dog &person & mean \\  \hline \hline
SSD & Mixed + DA & DT+PL \citep{inoue_crossdomain_2018} \footnote{The performance comes from the original paper \citep{inoue_crossdomain_2018}.}   & 76.5 & 54.9 & 46.0 & 37.4 & 38.5 & 72.3 & {\color{olive} \textbf{54.3}}$^{\star}$ \\
\hline \hline
 \multirow{3}{*}{\begin{tabular}{c}VGG16 \\IM \end{tabular} } & \multirow{3}{*}{\begin{tabular}{c}
Weakly \\
supervised\\
fine-tuning
\end{tabular}  } & WSDDN  \citep{bilen_weakly_2016} \repeatfootnote   &  1.5 & 26.0 & 14.6 & 0.4 & 0.5 & 33.3 & 12.7 \\
 & & SPN \citep{zhu_soft_2017} & 0.0 & 18.9 & 0.0 & 0.0 & 0.0 & 23.6 & 7.1 \\ 
 & & PCL \citep{tang_pcl_2018} & 0.0 & 0.0 & 0.0 & 0.0 & 0.0 & 0.0 & 0.0 \\
\hline \hline
\multirow{11}{*}{\begin{tabular}{c}
RES-  \\
152- \\
COCO
\end{tabular}} & \multirow{11}{*}{\begin{tabular}{c}Features \\ extraction\end{tabular}} & \MAX{} \citep{crowley_art_2016} &  76.0 & 33.8 & 33.0 & 20.8 & 22.7 & 19.8 & 34.3 \\
& & \MAXA{}  & 60.6 & 39.2 & 39.6 & 30.9 & 32.0 & 61.2 & 43.9 \\
\cline{3-10}
& & MI-SVM \citep{andrews_support_2003}  & 66.8 & 20.9 & 7.6 & 14.1 & 8.5 & 13.2 & 21.8 \\ %
& & mi-SVM  \citep{andrews_support_2003} & 10.6 & 10.9 & 1.4 & 2.0 & 0.8 & 5.9 & 5.3 \\ %
& &  MI\_Net \citep{wang_revisiting_2018} & 77.6   & 32.4   & 35.5  & 24.7   & 16.2  & 18.0   & 34.1   $\pm$ 1.0   \\  
 & & MI\_Net\_with\_DS \citep{wang_revisiting_2018}  & 73.4   & 22.4   & 25.8   & 17.6 & 11.2  & 10.3 & 26.8   $\pm$ 2.4   \\
& &   MI\_Net\_with\_RC \citep{wang_revisiting_2018}& 32.3   & 19.2   & 20.1   & 6.7  & 6.8  & 15.4   & 16.7   $\pm$ 6.3   \\  
 & & mi\_Net \citep{wang_revisiting_2018} & 66.4  & 30.3   & 14.9   & 14.4   & 8.6   & 20.5   & 25.8   $\pm$ 3.5   \\
\cline{3-10}
& & \MILS{}   & 84.1  & 47.4 & 48.2  & 30.9  & 27.9  & 58.2  & {\color{red} \bf 49.5 }  $\pm$ 0.9   \\ 
& & \MaxOfMaxS{}  & 77.8   & 44.7   & 45.5   & 25.6  & 26.7  & 59.2   & 46.6   $\pm$ 1.3   \\ 
& & \mimaxaddlayerS{} & 79.3  & 46.1  & 43.6  & 26.9  & 28.8  & 57.0  & 47.0   $\pm$ 1.6   \\ 
\hline
\end{tabular}
}
\end{table*}
\end{savenotes}

\begin{savenotes}
\begin{sidewaystable*}[h]
\caption{\textbf{Clipart1k (test set)} Average precision (\%).  Comparison of the proposed \MILS{}, \MaxOfMaxS{} and \mimodels{} methods to alternative approaches. In those case, we use a line search for \MAX{} and \MAXA{}. In green the best mixed supervised method and in red the best weakly supervised one.}
\label{table:Clipart1k_detection}
  \resizebox{\columnwidth}{!}{
\begin{tabular}{|c|c|c|cccccccccccccccccccc|c|}
\hline
 Net & Method & Model  &  aeroplane  &  bicycle  &  bird  &  boat  & bottle  &  bus  &  car  &  cat  &  chair  &cow  &  diningtable  &  dog  &  horse  &motorbike  &  person  &  pottedplant  &sheep  &  sofa  &  train  &  tvmonitor &  mean \\  \hline \hline
SSD & Mixed supervised  & DT+PL \citep{inoue_crossdomain_2018}\footnote{The performance comes from the original paper \citep{inoue_crossdomain_2018}.}   & 35.7 & 61.9 & 26.2 & 45.9  & 29.9 & 74.0 & 48.7  & 2.8 & 53.0 & 72.7 & 50.2 & 19.3 & 40.9 & 83.3 & 62.4 & 42.4 & 22.8 & 38.5 & 49.3 & 59.5 &  {\color{olive} \textbf{46.0}}$^{\star}$ \\
Yolov2 & with domain  & DT+PL \citep{inoue_crossdomain_2018}\repeatfootnote &  & & &  & & &  & & & &  & & & & & & &  &  &  &  39.9$^{\star}$ \\
Faster RCNN & adaptation & DT+PL \citep{inoue_crossdomain_2018}\repeatfootnote &  & & &  & & &  & & & &  & & & & & & &  &  &  &  34.9$^{\star}$ \\
\hline \hline
 \multirow{3}{*}{VGG16-IM} & Weakly  & WSDDN \citep{bilen_weakly_2016} \repeatfootnote & 1.6 & 3.6 & 0.6 & 2.3 & 0.1 & 11.7 &4.5 & 0.0 & 3.2 & 0.1 & 2.8 &  2.3 & 0.9 & 0.1 & 14.4 & 16.0 & 4.5 & 0.7 & 1.2 & 18.3 & 4.4  \\
  & supervised & SPN \citep{zhu_soft_2017}  & 0.0 & 12.5 & 0.8 & 0.1 & 0.0 & 12.5 & 1.0 & 0.0 & 0.1 & 4.8 & 6.4 & 0.0 & 5.3 & 5.0 & 2.3 & 0.0 & 0.0 & 0.0 & 22.5 & 2.5 & 3.8 \\
  & fine tuning & PCL \citep{tang_pcl_2018} & 0.4 & 0.0 & 0.3 & 1.1 & 0.1 & 0.0 & 5.9 & 0.0 & 0.9 & 0.0 & 0.3 & 3.8 & 0.3 & 0.0 & 3.6 & 1.5 & 0.0 & 0.7 & 0.0 & 4.4 & 1.2 \\
\hline \hline
\multirow{11}{*}{\begin{tabular}{c}RES- \\ 152- \\ COCO\end{tabular}}  & \multirow{11}{*}{\begin{tabular}{c}Features \\ extraction\end{tabular}} & \MAX \citep{crowley_art_2016} & 15.2 & 12.6 & 15.7 & 23.3 & 2.2 & 34.5 & 19.0 & 0.0 & 15.6 & 7.7 & 2.4 & 4.6 & 24.7 & 41.9 & 15.6 & 32.6 & 0.4 & 0.0 & 46.4 & 22.9 & 16.9 \\
& & \MAXA{} & 24.7 & 29.2 & 19.7 & 31.6 & 6.0 & 37.0 & 34.6 & 0.0 & 30.6 & 1.7 & 4.2 & 0.9 & 12.7 & 53.0 & 35.4 & 34.0 & 0.7 & 4.9 & 50.3 & 29.5 & 22.0 \\
\cline{3-24}
& & MI-SVM \citep{andrews_support_2003}  & 10.3 & 35.8 & 8.4 & 22.4 & 15.5 & 25.0 & 28.3 & 8.7 & 26.9 & 4.8 & 14.3 & 0.0 & 18.4 & 45.0 & 22.6 & 16.4 & 1.5 & 7.9 & 51.9 & 22.4 & 19.3 \\  %
& & mi-SVM no GS  \citep{andrews_support_2003} & 1.0 & 4.1 & 8.1 & 6.4 & 1.5 & 4.5 & 16.0 & 4.4 & 10.4 & 4.1 & 2.7 & 0.1 & 10.6 & 20.5 & 6.2 & 3.1 & 0.2 & 2.6 & 8.6 & 8.5 & 6.2 \\ %
& &  MI\_Net \citep{wang_revisiting_2018} & 21.3  & 45.6  & 26.8  & 22.2  & 37.4  & 47.6  & 42.8  & 18.4  & 40.0  & 28.1  & 21.7  & 4.3  & 24.8  & 24.3  & 27.9  & 22.2  & 7.2  & 29.7  & 47.0  & 53.9  & 29.7   $\pm$ 1.5 \\ 
 & & MI\_Net\_with\_DS \citep{wang_revisiting_2018}   & 12.9  & 44.1  & 15.0  & 12.1  & 25.1  & 30.5  & 11.8  & 14.0  & 26.4  & 14.4  & 16.8  & 4.3  & 8.9  & 12.6  & 16.4  & 15.2  & 5.1  & 23.5  & 30.5  & 39.1  & 18.9   $\pm$ 2.4   \\  
& &   MI\_Net\_with\_RC \citep{wang_revisiting_2018} & 1.6  & 2.0  & 0.2  & 0.0  & 0.6  & 0.1  & 3.2  & 0.4  & 0.6  & 0.6  & 0.1  & 0.0  & 0.5  & 0.3  & 2.2  & 1.9  & 0.3  & 0.6  & 2.3  & 0.0  & 0.9   $\pm$ 0.8   \\   
 & & mi\_Net \citep{wang_revisiting_2018} & 20.0  & 43.6  & 28.7  & 23.9  & 36.3  & 50.4  & 43.2  & 20.2  & 43.6  & 34.3  & 25.7  & 3.9  & 22.1  & 25.2  & 30.3  & 9.7  & 5.3  & 28.0  & 41.3  & 55.2  & 29.5   $\pm$ 1.2   \\  
 \cline{3-24}
& & \MILS{} & 42.4  & 46.4  & 25.0  & 45.6  & 45.6  & 52.6  & 43.7  & 24.0  & 45.5  & 42.4  & 29.1  & 5.9  & 35.5  & 52.3  & 55.5  & 50.0  & 2.1  & 15.7  & 60.3  & 47.9  & {\color{red} \bf 38.4  } $\pm$ 0.8  \\
& & \MaxOfMaxS{} &  32.6  & 36.3  & 15.7  & 27.8  & 32.6  & 52.8  & 42.3  & 7.1  & 41.5  & 20.8  & 14.4  & 2.0  & 30.5  & 57.6  & 54.7  & 32.9  & 1.7  & 10.2  & 58.1  & 38.4  & 30.5   $\pm$ 2.3  \\  
& & \mimaxaddlayerS{} & 31.8  & 46.6  & 25.5  & 31.3  & 45.1  & 41.6  & 43.1  & 8.6  & 46.9  & 33.9  & 8.7  & 3.7  & 29.8  & 43.5  & 54.4  & 51.9  & 2.7  & 14.6  & 48.6  & 47.7  & 33.0   $\pm$ 1.2 \\
\hline
\end{tabular}
}
\end{sidewaystable*}
\end{savenotes}

\begin{savenotes}
\begin{table*}[h]
\caption{\textbf{Comic2k (test set)} Average precision (\%). Comparison of the proposed \MILS{} method to alternative approaches. no GS means no Grid Search on the hyperparameters of the SVM otherwise it is the case.}
\label{table:Comic2k}
 \resizebox{\textwidth}{!}{
\begin{tabular}{|c|c|c|cccccc|c|}
\hline
 Net & Method & Model & bike &bird &car &cat &dog &person & mean \\  \hline \hline
\multirow{2}{*}{SSD} & Mixed supervised with  & \multirow{2}{*}{DT+PL} \citep{inoue_crossdomain_2018}\footnote{The performance comes from the original paper \citep{inoue_crossdomain_2018}.}   & \multirow{2}{*}{76.5} & \multirow{2}{*}{54.9} & \multirow{2}{*}{46.0} & \multirow{2}{*}{37.4} & \multirow{2}{*}{38.5} & \multirow{2}{*}{72.3 }&  \multirow{2}{*}{{\color{olive} \textbf{54.3}}$^{\star}$ }\\ 
& domain adaptation & & &  &  &  &  &  &  \\
\hline \hline
 \multirow{3}{*}{VGG16-IM} & Weakly & WSDDN \citep{bilen_weakly_2016} \repeatfootnote & 1.5 & 26.0 & 14.6 & 0.4 & 0.5 & 33.3 & 12.7  \\
& supervised & SPN \citep{zhu_soft_2017}  & 0.0 & 0.0 & 0.0 & 3.1 & 0.0 & 4.1 & 1.2 \\
 & fien tuning & PCL \citep{tang_pcl_2018} & 0.0 & 0.0 & 0.0 & 0.0 & 0.0 & 0.0 & 0.0 \\
\hline \hline
\multirow{11}{*}{\begin{tabular}{c}RES- \\ 152- \\ COCO\end{tabular}}  & \multirow{11}{*}{\begin{tabular}{c}Features \\ extraction\end{tabular}} & \MAX \citep{crowley_art_2016} & 15.2 & 2.7 & 29.4 & 2.3 & 16.8 & 4.9 & 11.9 \\ 
& & \MAXA{}  & 36.8 & 5.6 & 27.1 & 8.2 & 6.1 & 34.8 & 19.8 \\ 
\cline{3-10}
& & MI-SVM \citep{andrews_support_2003}  & 34.2 & 3.0 & 20.0 & 5.2 & 2.5 & 12.9 & 13.0 \\  %
& & mi-SVM no GS  \citep{andrews_support_2003} & 10.8 & 2.3 & 5.5 & 3.2 & 2.1 & 3.6 & 4.6 \\ %
& &  MI\_Net \citep{wang_revisiting_2018} & 42.9  & 15.5  & 33.1  & 11.8  & 13.4  & 20.4  & 22.8   $\pm$ 1.1    \\  
 & & MI\_Net\_with\_DS \citep{wang_revisiting_2018}  & 40.8  & 13.3  & 32.5  & 5.7  & 9.1  & 16.1  & 19.6   $\pm$ 1.6   \\  
& &   MI\_Net\_with\_RC \citep{wang_revisiting_2018}&  19.8  & 5.4  & 16.4  & 2.8  & 9.8  & 13.9  & 11.4   $\pm$ 4.4   \\   
 & & mi\_Net \citep{wang_revisiting_2018} & 42.1  & 10.9  & 24.5  & 8.8  & 8.8  & 22.1  & 19.5   $\pm$ 2.1   \\
 \cline{3-10}
& & \MILS{} & 45.3  & 9.7  & 33.7  & 14.4  & 21.6  & 37.0  & {\color{red} \bf 27.0 }  $\pm$ 0.8  \\ 
& &  \MaxOfMaxS{} & 44.9  & 5.2  & 26.2  & 14.1  & 11.0  & 38.4  & 23.3   $\pm$ 1.6    \\ 
& & \mimaxaddlayerS{} & 43.0  & 5.1  & 31.5  & 11.8  & 13.8  & 36.4  & 23.6   $\pm$ 0.5   \\
\hline
\end{tabular}
}
\end{table*}
\end{savenotes}

\begin{savenotes}
\begin{table*}[h]
\caption{\textbf{CASPA paintings (test set)} Average precision (\%). Comparison of the proposed \MILS{} method to alternative approaches. no GS means no Grid Search on the hyperparameters of the SVM otherwise it is the case.}
\label{table:CASPApaintings}
  \resizebox{\textwidth}{!}{
\begin{tabular}{|c|c|c|cccccccc|c|}
\hline
 Net & Method & Model & bear  & bird & cat  & cow & dog & elephant & horse & sheep & mean \\  \hline \hline
 \multirow{2}{*}{VGG16-IM} & Weakly supervised & SPN \citep{zhu_soft_2017}  & 0.5 & 0.1 & 1.6 & 0.9 & 0.5 & 1.4 & 0.6 & 0.0 & 0.7 \\
  & fine tuning & PCL \citep{tang_pcl_2018}
 & 0.0 & 0.0 & 0.0 & 0.0 & 0.0 & 0.0 & 0.0 & 0.0 & 0.0 \\ \hline
\hline
\multirow{11}{*}{\begin{tabular}{c}RES- \\ 152- \\ COCO\end{tabular}}  & \multirow{11}{*}{\begin{tabular}{c}Features \\ extraction\end{tabular}} & \MAX \citep{crowley_art_2016} & 22.0 & 2.1 & 14.5 & 3.5 & 14.2 & 8.8 & 12.8 & 0.5 & 9.8 \\
 & & \MAXA{} & 26.3 & 13.1 & 26.9 & 5.4 & 8.3 & 18.1 & 14.9 & 3.9 & 14.6 \\ 
 \cline{3-12}
 & & MI-SVM \citep{andrews_support_2003}  & 9.3 & 0.2 & 6.7 & 1.5 & 0.1 & 0.6 & 0.9 & 0.4 & 2.5 \\  %
 & & mi-SVM no GS  \citep{andrews_support_2003} & 1.3 & 1.6 & 3.0 & 0.8 & 1.0 & 0.3 & 1.5 & 0.3 & 1.2 \\ %
 & &  MI\_Net \citep{wang_revisiting_2018} & 32.8  & 5.4  & 14.1  & 5.2  & 6.2  & 15.0  & 11.1  & 4.2  & 11.7   $\pm$ 1.6   \\    
  & & MI\_Net\_with\_DS \citep{wang_revisiting_2018}  & 29.0  & 1.6  & 8.3  & 3.0  & 3.2  & 5.9  & 7.1  & 2.6  & 7.6   $\pm$ 1.2   \\    
 & &   MI\_Net\_with\_RC \citep{wang_revisiting_2018}  & 16.9  & 0.9  & 6.6  & 2.6  & 2.9  & 8.2  & 4.7  & 2.1  & 5.6   $\pm$ 2.1   \\  
  & & mi\_Net \citep{wang_revisiting_2018} & 26.7  & 8.9  & 12.5  & 1.5  & 3.4  & 7.1  & 5.1  & 2.4  & 8.4   $\pm$ 1.7   \\  
  \cline{3-12}
 & & \MILS{} & 28.3  & 15.7  & 25.6  & 5.3  & 13.7  & 17.2  & 18.8  & 5.1  & {\color{red} \bf 16.2  } $\pm$ 0.4   \\ 
 & &  \MaxOfMaxS{}  &  26.2  & 16.9  & 23.9  & 5.4  & 10.1  & 9.7  & 18.8  & 4.5  & 14.4   $\pm$ 0.7  \\   
 & & \mimaxaddlayerS{}  & 26.5  & 15.7  & 26.3  & 4.8  & 14.2  & 10.1  & 11.5  & 6.2  & 14.4   $\pm$ 0.9   \\
 \hline
\end{tabular}
}
\end{table*}
\end{savenotes}

\begin{savenotes}
\begin{table*}[h]
\caption{\textbf{\NB{} detection test set} \textit{detection} average precision (\%) at  IoU $\geqslant$0.5. Comparison of the proposed \MILS{}, \MaxOfMaxS{} and \mimodels{} methods to alternative approaches. In those case, we use a grid search for \MAX{} and \MAXA{}. In red, the best weakly supervised method. %
}
\label{table:IconArt_PCAtest_IuO05}
 \resizebox{\textwidth}{!}{
\begin{tabular}{|c|c|c|ccccccc|c|} 
\hline
  Net & Method & Model & angel & JCchild & crucifixion & Mary & nudity & ruins & StSeb & mean \\ 
\hline   \hline
\multirow{2}{*}{VGG16-IM} &  Weakly supervised   & SPN \citep{zhu_soft_2017} 
&  0.0 & 0.8 & 22.3 & 12.0 & 6.8 & 10.4 & 1.2 & 7.7 \\
 & fien tuning & PCL\footnote{Trained with the following hyperparameters: batch size = 2, learning rate = 0.001, epochs = 13 and number of clusters by default.} \citep{tang_pcl_2018} &  2.9 & 0.3 & 1.0 & 26.3 & 2.3 & 7.2 & 1.4 & 5.9 \\
   \hline \hline
\multirow{11}{*}{\begin{tabular}{c}RES- \\ 152- \\ COCO\end{tabular}}  & \multirow{11}{*}{\begin{tabular}{c}Features \\ extraction\end{tabular}} & \MAX  \citep{crowley_art_2016} & 1.4 & 1.3 & 11.5 & 2.8 & 3.8 & 0.3 & 4.5 & 3.7 \\
& & \MAXA{}  & 1.3 & 4.4 & 18.2 & 28.0 & 15.3 & 0.2 & 16.4 & 12.0 \\
\cline{3-11}
 & &  MI-SVM \citep{andrews_support_2003}  & 0.7 & 4.4 & 21.6  &0.6  & 1.0 & 0.0 & 0.0  & 4.0 \\ %
 &  & mi-SVM \citep{andrews_support_2003}  & 1.3 & 5.1 & 3.9 & 3.6 & 2.9 & 0.3 & 2.2  & 2.8 \\ %
& &  MI\_Net \citep{wang_revisiting_2018} & 9.7   & 42.6   & 21.1 & 6.9 & 17.6  & 5.1  & 2.5   & {\color{red} \bf 15.1  } $\pm$ 1.5   \\ 
& & MI\_Net\_with\_DS \citep{wang_revisiting_2018} & 8.6   & 35.6   & 19.6   & 5.3   & 15.9   & 3.2   & 3.1   & 13.0   $\pm$ 1.7   \\ 
& &   MI\_Net\_with\_RC \citep{wang_revisiting_2018} & 8.2  & 36.9   & 20.5   & 4.8  & 16.2   & 1.6   & 0.9  & 12.7   $\pm$ 1.6   \\  
& & mi\_Net \citep{wang_revisiting_2018} & 8.2   & 28.4   & 15.1   & 11.2   & 15.8   & 6.8  & 4.5   & 12.9   $\pm$ 1.2   \\ 
\cline{3-11}
& & \MILS{}   & 0.3   & 0.1  & 42.7   & 4.4 & 21.9   & 0.6  & 13.7   & 12.0   $\pm$ 0.9   \\ 
& & \MaxOfMaxS{} & 3.1   & 9.8   & 33.0  & 7.4   & 29.2   & 0.1 & 8.5   &  13.0  $\pm$ 2.2   \\  
& & \mimaxaddlayerS{}  & 4.3  & 6.7  & 35.7  & 15.6  & 24.0  & 0.1  & 15.2  &  14.5  $\pm$ 1.8   \\
\hline
\end{tabular}
}
\end{table*}
\end{savenotes}

As explained above, we concentrate on non-photographic databases for which a ground truth is available for object detection on the test set. We report in Tables \cref{tab:People-Art_detection,table:Watercolor2k_detection,table:Clipart1k_detection,table:Comic2k,table:CASPApaintings,table:IconArt_PCAtest_IuO05}  the performances for the weakly supervised object detection task for 6 different non-photographic datasets: PeopleArt \citep{westlake_detecting_2016}, Watercolor2k, Clipart1k, Comic2k \citep{inoue_crossdomain_2018}, IconArt \citep{gonthier_weakly_2018} and CASPApaintings \citep{thomas_artistic_2018}.
CASPApaintings is the paintings subset of the CASPA dataset\footnote{\url{http://people.cs.pitt.edu/~chris/artistic_objects/}} proposed in \citep{thomas_artistic_2018} with bounding boxes associated to 8 visual categories (only animals) for most of the images. 

When the method is not too costly we provide standard deviation and mean score computed on 10 runs of it. %

First, we can see that for all databases, the end-to-end weakly  supervised methods (WSDDN, SPN and PCL) yield relatively poor results. Possible explanations are that the model overfits on the training set or that the model is stuck in bad local minima, so that the weakly supervised setting is not adequate with a relatively small training dataset.
Moreover in the case of PCL, the boxes are proposed by the Selective Search algorithm \citep{uijlings_selective_2013} which, as shown in Table \ref{tab:Recall_boxes}, completely fails on the considered non-photographic datasets. That alone can explain the poor results of PCL on those datasets. Recall also that these methods do use features inherited from systems such as FasterCNN that are pretrained with bounding box annotations.

When comparing the performances of the different multiple instance neural networks, we can see that MI\_Net (Maximum Bag Margin Formulation) outperforms the other MIL networks on three datasets. Moreover the multiple instance neural network outperforms the multiple instance SVM (mi-SVM and MI-SVM), which can be due to the fact that a linear SVM that are not complex enough. 

We can notice that the Maximum Pattern margin methods (mi-SVM and mi\_Net) never perform better that the Bag margin ones. This is rather unexpected since those models are designed to better take into account the whole positive bag by assigning an individual label per instance. These models appear to be badly suited for the task of weakly supervised detection in non-photographic databases.

When comparing our \MILS{} and \MaxOfMaxS{} models to the baseline \MAX{} and \MAXA{}, we observe that our models consistently perform better. Nevertheless the \MAXA{} model performs well especially on the \NB{} or CASPApaintings databases, probably because this model uses all the regions of the negatives images, yielding good discrimination of background regions during inference. The \MAX{} baseline sometimes provides equivalent performances to more complex methods (such as MI-SVM or MI\_Net), illustrating the fact that the objectness score (used for selecting candidates in \MAX{}) contains useful information.
Also observe that it is faster to train a multiple instance perceptron than several linear SVMs, as is needed for MI-SVM or mi-SVM. This is quantified in Section \ref{sec:execution_time}.

Finally, we observe that both our models \MILS{} and \MaxOfMaxS{} provides better results than the others methods on PeopleArt, CASPApaintings, Comic2k, Clipart1k and Watercolor2k datasets.

The dataset \NB{} appear to be much more challenging. In this case, our multiple instance methods provide equivalent performances compared to the multiple instance networks. The best performance is obtained by the MI\_Net, the \mimaxaddlayerS{} performance being very similar. 

\subsubsection{Execution Time}
\label{sec:execution_time}

One advantage of our method is the relativel short time needed for training, as can be seen in Table \ref{tab:ExecutionTime_boxes}.
As can be expected, the SPN and PCL methods are the longest to train due to the fine-tuning of the whole network.
Observe also that the traingin time for our method \MILS{} is almost independent of the number of classes and restarts, which is a strong advantage compared to the MI-SVM, mi-SVM, MI\_Net and mi\_Net models which all need one full training per class and per re-initialization. The SVM based methods are more costly because they don't take advantage of GPU computational power.

Nevertheless, due to the aggregation of several hyperplan with a maximum operator in the \MaxOfMaxS{} model, we need to do 10 time more epochs that when using \MILS{}, which explain the strong overload.

\begin{table*}[h]
\centering
\caption{Execution time of the different models for datasets Watercolor2k and Comic2k, with 1000 images in the training set and 6 visual categories.} 
\label{tab:ExecutionTime_boxes}
  \resizebox{\textwidth}{!}{
\begin{tabular}{|c|c|c|c|}
\hline
Method & Training Duration  & Linear to number of class & Linear to number of restarts \\
\hline 
No Boxes proposals & & & \\
SPN \citep{zhu_soft_2017} & 3000s (20 epochs) & No & $\bullet$  \\ %
\hline
Selective Search Bounding Boxes proposal& 6600s &  &  \\
PCL  \citep{tang_pcl_2018} & 12000s (13 epochs) & No & $\bullet$  \\ %
\hline
Faster RCNN Features and boxes proposals &  200s &  &  \\
 \MAX{}   & 52s & Yes & $\bullet$  \\ %
 \MAXA{}   & 2000s & Yes & $\bullet$  \\ %
MI-SVM \citep{andrews_support_2003} & 3000s & Yes & Yes \\
mi-SVM \citep{andrews_support_2003} &  30000s  & Yes & Yes\\
MI\_Net \citep{wang_revisiting_2018} & 1200s (20 epochs)& Yes & Yes \\ %
MI\_Net\_with\_DS \citep{wang_revisiting_2018} & 1800s (20 epochs)& Yes & Yes \\
MI\_Net\_with\_RC \citep{wang_revisiting_2018} &  1600s (20 epochs)& Yes & Yes \\
mi\_Net \citep{wang_revisiting_2018} & 1800s (20 epochs) & Yes & Yes \\
\MILS{} & 130s (300 epochs) & No & No \\  %
\MaxOfMaxS{} & 1100s (3000 epochs) & No & No \\
\mimaxaddlayerS{}& 3000s (300 epochs) & No & Yes \\
\hline
\end{tabular}}
\end{table*}

\subsection{Fine \MILS{} models Analysis}

 In this section we discuss the details of our models and some variations. In particular, we provide an ablation study where we analyze how the choices of a different loss, different set of features and use of  the objectness  score impact the performances of our models. In Section \ref{sec:Param} a thorough investigation of the main parameters' influence is conducted. From this study we are able to recommend a set of parameters that are suited for our models, thus providing the user with a safe baseline for re-using them. Then, we experimentally show that our method also permits to transfer easily the knowledge between datasets and artistic modalities. In section \ref{sec:KnowledgeTransfer_crossMod}, we also evaluate the generalization ability of our models across different modalities of images (using classes shared by the different datasets).  Finally, in section \ref{sec:visualResults} some visual results are commented to give an insight on the strengths and shortcomings of our model. 

\subsubsection{Ablation study}
\label{sec:ablationStudy}
 {\bf Choice of the loss function:}
In Table \ref{tab:Score_and_hingeLoss_Study}, we gather different versions of the two models \MILS{} and \MaxOfMaxS{} with two possible modifications. First we replace the $\mathop{Tanh}$ based loss in equation \eqref{eqn:perexampleloss} by the Hinge loss. Second we suppress the objectness score in the loss function (see section \ref{sec:objectTrick}).

The first conclusion that can be drawn is that the use of objectness score significantly increase the performances of our models. This is especially true for the PeopleArt dataset where the performances very srongly decrease without using the objectness score. For the other datasets the performances are always significantly lower without the objectness score. Note that for some classes this drop in detection score is due to the fact that the model detects parts of the object instead of the whole object when the objectness score is ignored. Such an example can be seen in figure \ref{fig:Bowes_without_and_withScore} section \ref{sec:visualResults}, where the class for Saint Sebastian is confused with arrows, which is understandable in this case but not desirable. The use of the objectness score often helps avoiding such partial detection cases. 

The second conclusion is that replacing the $\mathop{Tanh}$ based loss function in equation \eqref{eqn:perexampleloss} by a Hinge loss $l(y,\hat{y})=1-\max(0,1-y\hat y)$ generally hinders the performances, except for two cases among the 12 cases of the (dataset,model) possible combinations. In particular the \MaxOfMaxS{} methods never benefits from a different loss function. This may be due to the fact that, given the difficulty of the task, errors are likely to happen and the $\mathop{Tanh}$ function may be more robust and forgiving than the Hing loss which will try hard to correct any errors, especially those with a high negative margin.

\begin{savenotes}
\begin{table*}[h]
\centering
\caption{Mean average precision over the classes of the different datasets (\%). Comparison of the proposed \MILS{} and \MaxOfMaxS{} methods with different settings. Standard deviation is computed on 10 runs of the method.} %
\label{tab:Score_and_hingeLoss_Study}
 \resizebox{\textwidth}{!}{
\begin{tabular}{|c||c|c|c|c||c|c|c|c|}
\hline
\multirow{3}{*}{Dataset} &  \multicolumn{4}{c||}{\MILS{}} & \multicolumn{4}{c|}{\MaxOfMaxS{}} \\ 
 & \multirow{2}{*}{Main Model} & \multirow{2}{*}{Without score} & \multirow{2}{*}{Hinge loss} & Without score & \multirow{2}{*}{Main Model} & \multirow{2}{*}{Without score} & \multirow{2}{*}{Hinge loss}  & Without score \\
 & & & & and hinge loss & & & & and hinge loss \\
\hline
PeopleArt & 55.5 $\pm$ 1.0 & 0.9  $\pm$ 0.4 & 57.6  $\pm$ 1.0 & 1.7  $\pm$ 0.9 & 58.3  $\pm$ 1.2 & 10.1  $\pm$ 3.3 & 56.6  $\pm$ 4.4 & 18.1  $\pm$ 8.6 \\
Watercolor2k & 49.5   $\pm$ 0.9  & 32.8   $\pm$ 2.2 & 46.7   $\pm$ 1.5  & 33.8   $\pm$ 1.6 & 46.6   $\pm$ 1.3 & 18.3   $\pm$ 4.7 & 37.5   $\pm$ 2.1  & 24.8   $\pm$ 3.3 \\
Clipart1k & 38.4   $\pm$ 0.8  & 24.2   $\pm$ 1.6 & 34.8   $\pm$ 1.2  & 22.2   $\pm$ 1.8  & 30.5   $\pm$ 2.3  & 11.9   $\pm$ 2.6  &  16.5   $\pm$ 1.2  & 5.1   $\pm$ 1.1   \\
Comic2k & 27.0   $\pm$ 0.8 & 17.4   $\pm$ 1.5 & 25.5   $\pm$ 1.1  &  17.3   $\pm$ 1.1   & 23.3   $\pm$ 1.6  & 11.6   $\pm$ 2.8   & 15.0   $\pm$ 1.8  &  9.5   $\pm$ 1.8 \\
CASPA paintings &  16.2   $\pm$ 0.4 &  18.7   $\pm$ 0.8  & 16.1   $\pm$ 0.5& 12.6   $\pm$ 0.9   & 14.4   $\pm$ 0.7  &  8.6   $\pm$ 1.4   & 9.0   $\pm$ 0.9  & 3.2   $\pm$ 0.6  \\
IconArt & 12.0   $\pm$ 0.9  & 6.7   $\pm$ 2.5  & 14.3   $\pm$ 2.1  & 8.2   $\pm$ 2.3  & 13.0   $\pm$ 2.2 & 6.4   $\pm$ 2.3   & 13.3  $\pm$ 2.8 &   8.3   $\pm$ 2.0  \\
\hline
\end{tabular}
}
 \end{table*}  
 \end{savenotes}

{\bf Features extraction and region proposals choices:}
We have investigated alternative choices for the Faster RCNN's features and box proposals: for the boxes we used the unsupervised box proposal algorithm EdgeBoxes \citep{zitnick_edge_2014} and for the features we used a ResNet-152 trained on ImageNet applied to each proposed box. By doing so we must drop the objectness score that is not included in the output of EdgeBoxes. 

We can see in Table \ref{tab:EdgeBoxes} the performances of the model \MILS{} (without the objectness score) using  those features/boxes compared to the Faster RCNN features/boxes (without objectness score for fair comparison).
Regarding the detection task the performances clearly drop when using EdgeBoxes.
To further investigate this drop of performance we present in Table \ref{tab:Recall_boxes}  the recall score of three box proposals methods (the percentage of ground-truth boxes that are present in the set of all proposed boxes). We can see that EdgeBoxes performs very poorly on a data-set like PeopleArt and never matchs the boxes proposed by Faster RCNN.

For the classification task  we can see that the \MILS{} method without objectness score performs honorably in this setting when compared to the use of Faster RCNN's  boxes/features (even slightly better on  the IconArt database). This is another proof that bag-level classification (the aim of the training of a MIL algorithm) is not a good proxy for instance-level classification (which is the aim of a detection algorithm). The objectness score can be seen as a very helpful cue to guide the training of a WSOD method. As shown by \cite{donahue_decaf_2013} for classification task, transfer learning of deep models trained for detection tasks is the best way to obtain a detector on new domains even when no  bounding boxes are available.

\begin{table*}[h]
\caption{Average precision for detection and classification (\%). Two different feature extraction methods are considered in this table (both without objectness score).}
\label{tab:EdgeBoxes}
\center
\begin{tabular}{|c|c|c|c|}
\hline
Dataset & Metric & Faster RCNN & EdgeBoxes \\ \hline \hline 
\multirow{2}{*}{PeopleArt} & AP IuO $\geqslant$0.5 & 0.9  $\pm$ 0.4 & 0.0  $\pm$ 0.0    \\ 
 & Classif AP & 92.5  $\pm$ 0.3 & 92.1  $\pm$ 0.2 \\
  \hline
  \multirow{2}{*}{Clipart1k} & AP IuO $\geqslant$0.5 &  24.2   $\pm$ 1.6 & 3.1   $\pm$ 0.3  \\
 & Classif AP & 59.4   $\pm$ 1.7 & 42.8   $\pm$ 1.3  \\
  \hline 
  \multirow{2}{*}{Comic2k} & AP IuO $\geqslant$0.5 &  17.4   $\pm$ 1.5  & 1.8   $\pm$ 0.3   \\
 & Classif AP & 54.9   $\pm$ 2.0 & 47.9   $\pm$ 1.5    \\ \hline
\multirow{2}{*}{Watercolor2k} & AP IuO $\geqslant$0.5 &32.8   $\pm$ 2.2 & 2.7   $\pm$ 0.5 \\
 & Classif AP & 78.0   $\pm$ 1.2 & 71.8   $\pm$ 1.3   \\
 \hline
 \multirow{2}{*}{CASPA} & AP IuO $\geqslant$0.5 & 12.6   $\pm$ 0.5 & 0.3   $\pm$ 0.1 \\
 & Classif AP &  48.6   $\pm$ 0.6 &  45.0   $\pm$ 1.2    \\
 \hline
\multirow{2}{*}{IconArt} &  AP IuO $\geqslant$0.5 & 6.7   $\pm$ 2.5   & 5.3   $\pm$ 0.3 \\  
 & Classif AP & 60.4   $\pm$ 1.1 & 69.2   $\pm$ 0.3 \\
 \hline
\end{tabular}
 \end{table*}

\begin{table*}[h]
\centering
\caption{Recall (\%) at IuO $\geqslant$0.5 of the boxes proposals for the different methods and databases. Mean over the classes.} %
\label{tab:Recall_boxes}
\begin{tabular}{|c|c|c|c|}
\hline
\multirow{2}{*}{Dataset} & RPN of Pre-trained  & EdgeBoxes &
Selective Search \\
& Faster RCNN \citep{ren_faster_2015} &\citep{zitnick_edge_2014}  & \citep{uijlings_selective_2013} \\ 
\hline
Number of boxes & 300 & 300 & 3000-5000 \\
\hline
PeopleArt & 94.0 & 15.4 & 55.7 \\
Clipart1k & 91.4 & 14.4 & 49.4 \\
Comic2k & 82.7 & 54.1 & 46.2 \\
Watercolor2k & 93.6 & 61.4 & 56.8 \\
CASPA & 76.6 & 34.3 & 51.6\\
IconArt & 75.9 & 60.0 & 56.9 \\
\hline
\end{tabular}
\end{table*}  

\subsubsection{Influence of the parameters of the model}
\label{sec:Param} 

\begin{figure*}[h!]
\centering
  \hfill
     \includegraphics[width=0.32\textwidth]{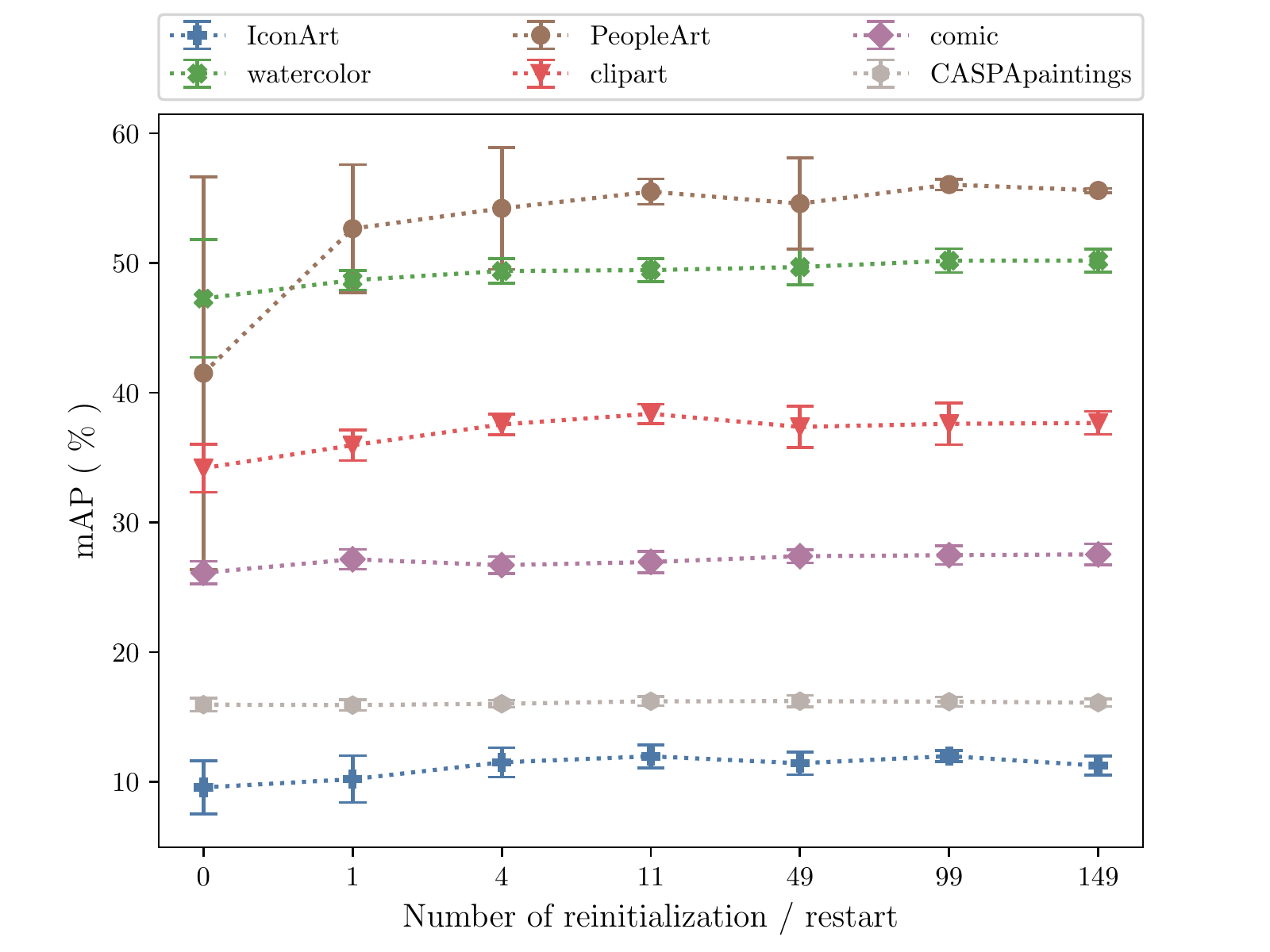}\hfill  
     \includegraphics[width=0.32\textwidth]{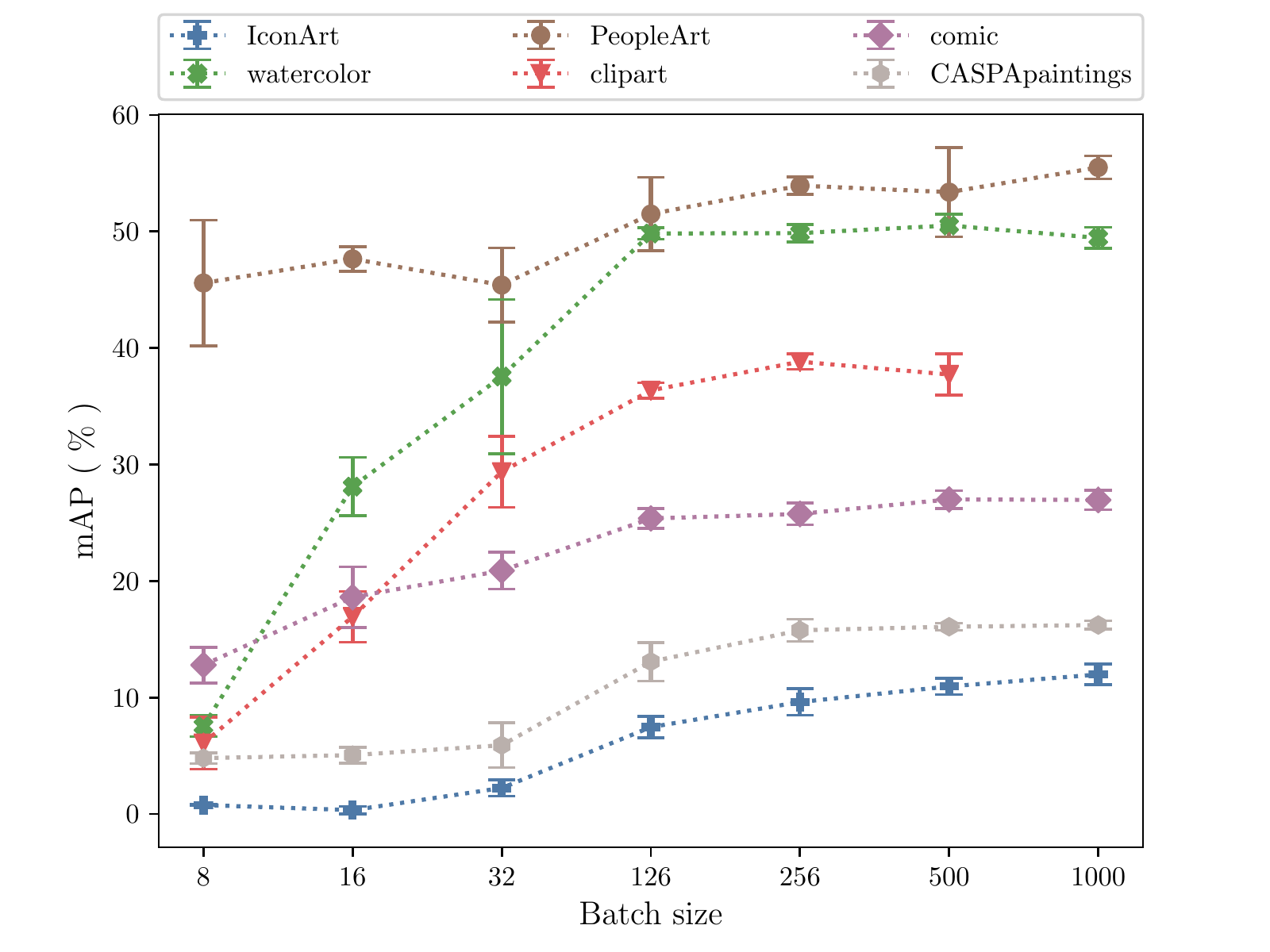}\hfill
     \includegraphics[width=0.32\textwidth]{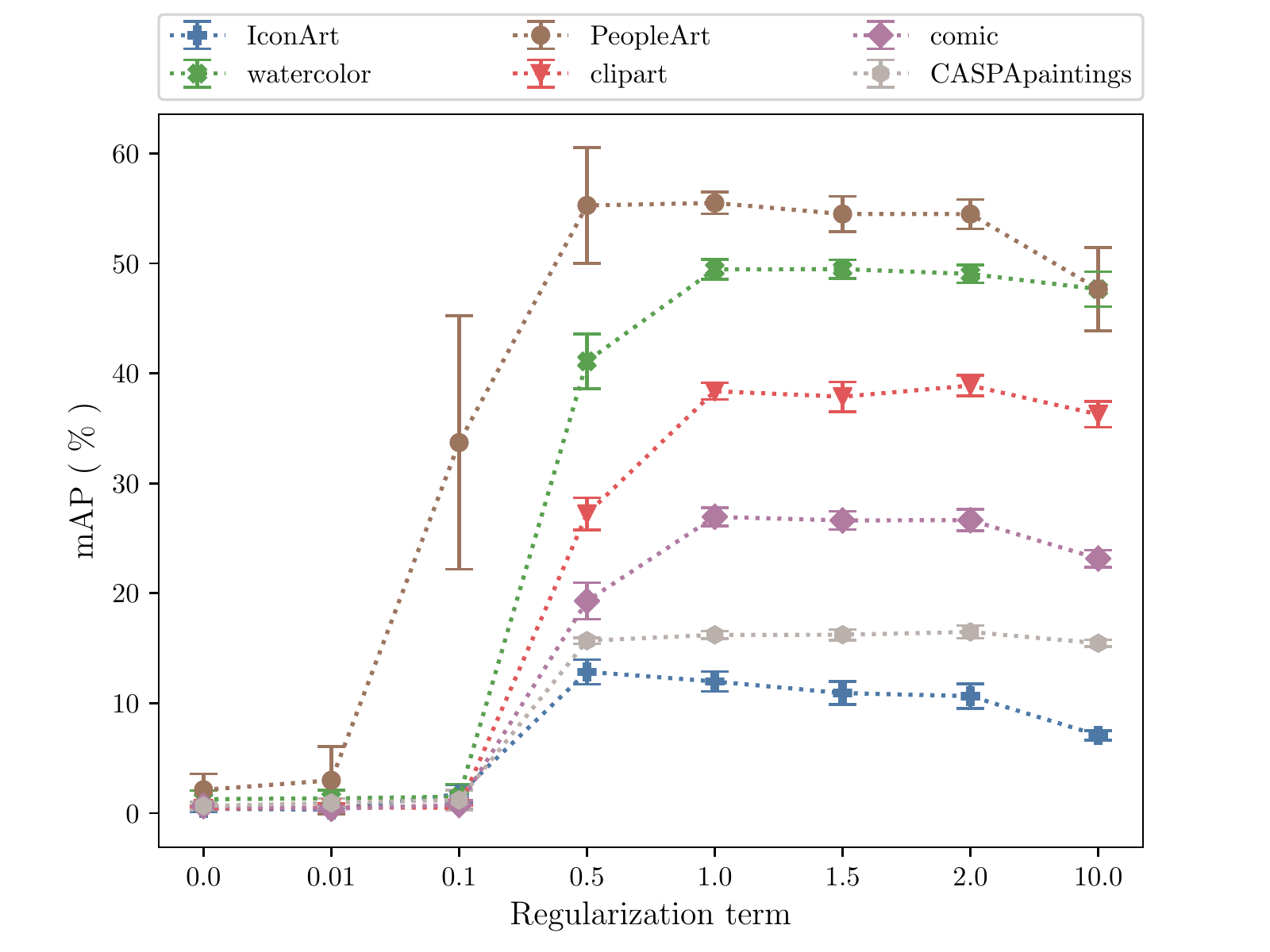}\hspace*{\fill} 
    \caption{Impact of the different hyperparameters on the \MILS{} model. Figure must be seen in color.}
    \label{fig:MIMAX_hyperParam}
\end{figure*}

In this section, we analyse the influence of the different hyperparameters of our \MILS{} model. We show in Figure \ref{fig:MIMAX_hyperParam} the performances with respect to each of the three following parameters: the number of restarts, the batch size and the regularization term $C$. We vary one parameter at a time while keeping the others fixed to the already mentioned values (i.e. 11 for the number of restarts, 1000 for the batch size and 1.0 for $C$).

Although the study in \citep{doran_theoretical_2014} shows that restarts from random points is not always useful for nonconvex models, we find that having about 10 restarts slightly improves the performances and can be taken as a rule of thumb for our models. Notice that the variance of the outcomes is also reduced for such a parameter choice. 
We also found experimentally that restarts for mi-SVM or MI-SVM reduce the performance in accordance with the experiments in \citep{doran_theoretical_2014}. Then, we observe that increasing the batch size provides better results and often yields a reduction of the variance. 
For the regularization term, we observe relatively constant performances between 1.0 and 2.0. The value 0.5 seems to be the best for 2 of the datasets (PeopleArt and IconArt, but with a great variance). These experiments also show the necessity of using a regularization term in the loss function.

\subsubsection{Cross modalities Knowledge Transfer}
\label{sec:KnowledgeTransfer_crossMod}

Tables \cref{tab:PolyhedralMILS_KLtransfer,tab:MILS_KLtransfer} present across-domain performance for two our models \MaxOfMaxS{} and \MILS{}. We compute the performances of detection for the classes that are shared between the different datasets. Those performances (one run) are compared to the mean performance on the same modality (several runs as before).
This experiment illustrates the fact that our method can be transferred to other modality of images. This is sometimes called the "Cross-Depiction Problem" \citep{hall_crossdepiction_2015}: recognizing visual objects regardless of whether they are painted or depicted in different artistic style.

First, we can see that the \MaxOfMaxS{} model trained on PeopleArt outperforms the one learned on the target modality for 2 of the 3 datasets (first line). This can be due to the fact the PeopleArt dataset contains many different artistic style. We also observe that the \MILS{} model badly fails on those three datasets and that the \MaxOfMaxS{} model generalizes better. Observe also that the fact that the class person is well detected can also be due to the Faster RCNN features that have been trained on a dataset (MS COCO) containing this class.

Finally, we can notice that some datasets such as CASPApaintings and Clipart1k are more challenging that the other maybe due to the difference in the modality for the second one.

This experiment illustrates the fact that our model \MaxOfMaxS{} generalize well but also that providing a diverse and numerous training set can help to get a better detector trained in a weakly supervised manner.

\begin{table*}[h]
\centering
\caption{Mean AP (\%) at IuO $\geqslant$0.5 for the common classes between the source and target sets with the \MILS{} model. In parenthesis the mean performance obtained by learning the detection on the same set (modality).} %
\label{tab:MILS_KLtransfer}
\begin{tabular}{|l||*{5}{c|}}\hline
\backslashbox{source set}{target set}
&PeopleArt & Watercolor2k &Comic2k  
& Clipart1k & CASPApaintings\\\hline\hline
PeopleArt & - & 0.0 (58.2) & 0.0 (37.0)  & 0.0 (55.5) & / \\\hline
Watercolor2k &  47.4  (55.5)  & -  & 25.8 (27.0) & 12.2 (33.4)  & 15.6 (18.3) \\\hline
Comic2k & 50.4 (55.5) & 47.3 (49.5)  & - &  10.0 (33.4) & 15.0 (18.3) \\\hline
Clipart1k &  36.2  (55.5) & 44.3  (49.5)  &  25.2  (27.0) & -  & 10.8 (14.0) \\\hline
CASPApaintings & / & 33.4 (35.4)  & 12.2 (15.2) & 4.7 (22.5) & - \\\hline
\end{tabular}
\end{table*} %

\begin{table*}[h]
\centering
\caption{Mean AP (\%) at IuO $\geqslant$0.5 for the common classes between the source and target sets with the \MaxOfMaxS{} model. The mean performance obtained by learning the detection on the same set (modality) is displayed between brackets.} %
\label{tab:PolyhedralMILS_KLtransfer}
\begin{tabular}{|l||*{5}{c|}}\hline
\backslashbox{source set}{target set}
&PeopleArt & Watercolor2k &Comic2k  
& Clipart1k & CASPApaintings\\\hline\hline
PeopleArt & - & 60.0 (59.2) & 42.1 (39.5)  & 54.3 (55.4) & / \\\hline
Watercolor2k & 56.0 (57.3)  & -  & 23.1 (24.1) &  11.2 (24.6) & 13.8 (18.3)\\\hline
Comic2k & 48.9 (57.3) & 42.4 (46.6) & - & 7.2 (24.6)  & 12.5 (18.3) \\\hline
Clipart1k & 52.0 (57.3)  & 36.7 (46.6) & 19.6 (24.1)  & -  & 7.7 (13.6) \\\hline
CASPApaintings& / & 27.5 (39.0) & 9.9 (18.1) & 4.2 (12.5) & - \\\hline
\end{tabular}
\end{table*}

\subsubsection{Visual results from the \MaxOfMaxS{} model.}
\label{sec:visualResults}

In order to give some intuitive insight on the ability of the proposed method, we show some visual illustrations of the performance of the proposed model \MaxOfMaxS{}, both in successful and failure cases. 

\begin{figure}
\centering
\setlength\tabcolsep{1pt}
\renewcommand{\arraystretch}{0.5}
\begin{tabular}{cc}
 \includegraphics[height=\heightimageCASPA]{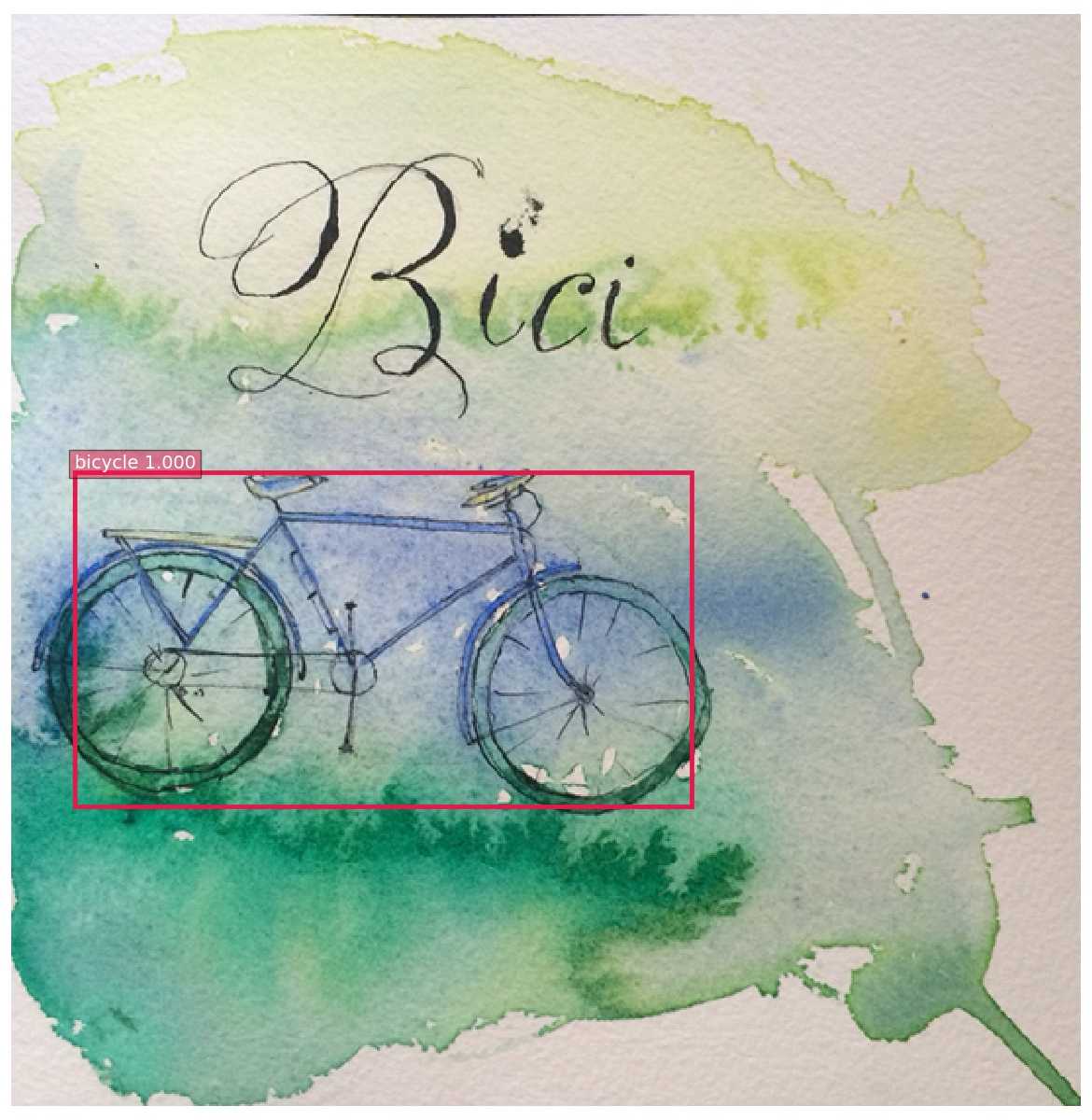}& 
     \includegraphics[height=\heightimageCASPA]{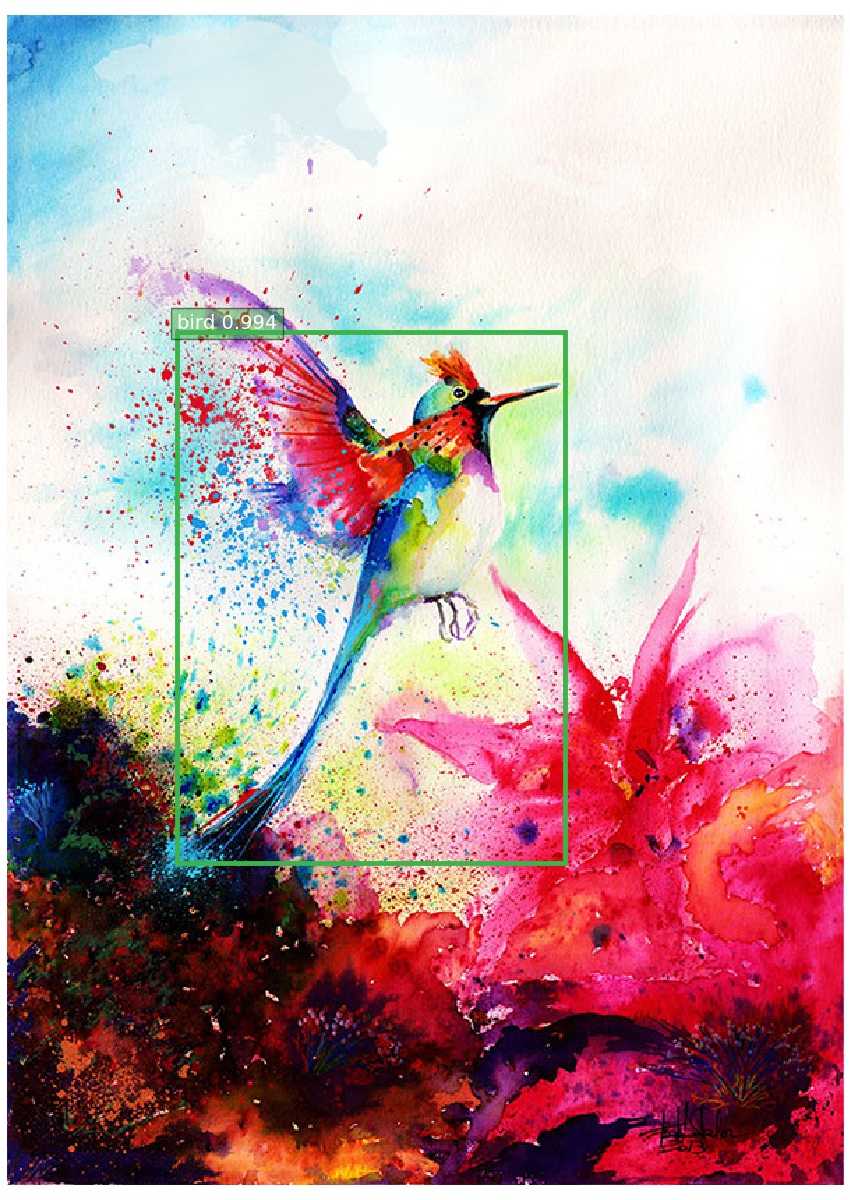}\\
          {\color{red} \footnotesize{Bike 1.0}} &    {\color{darkpastelgreen} \footnotesize{Bird 0.994} } \\
     \includegraphics[height=\heightimageCASPA]{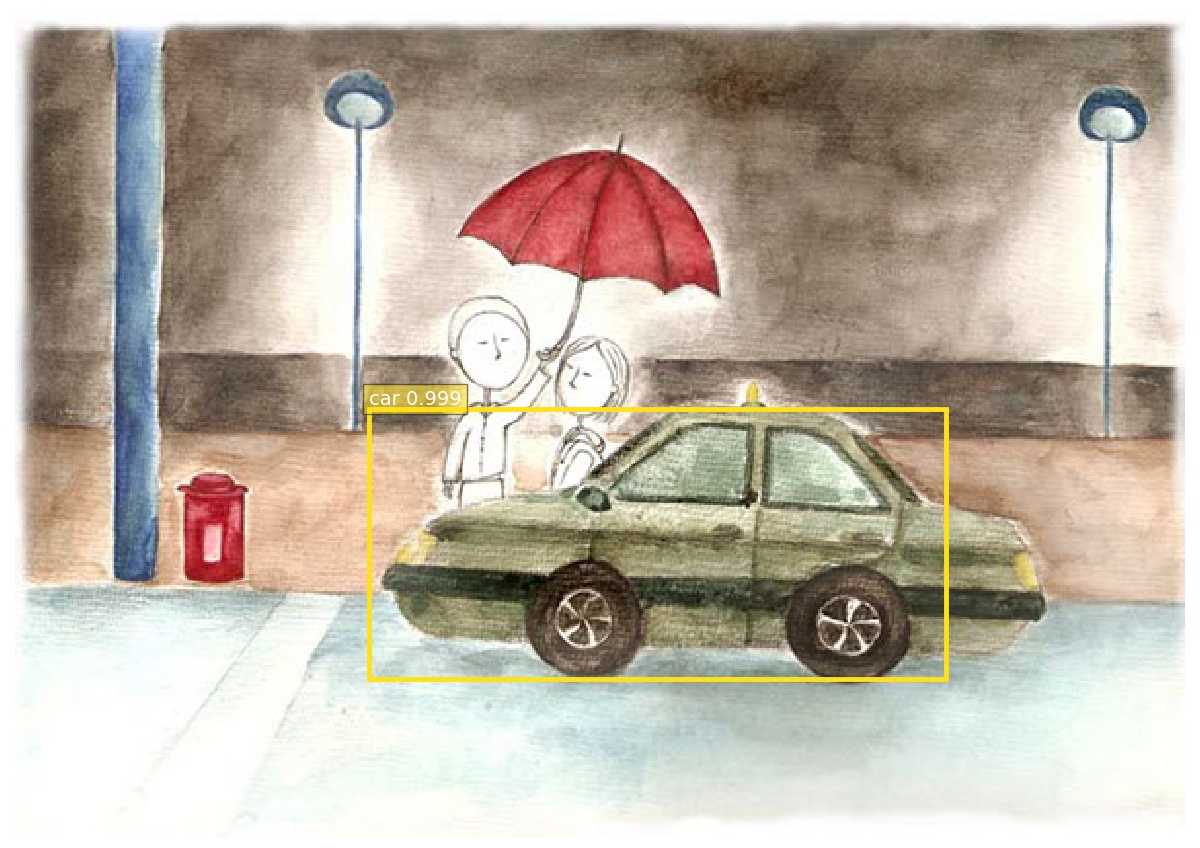}&
     \includegraphics[height=\heightimageCASPA]{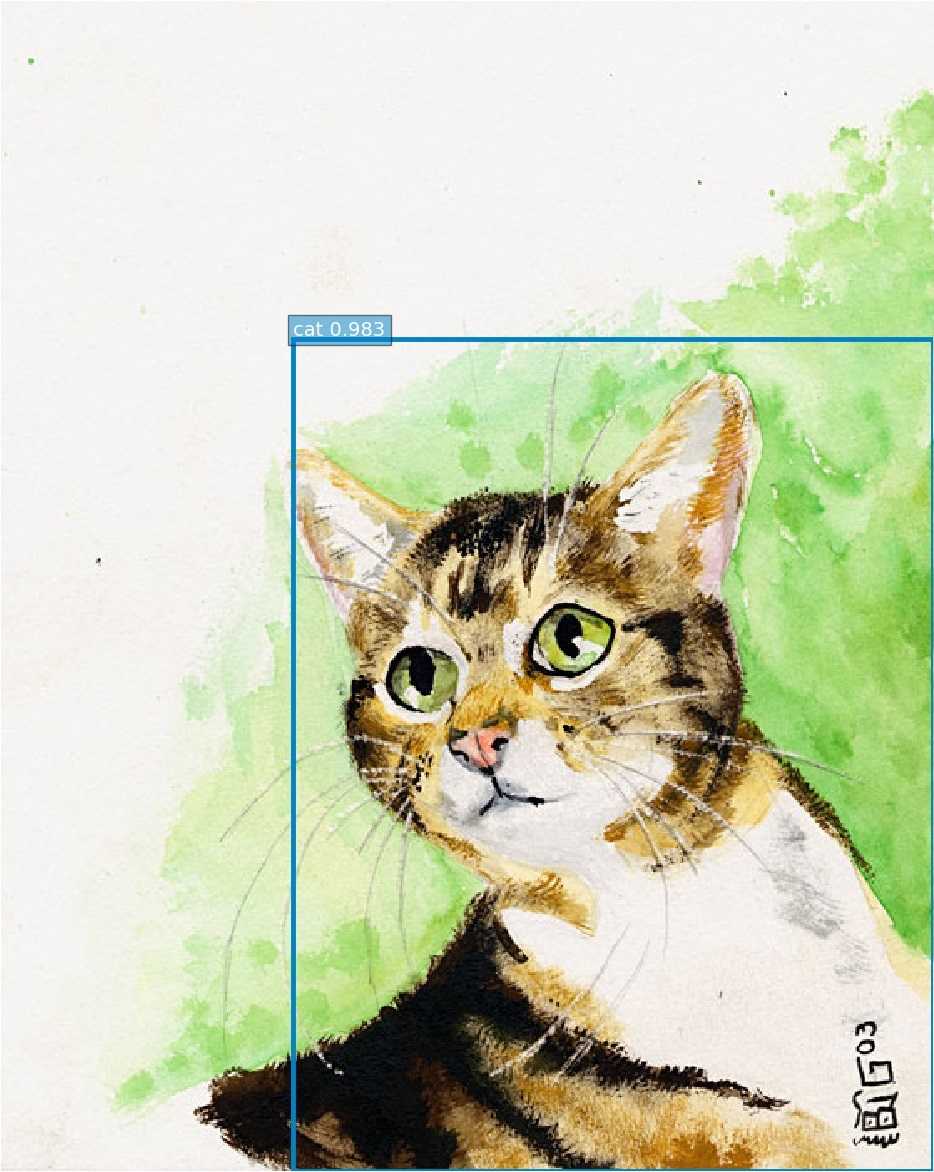}\\
        {\color{electricyellow} \footnotesize{Car 0.999} } & {\color{trueblue} \footnotesize{Cat 0.983} } \\
     \includegraphics[height=\heightimageCASPA]{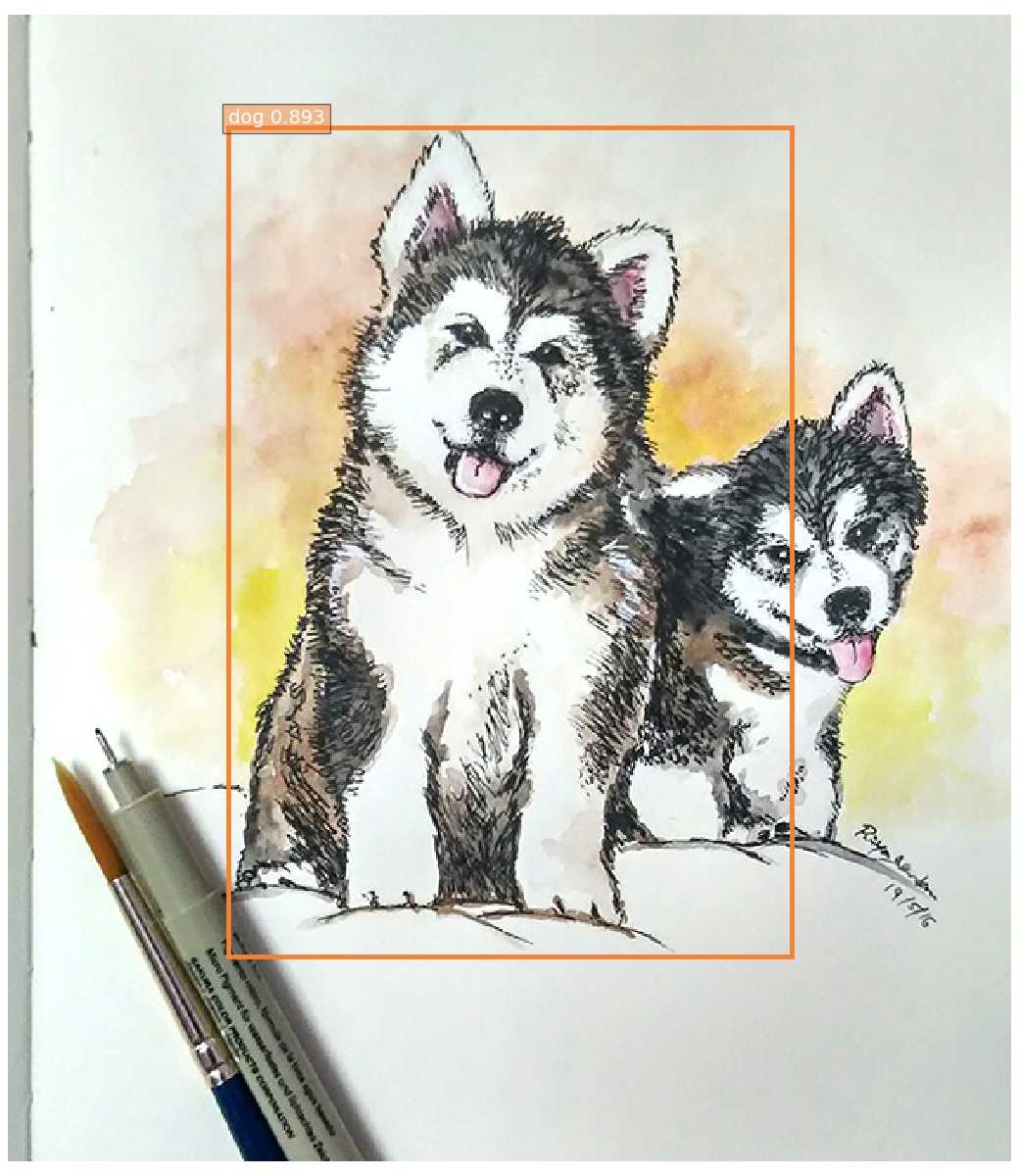}&
     \includegraphics[height=\heightimageCASPA]{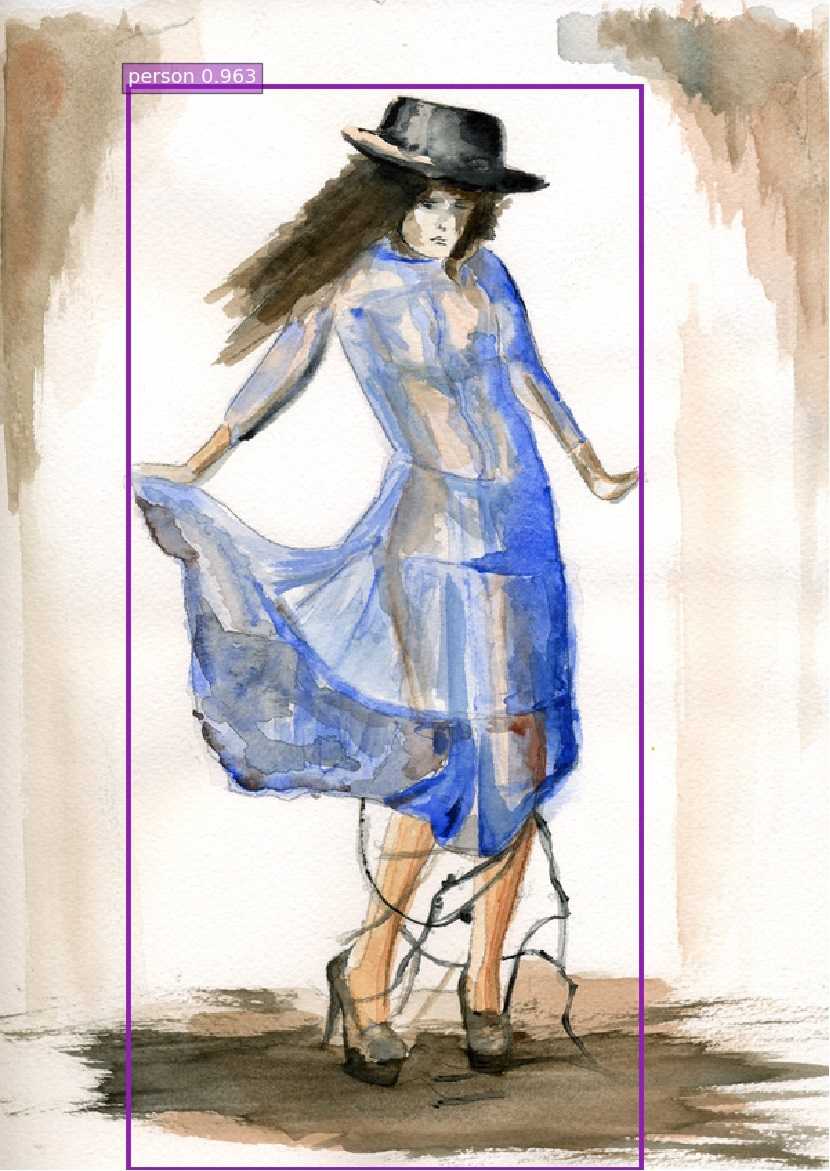}\\
        {\color{carrotorange} \footnotesize{Dog 0.893}} & {\color{purpleheart} \footnotesize{Person 0.963}} \\
\end{tabular}
    \caption{One successful example per class using our \MaxOfMaxS{} detection scheme on Watercolor2k test set. We only show boxes whose scores are over 0.75. Figure must be seen in color.}
    \label{fig:WatercolorSuccessfulDetection}
\end{figure}

\begin{figure}
\centering
\setlength\tabcolsep{1pt}
\renewcommand{\arraystretch}{0.5}
\begin{tabular}{cc}
      \includegraphics[height=\heightimageCASPA]{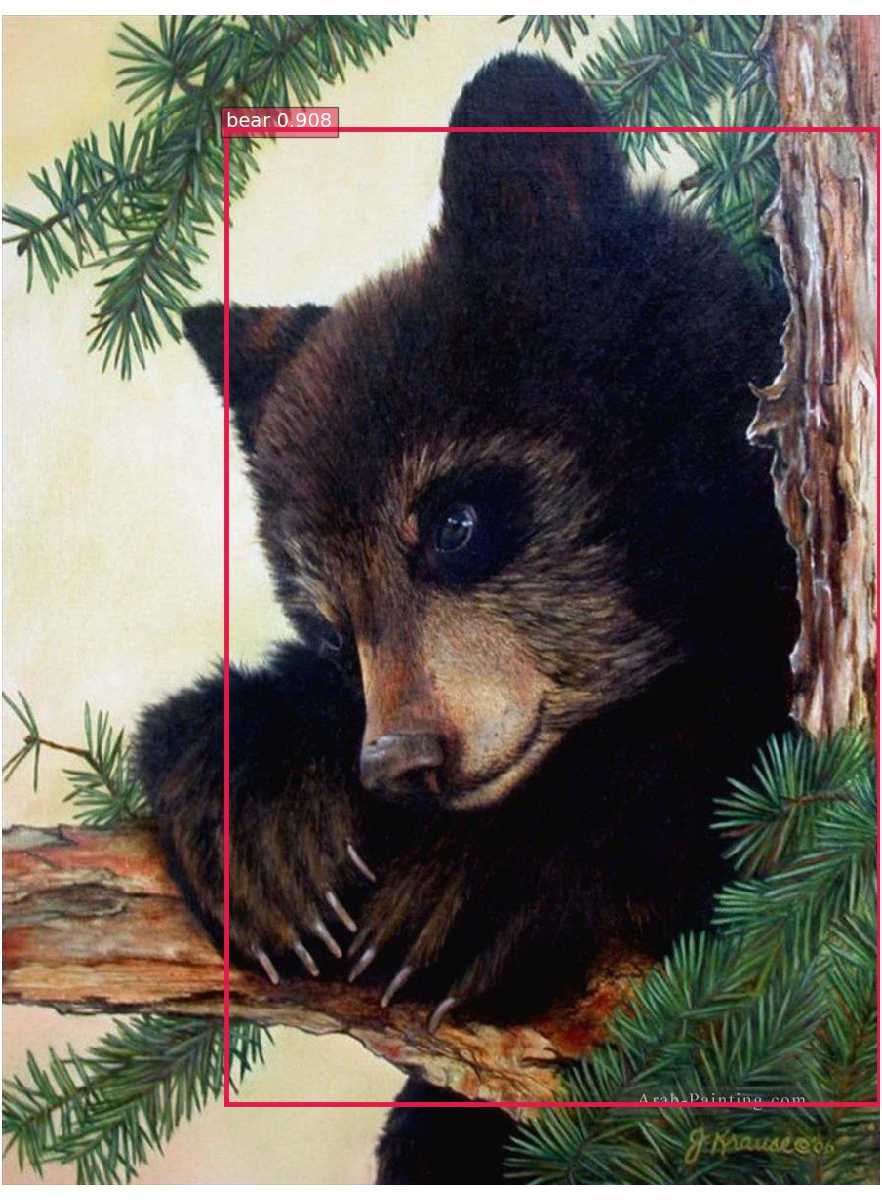}&  
     \includegraphics[height=\heightimageCASPA]{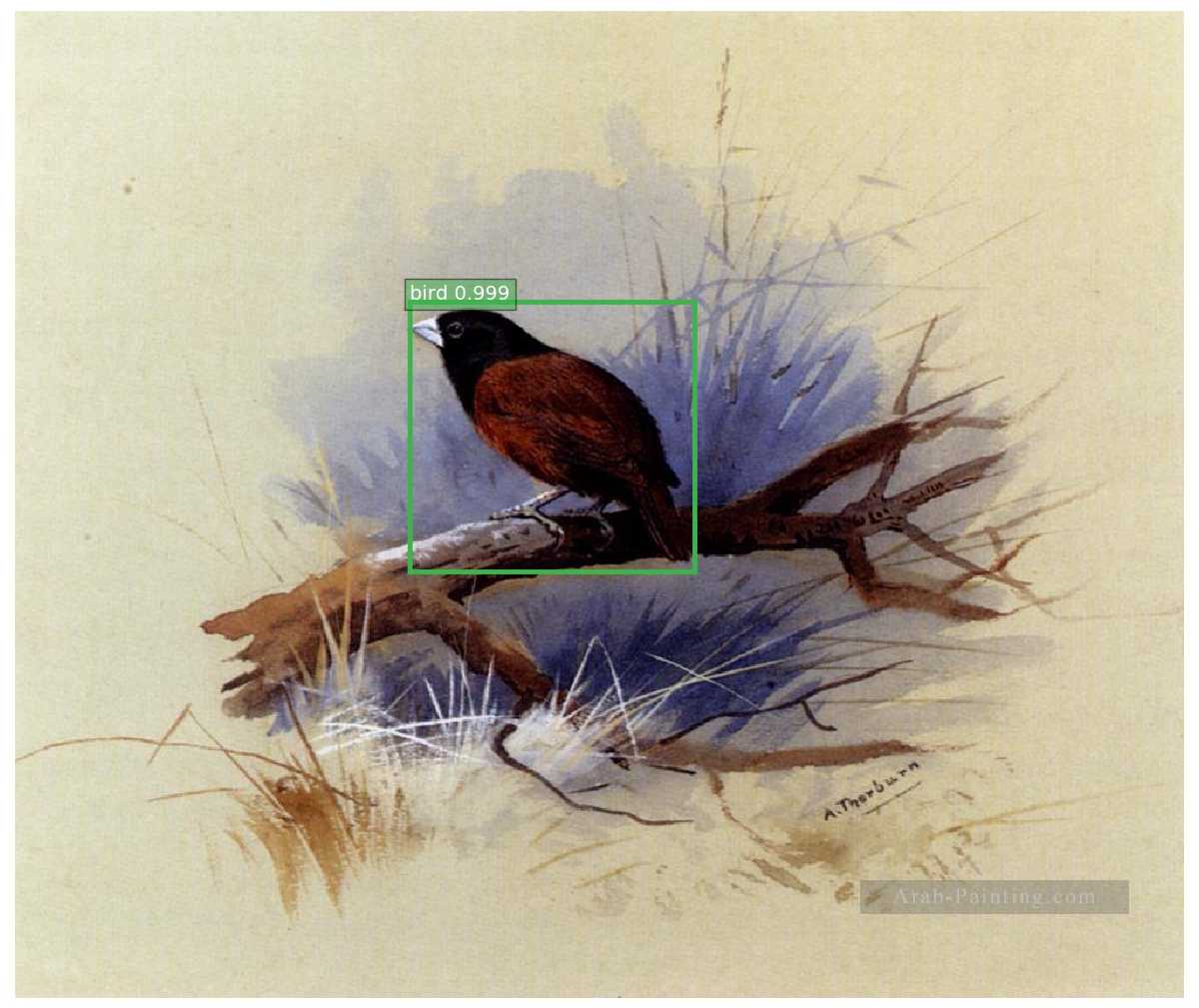} \\
        {\color{red} \footnotesize{Bear 0.908}} &
   {\color{darkpastelgreen} \footnotesize{Bird 0.999} } \\
      \includegraphics[height=\heightimageCASPA]{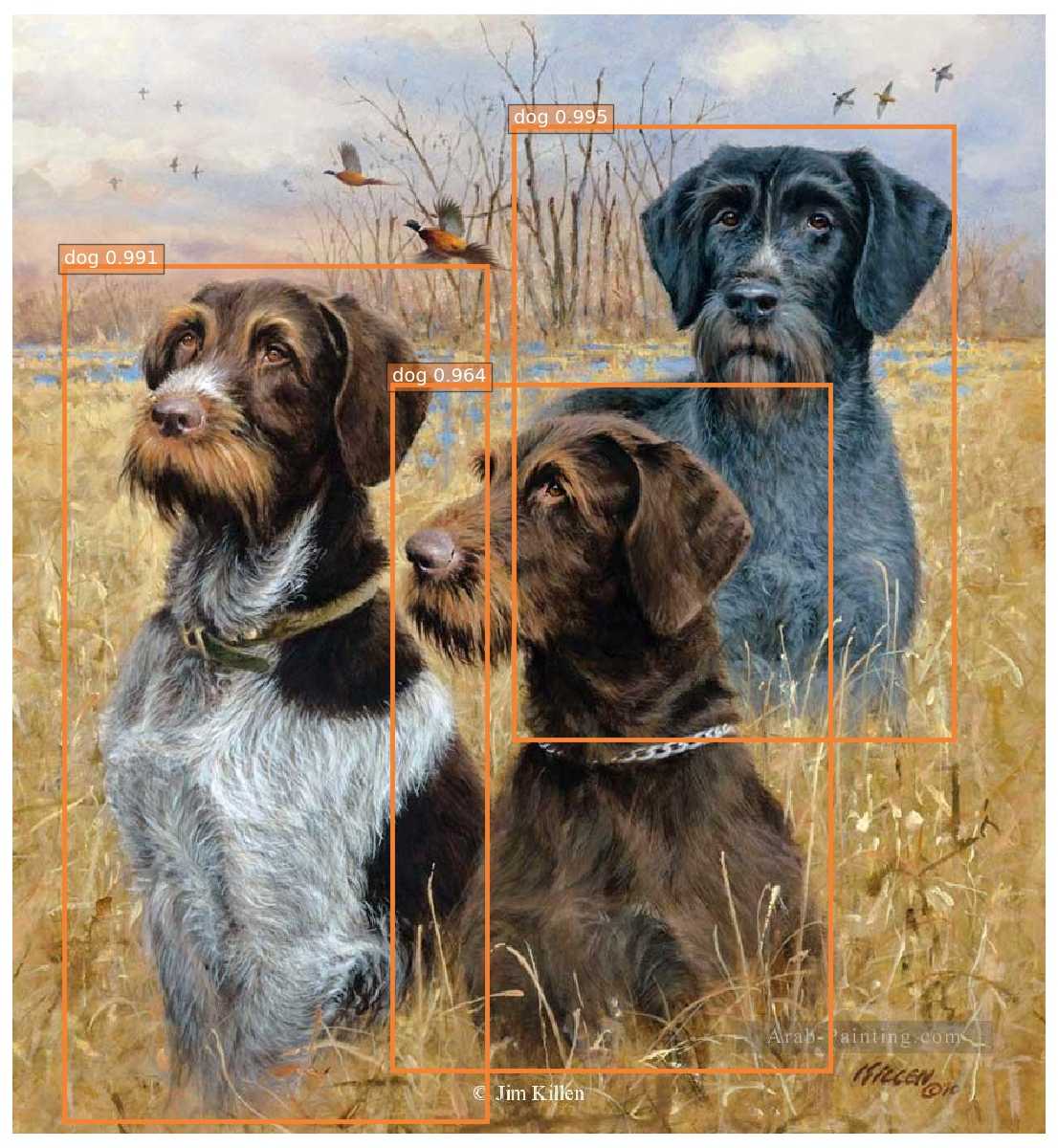}  &
     \includegraphics[height=\heightimageCASPA]{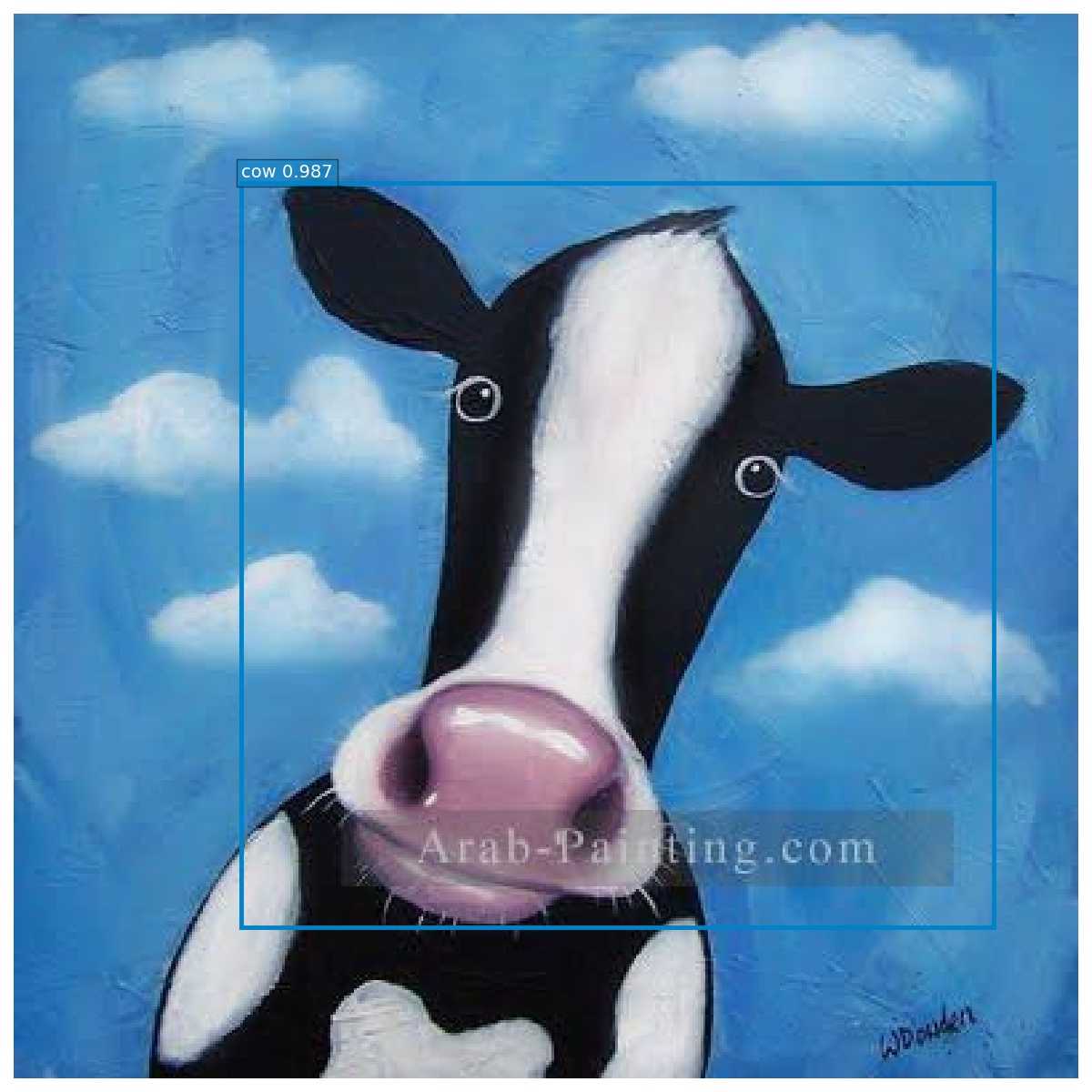} \\ 
     {\color{carrotorange} \footnotesize{Dog 0.995 0.991 0.964}}& {\color{trueblue} \footnotesize{Cow 0.987} } \\
      \includegraphics[height=\heightimageCASPAzeroSept]{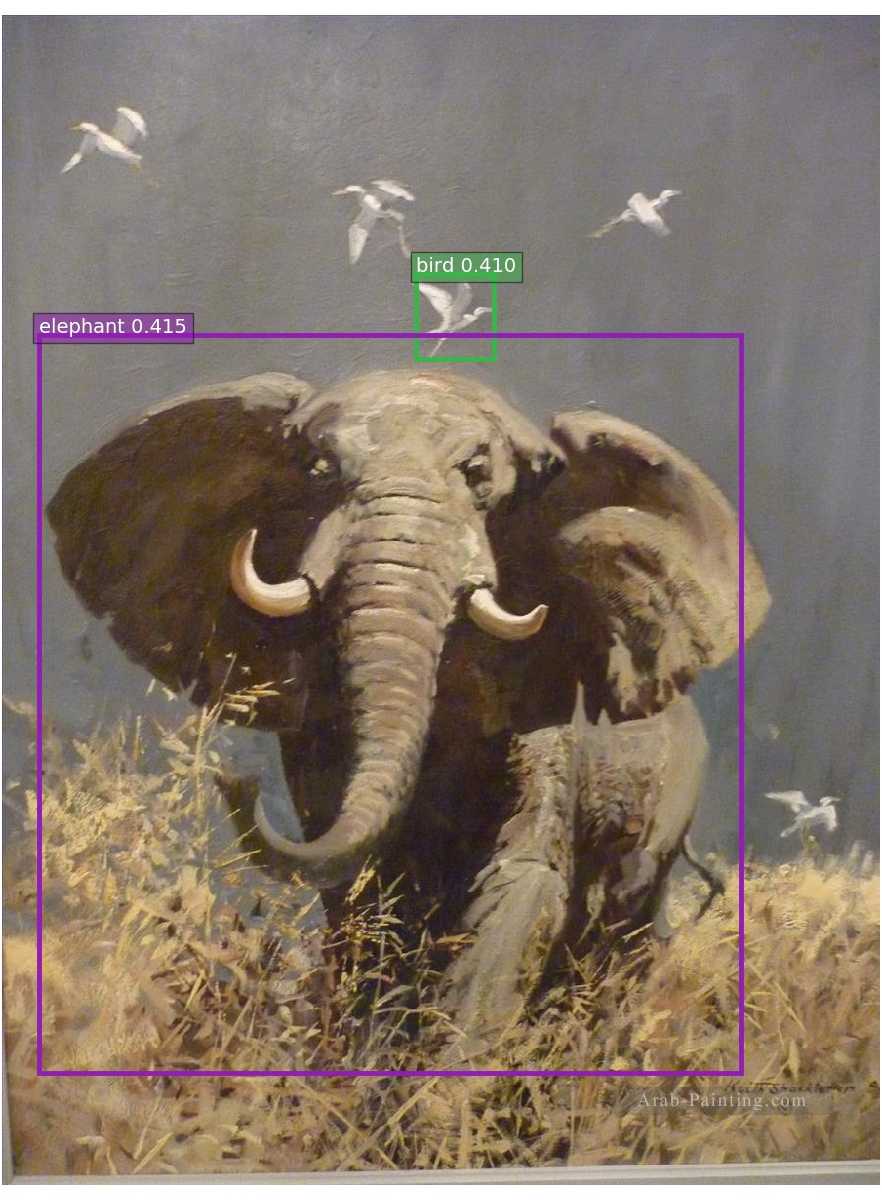} & \includegraphics[height=\heightimageCASPAzeroSept]{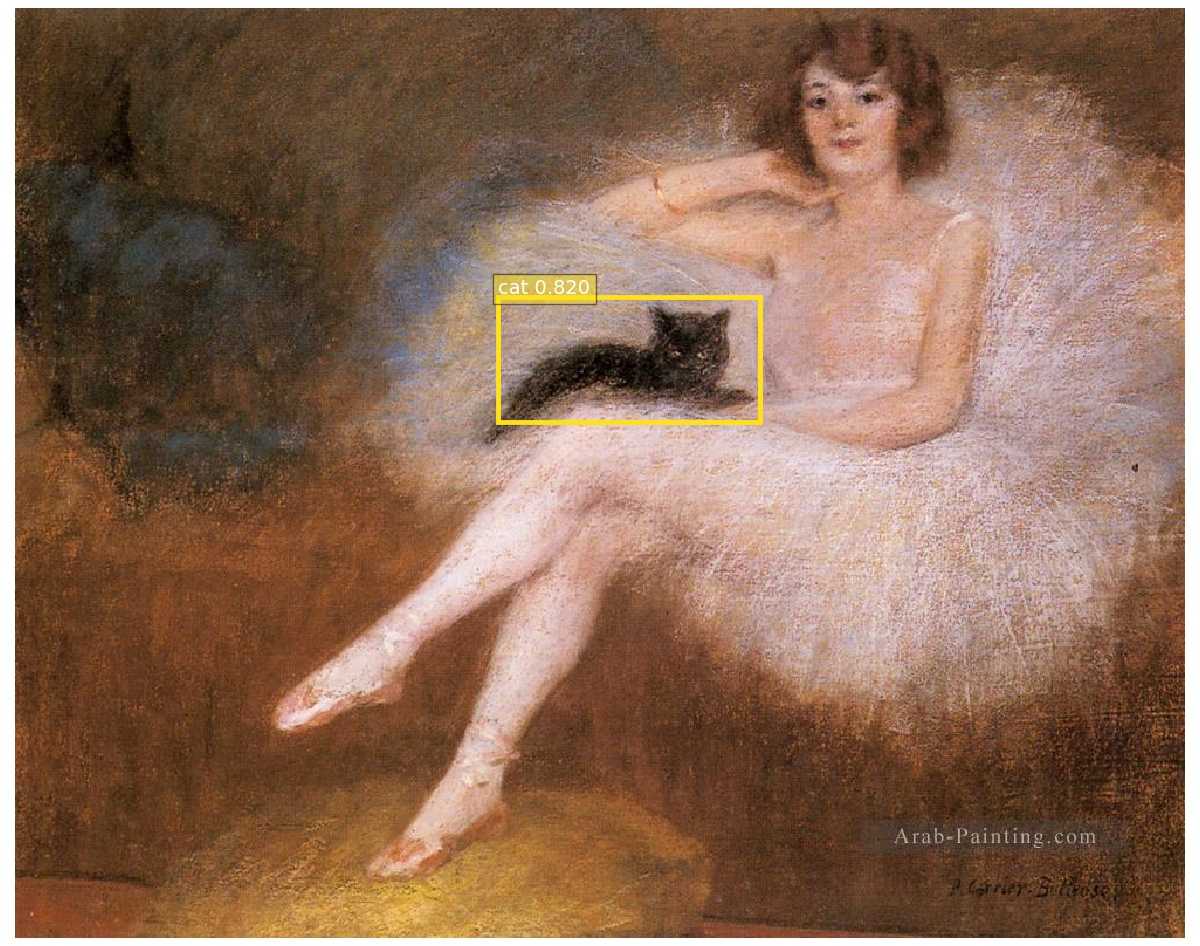} \\ 
     {\color{purpleheart} \footnotesize{Elephant 0.415}}  {\color{darkpastelgreen} \footnotesize{Bird 0.410}} & {\color{electricyellow} \footnotesize{Cat 0.820} }\\
     \includegraphics[height=\heightimageCASPA]{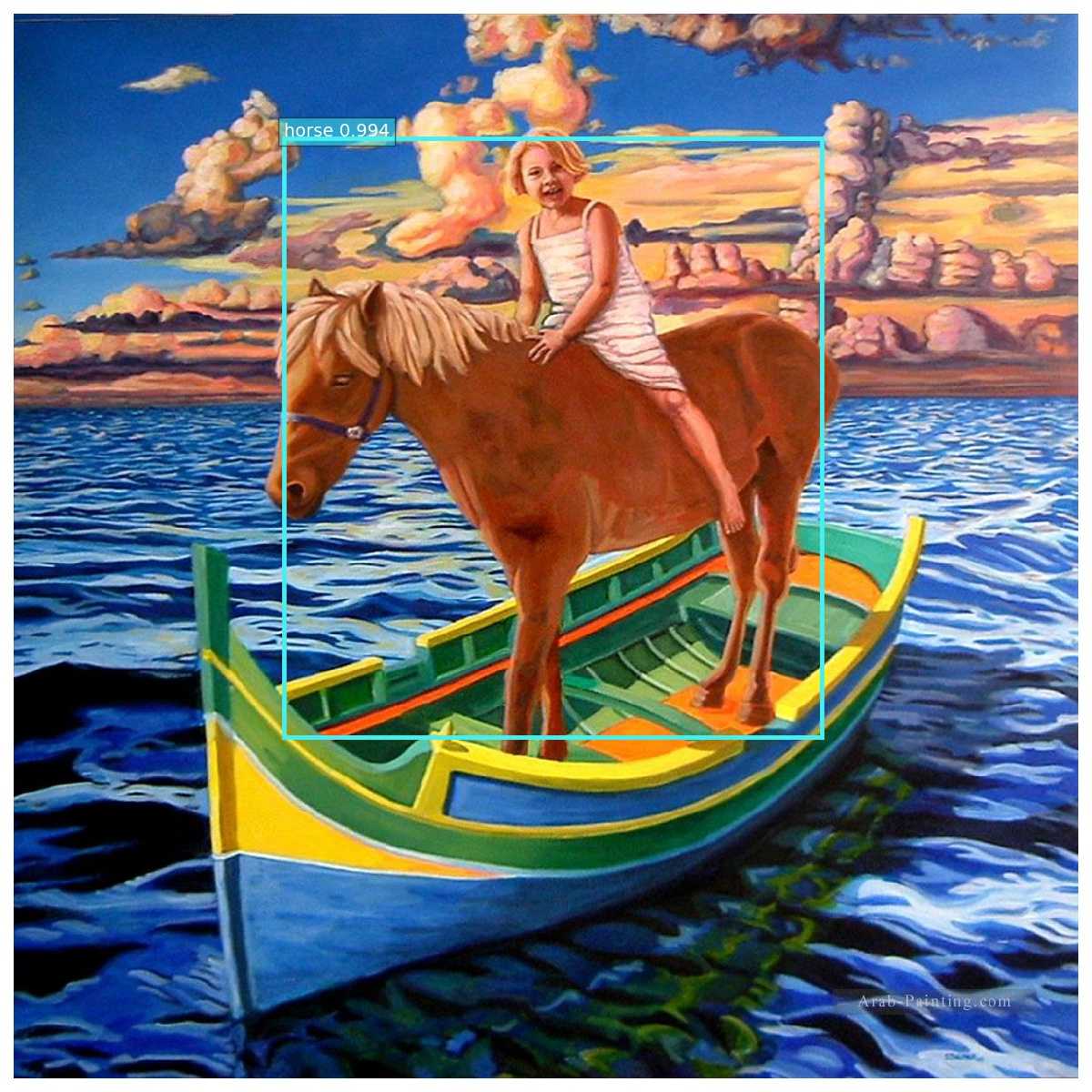} &
          \includegraphics[height=\heightimageCASPA]{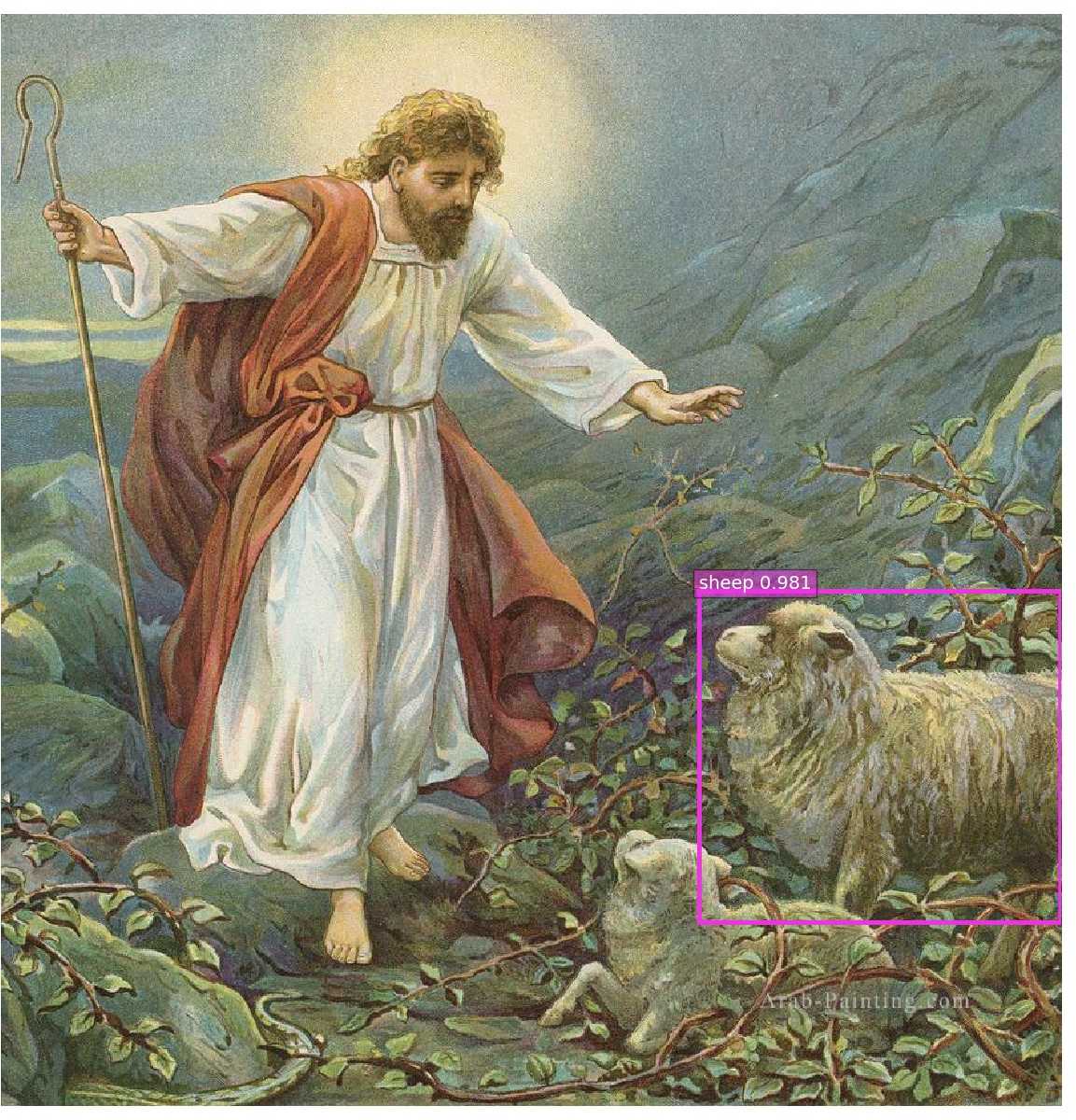}  \\
           {\color{cyan} \footnotesize{Horse 0.994}} & {\color{fuchsiapink} \footnotesize{Sheep 0.981} }
   \end{tabular}
    \caption[Successful detection on CASPA paintings with \MaxOfMaxS{} model]{Successful examples of animal detection using \MaxOfMaxS{} on CASPA paintings test set (there is no "person" class in the training set). We only show boxes whose scores are over 0.75, except for the elephant image. Figure must be seen in color.}
    \label{fig:CASPApaintingsSuccessfulDetection}
\end{figure}

\begin{figure}
\centering
\setlength\tabcolsep{1pt}
\renewcommand{\arraystretch}{0.5}
\begin{tabular}{cc}
    \includegraphics[height=\heightimageCASPA]{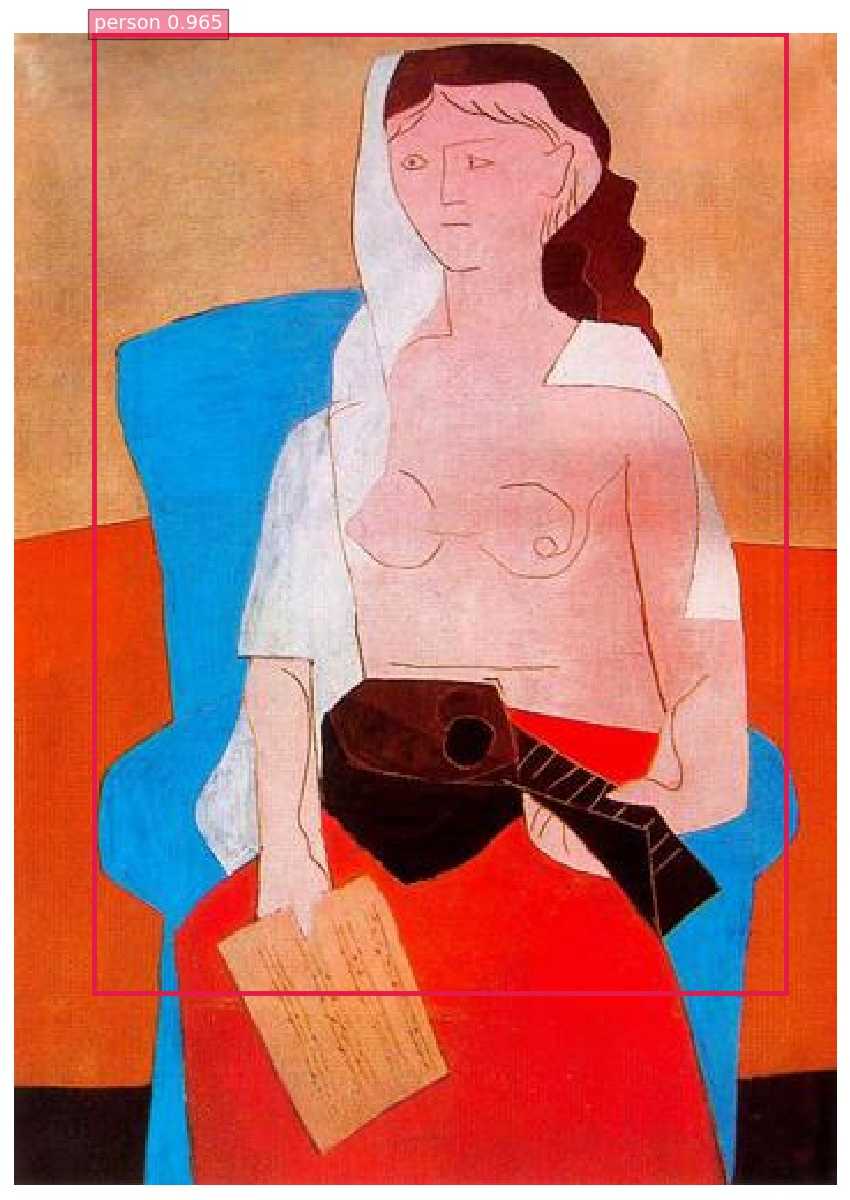} & \includegraphics[height=\heightimageCASPA]{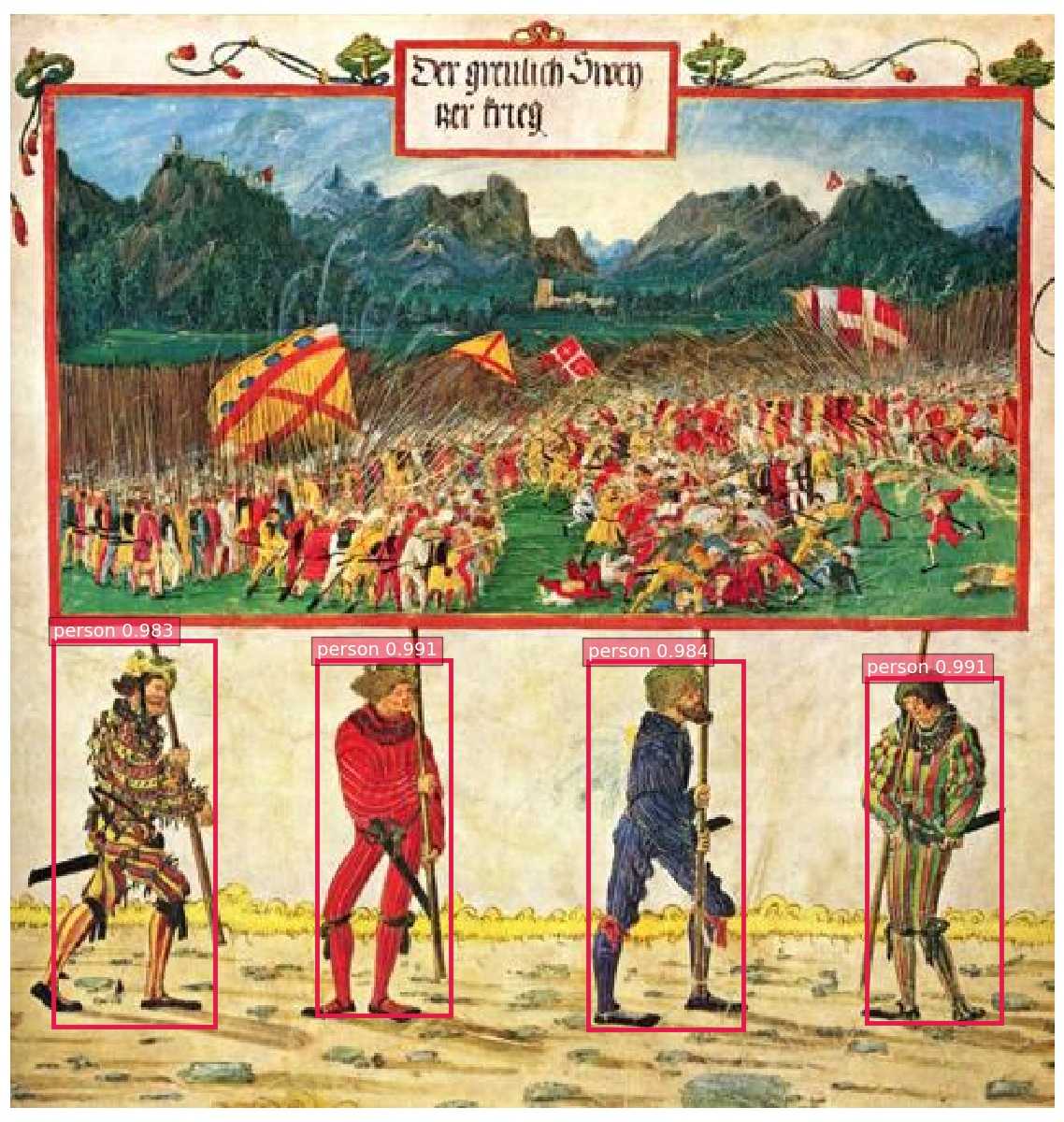}
    \\
      {\color{red} \footnotesize{Person 0.965}} & {\color{red} \footnotesize{Person 0.983 0.991 0.984 0.9991}} \\
     \includegraphics[height=\heightimageCASPAzeroSept]{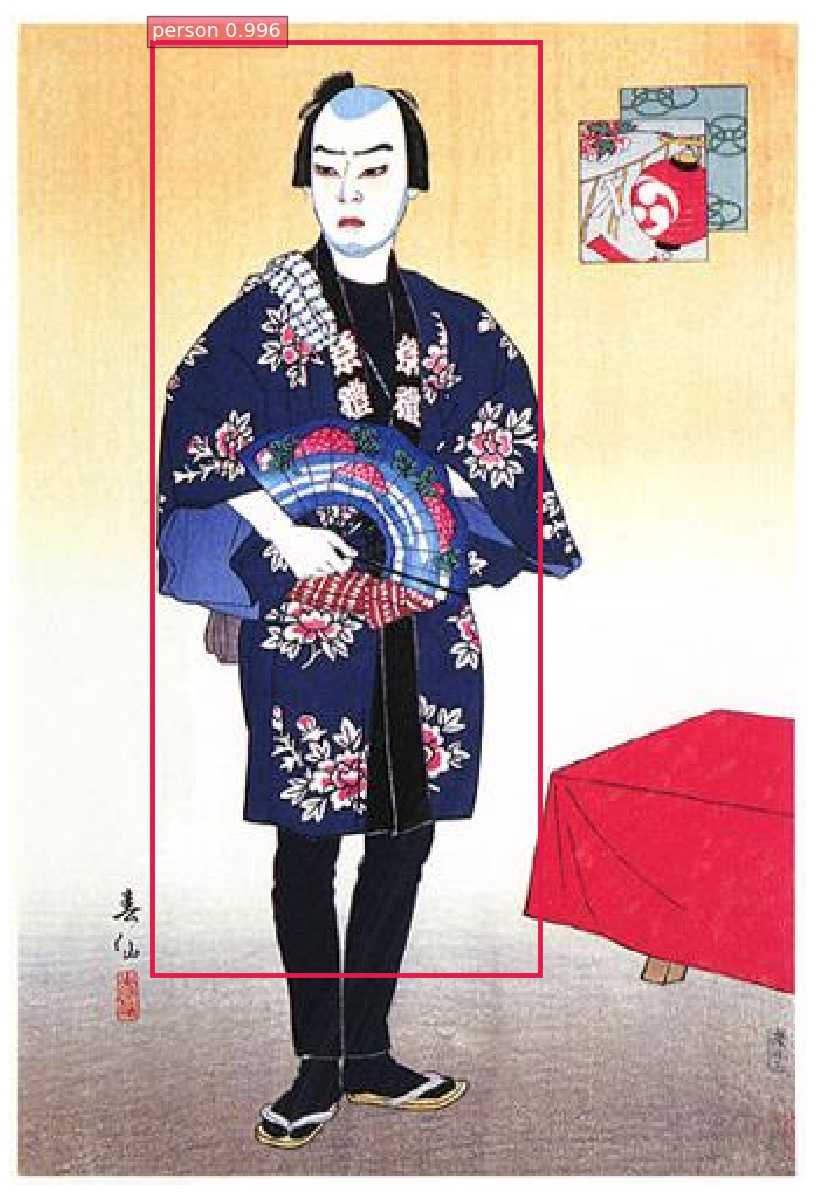} &
      \includegraphics[height=\heightimageCASPAzeroSept]{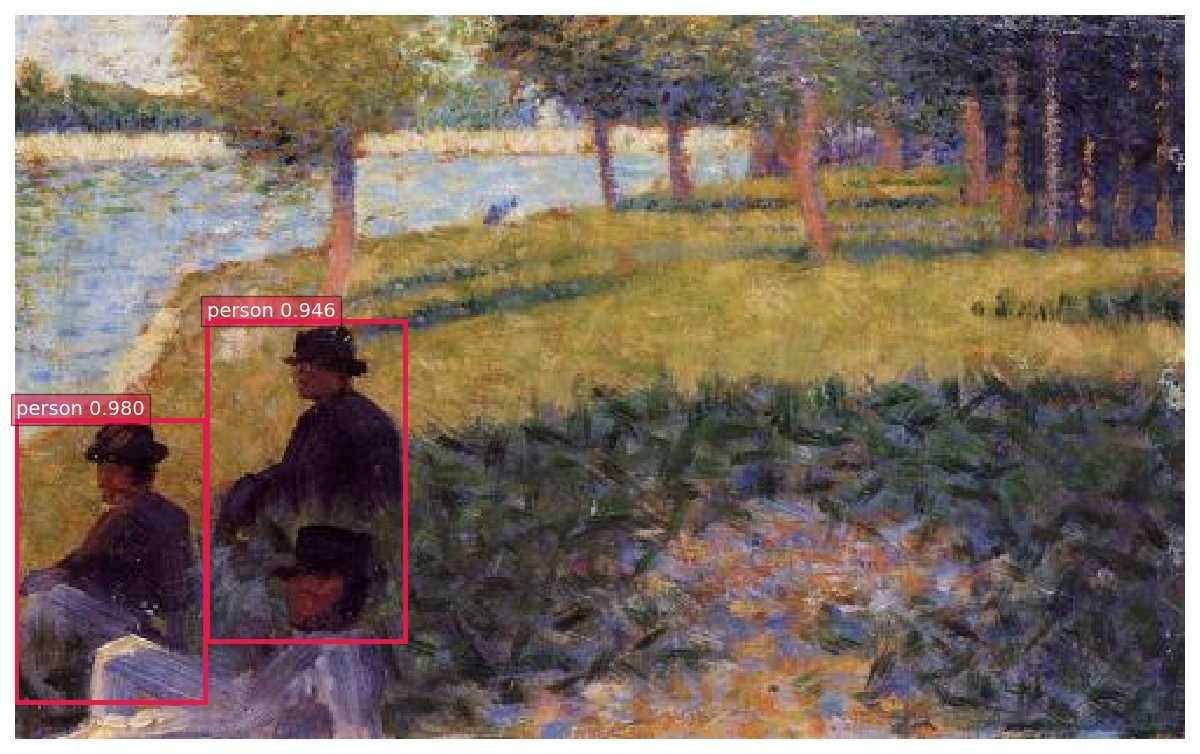} \\
          {\color{red} \footnotesize{Person 0.996}} &  {\color{red} \footnotesize{Person 0.980 0.946}}  \\
   \end{tabular}
    \caption{Successful examples using our \MaxOfMaxS{} detection scheme on PeopleArt test set. One can observe the strong stylistic differences between the images. We only show boxes whose scores are over 0.75. Figure must be seen in color.}
    \label{fig:PeopleArtSuccessfulDetection}
\end{figure}

\begin{figure}
\centering
\setlength\tabcolsep{1pt}
\renewcommand{\arraystretch}{0.5}
\begin{tabular}{ccc}
 \includegraphics[height=\heightimageCASPA]{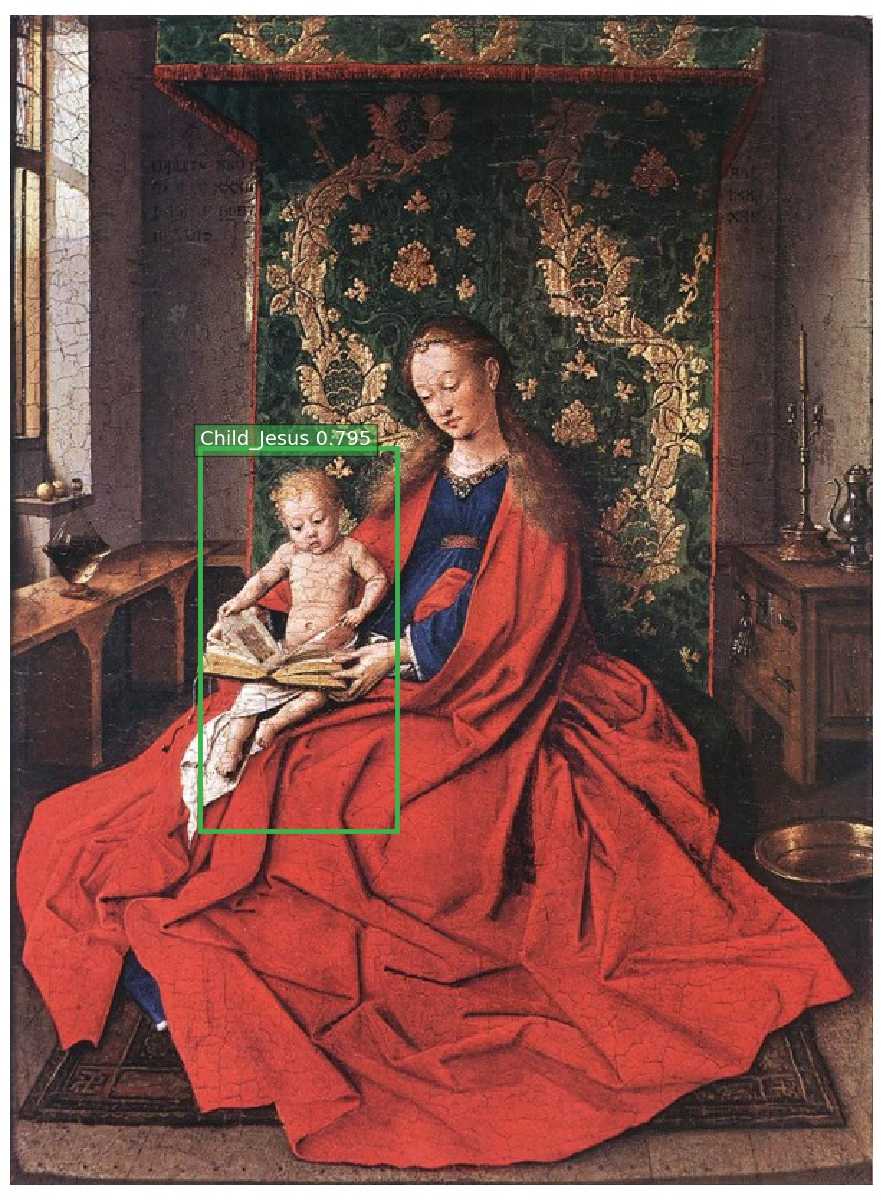} &
     \includegraphics[height=\heightimageCASPA]{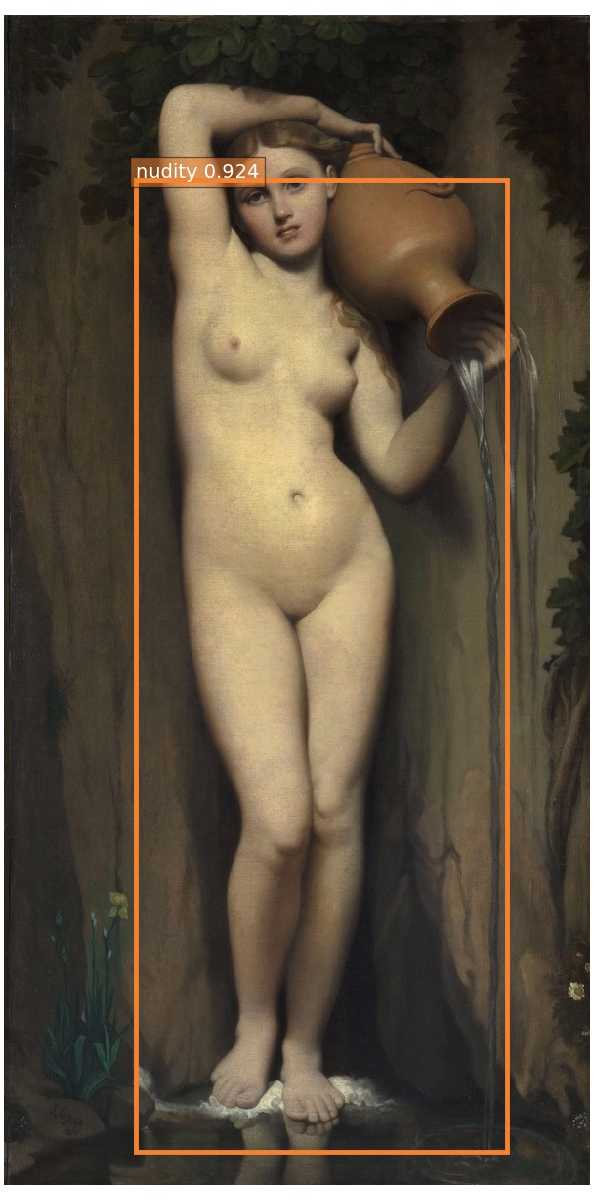} & \includegraphics[height=\heightimageCASPA]{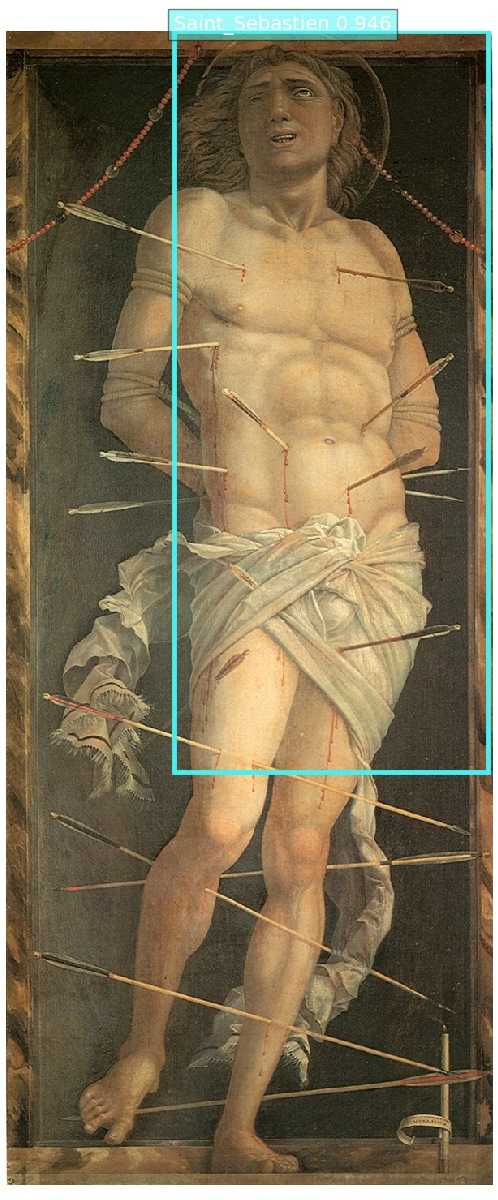} \\
     {\color{darkpastelgreen} \footnotesize{Jesus Child 0.795} } &  {\color{carrotorange} \footnotesize{Nudity 0.924}} & {\color{cyan} \footnotesize{Saint Sebastian 0.946} }   \\
     \multicolumn{3}{c}{\includegraphics[height=\heightimageCASPA]{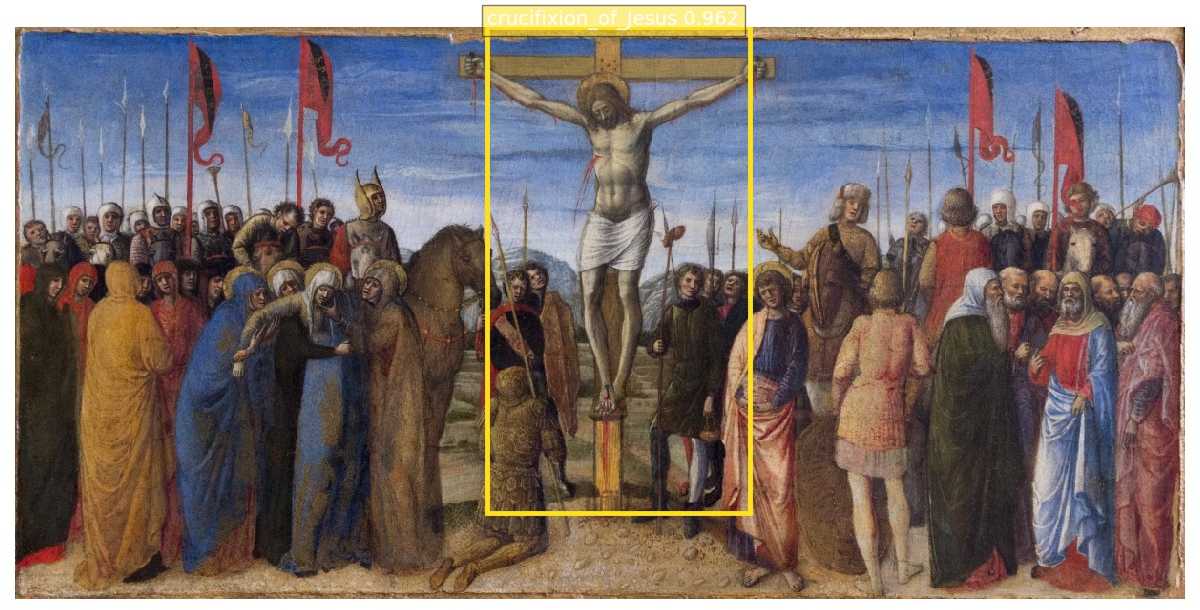}}\\
    \multicolumn{3}{c}{{\color{electricyellow} \footnotesize{Crucifixion  0.962} }} \\
   \end{tabular}
    \caption{Successful examples of detection of iconographic characters using our \MaxOfMaxS{} detection scheme on IconArt test set. We only show boxes whose scores are over 0.75. Figure must be seen in color.}
    \label{fig:IconArtv1SuccessfulDetection}
\end{figure}

{\bf Successful detections:}
We show successful results on various datasets. 
In \cref{fig:WatercolorSuccessfulDetection,fig:CASPApaintingsSuccessfulDetection} we show various examples of the visual categories we are able to detect, respectively on Watercolor2k and CASPApainting datasets. On Figure \ref{fig:PeopleArtSuccessfulDetection}, we can see the large stylistic diversity that the model is able to detect for a same class, namely person, on the PeopleArt dataset.
On Figure \ref{fig:IconArtv1SuccessfulDetection}, one can see some detections on the challenging IconArt dataset.

\begin{figure}
\centering
\setlength\tabcolsep{1pt}
\renewcommand{\arraystretch}{0.5}
\begin{tabular}{cc}
  \includegraphics[height=\heightimageCASPA]{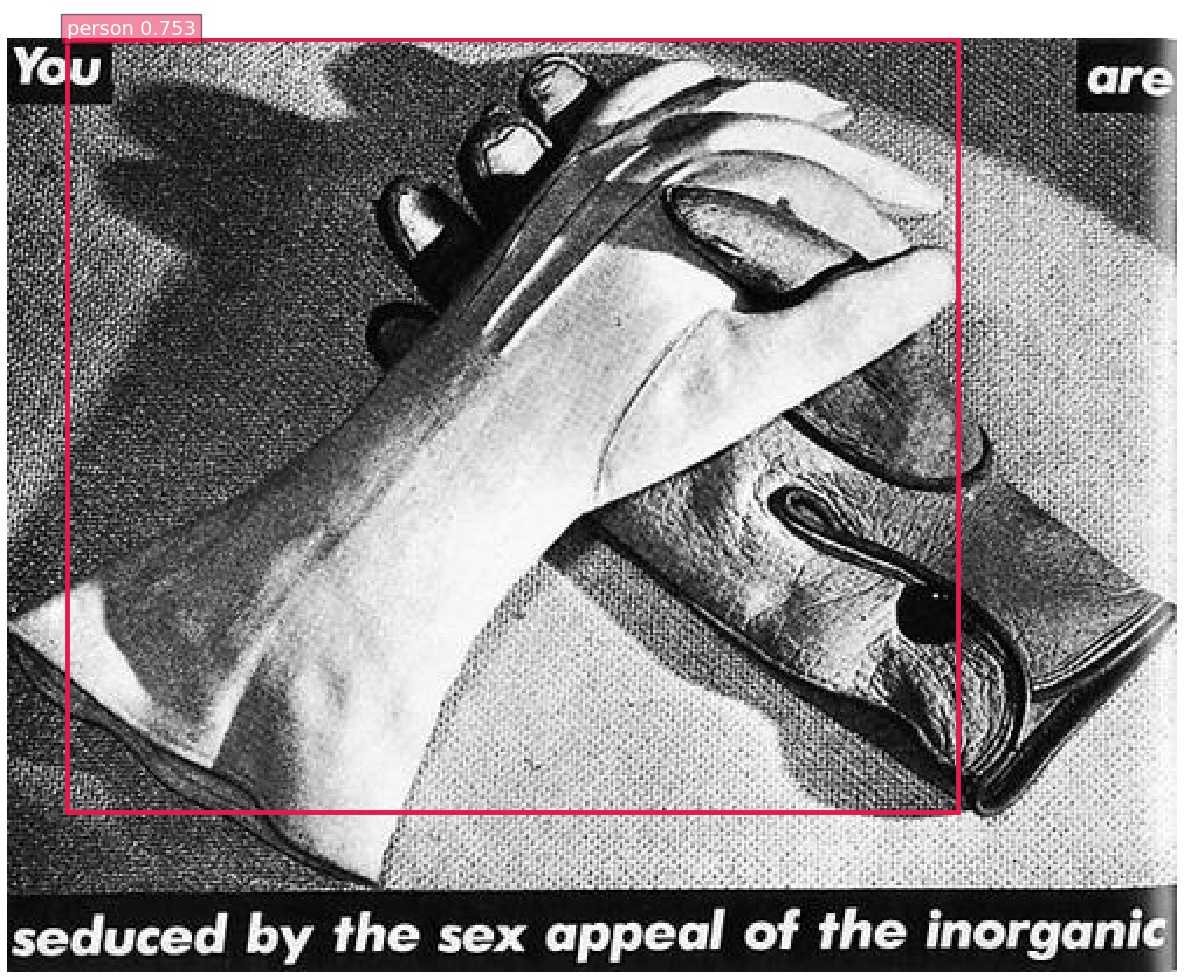} &
       \includegraphics[height=\heightimageCASPA]{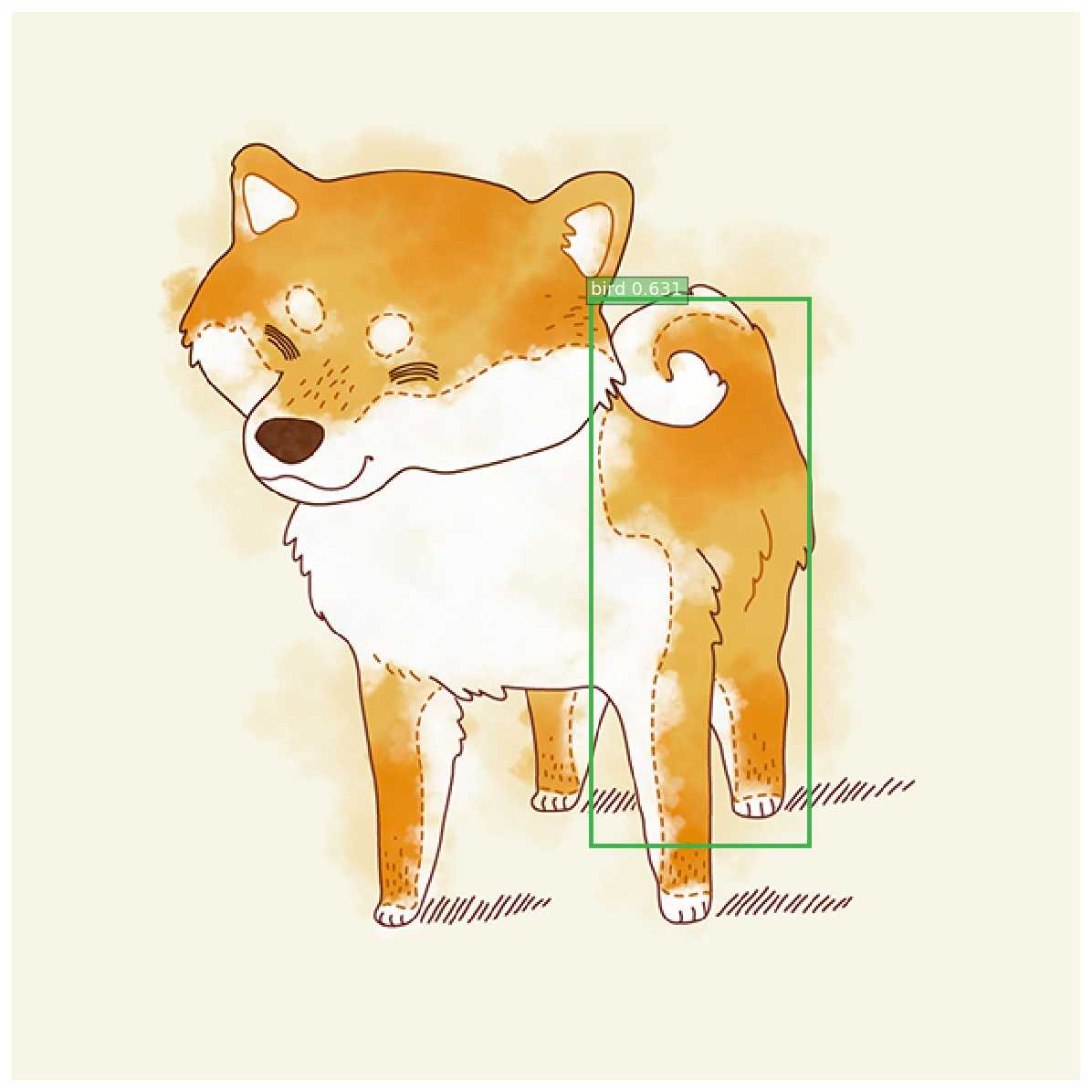}  \\
 {\color{red} \footnotesize{Person 0.753} } & {\color{darkpastelgreen} \footnotesize{Bird 0.631} }  \\
\end{tabular}
    \caption{Failure examples using our our \MaxOfMaxS{} detection scheme on different datasets. We only show boxes whose scores are over 0.75. The most discriminative boxes correspond to parts of the whole objects. On the first image, the gloves are detected instead of a person. On the second one, the  back legs and tail are detected as a dog. On the last one, the legs are detected as nudity. Figure must be seen in color.}
    \label{fig:DetectionOfDiscriminativeElements}
\end{figure}

\begin{figure}[h]
\centering
\setlength\tabcolsep{1pt}
\renewcommand{\arraystretch}{0.5}
\begin{tabular}{ cc  }
\multicolumn{2}{c}{ \MIL{}}  \\
  \includegraphics[height=\heightimageCASPA]{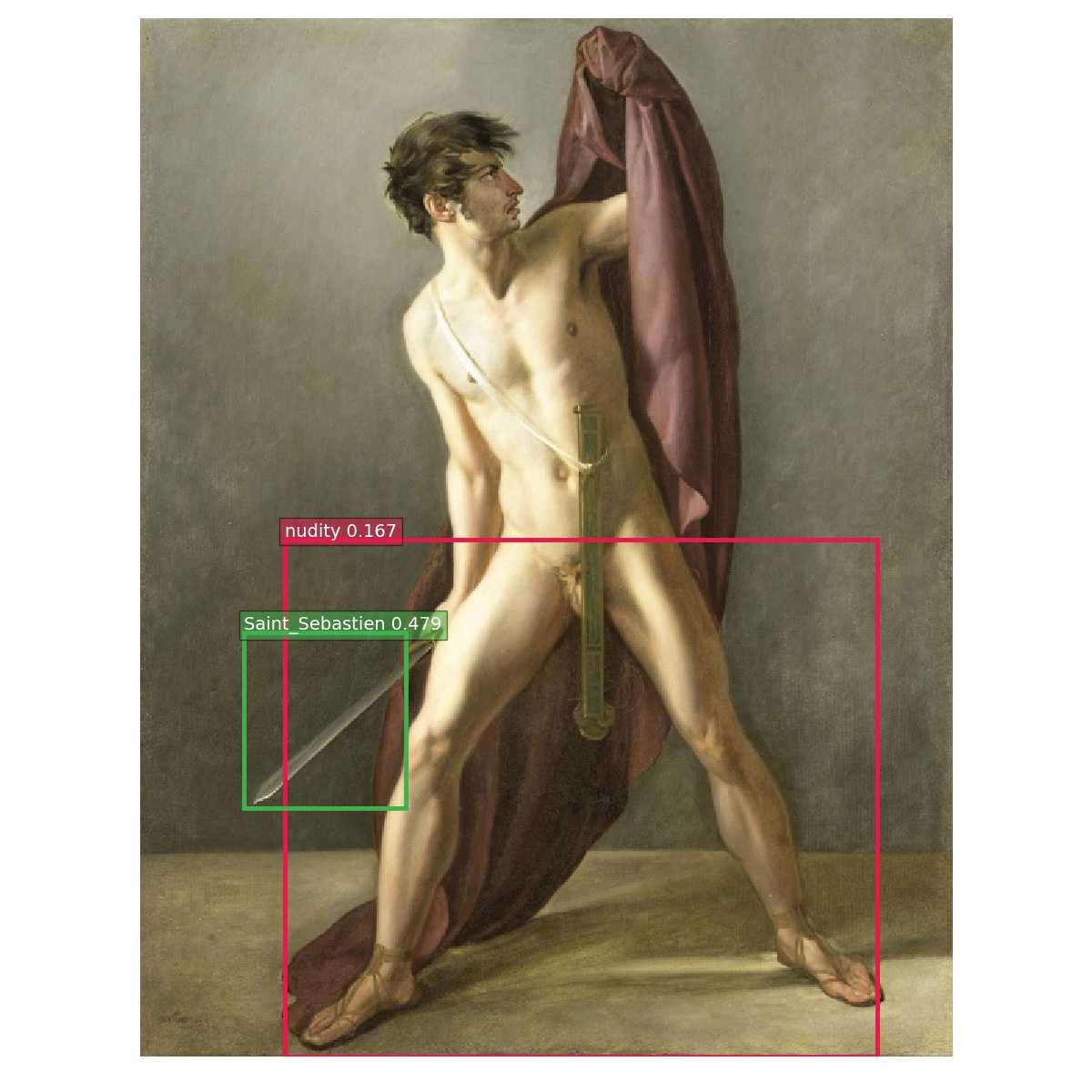}  &
     \includegraphics[height=\heightimageCASPA]{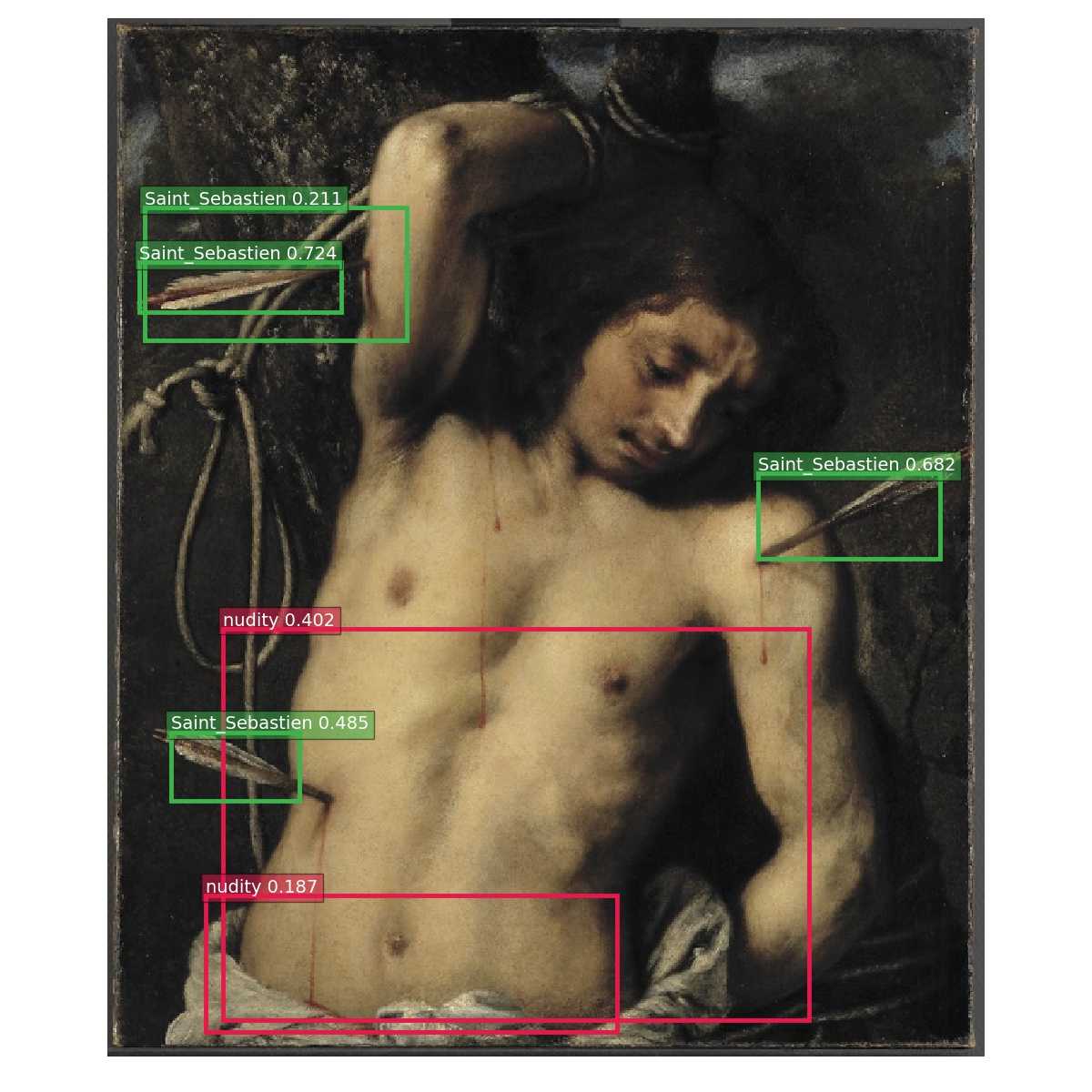} \\
      {\color{darkpastelgreen} \footnotesize{St Sebastian 0.479} }  &    {\color{darkpastelgreen} \footnotesize{St Sebastian 0.221 0.724 0.682 0.485} }  \\
      {\color{red} \footnotesize{Nudity 0.167} } & {\color{red} \footnotesize{Nudity 0.402 0.187} }  \\
     \multicolumn{2}{c}{ \MILS{} with score} \\
     \includegraphics[height=\heightimageCASPA]{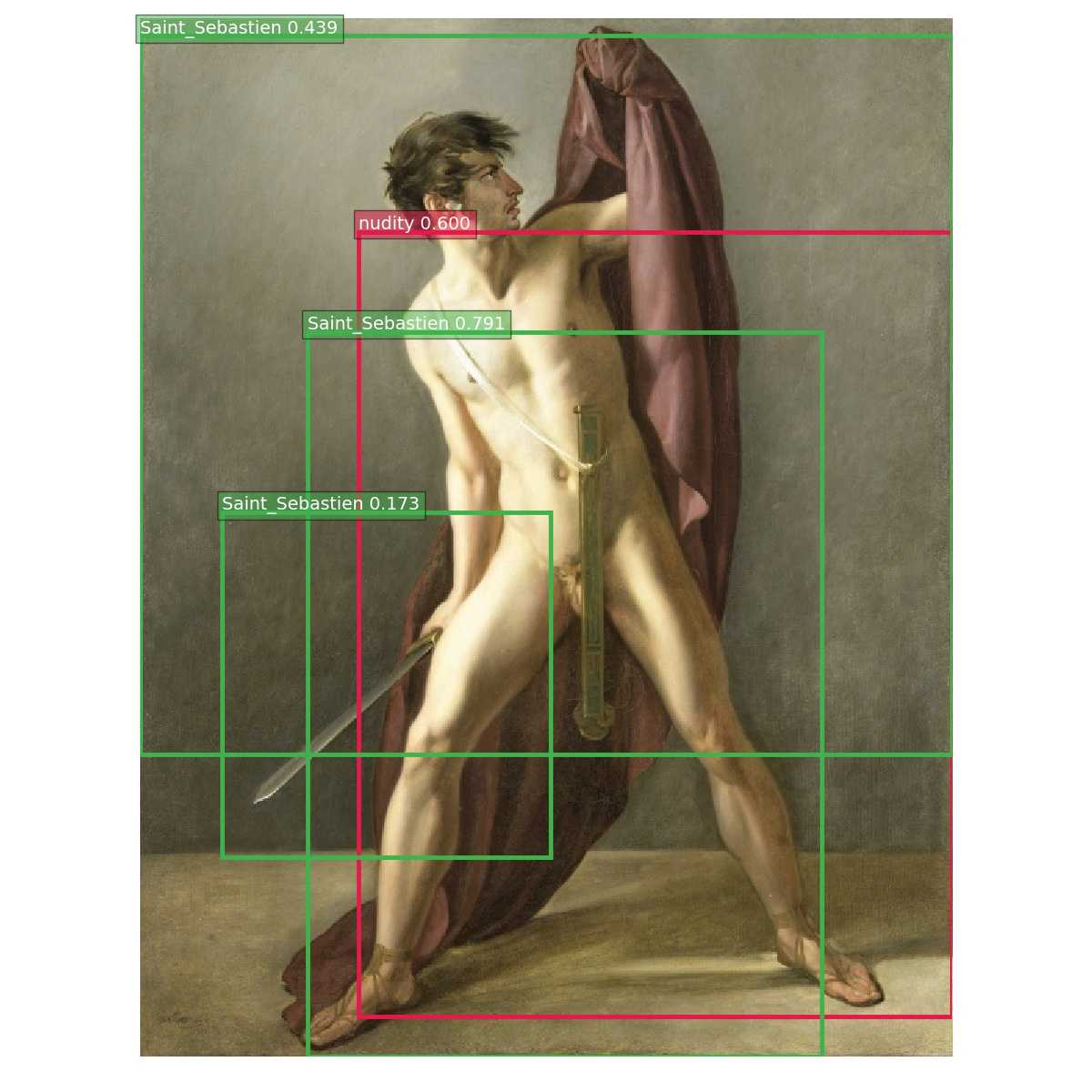}   &
 \includegraphics[height=\heightimageCASPA]{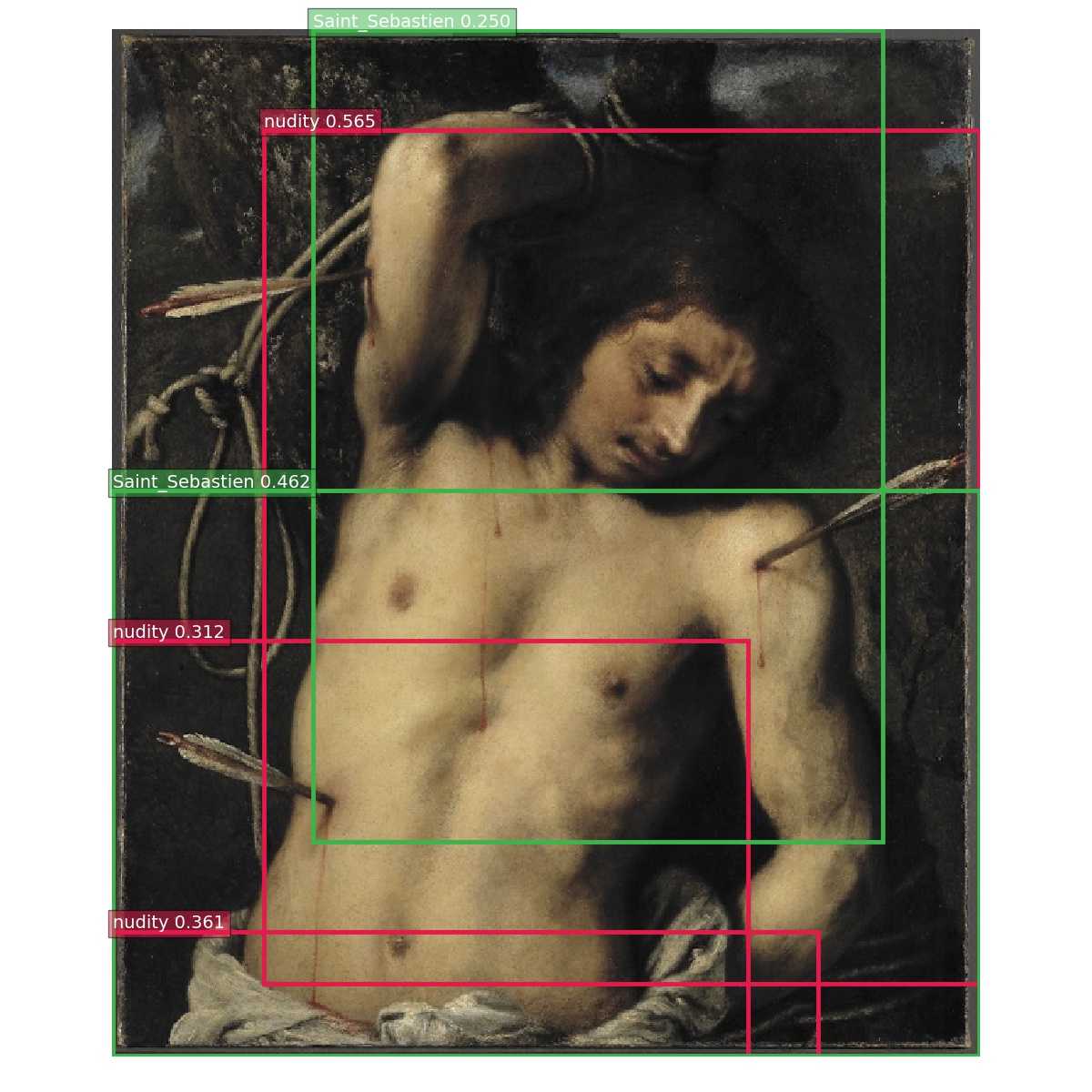} \\
  {\color{darkpastelgreen} \footnotesize{St Sebastian 0.439 0.791 0.173} }  &    {\color{darkpastelgreen} \footnotesize{St Sebastian 0.25 0.462} }  \\
  {\color{red} \footnotesize{Nudity 0.6} }  &  {\color{red} \footnotesize{Nudity 0.565 0.312 0.361} } \\
  \end{tabular}
    \caption{An example of wrongly detected object at test time, when using \MILS{} without or with the objectness score. In the first case, arrows or spike are detected instead of Saint Sebastian. Figure must be seen in color.}
    \label{fig:Bowes_without_and_withScore}
\end{figure}

\begin{figure}
\centering
\setlength\tabcolsep{1pt}
\renewcommand{\arraystretch}{0.5}
\begin{tabular}{cc}
          \includegraphics[height=\heightimageCASPA]{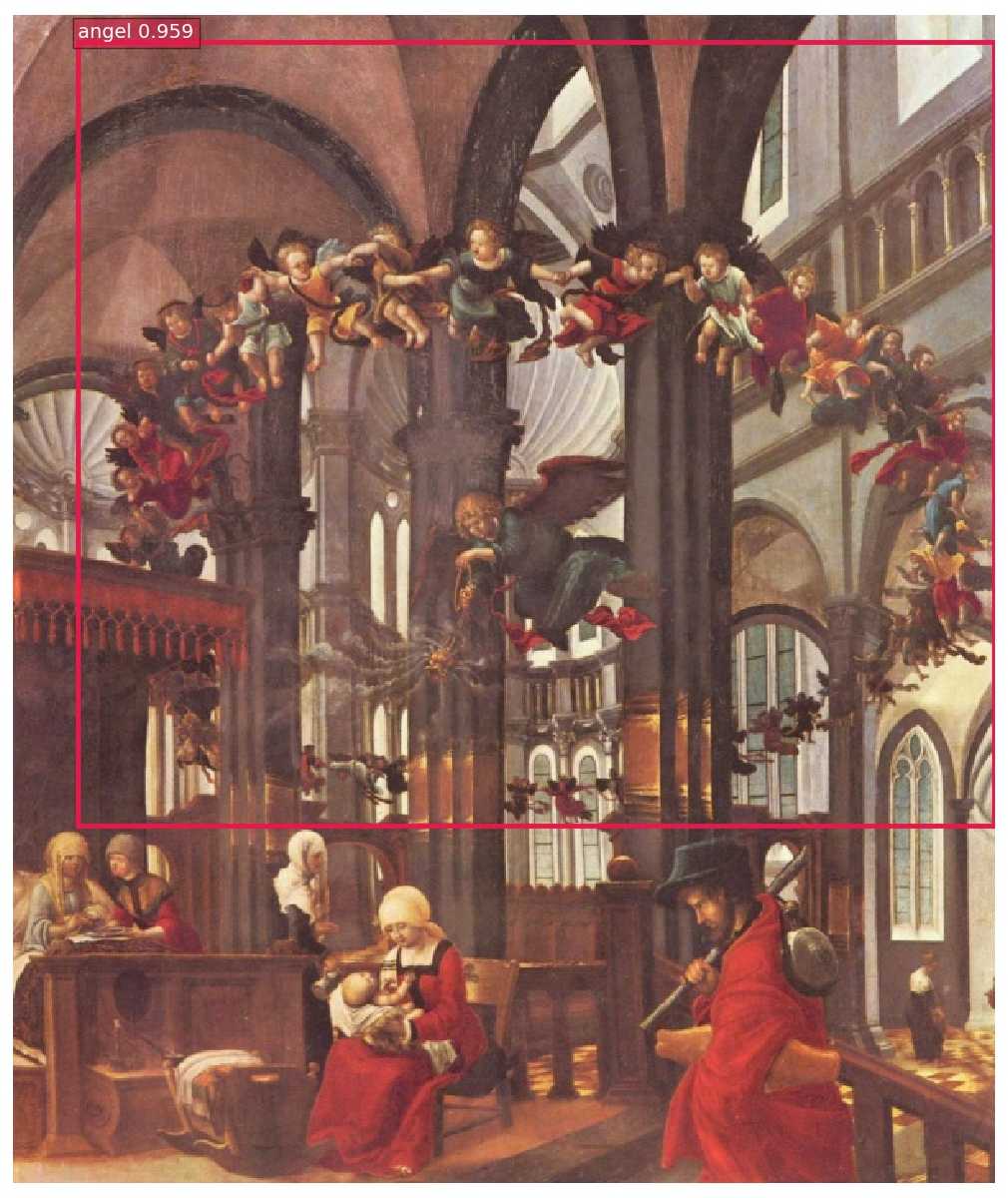} &
     \includegraphics[height=\heightimageCASPA]{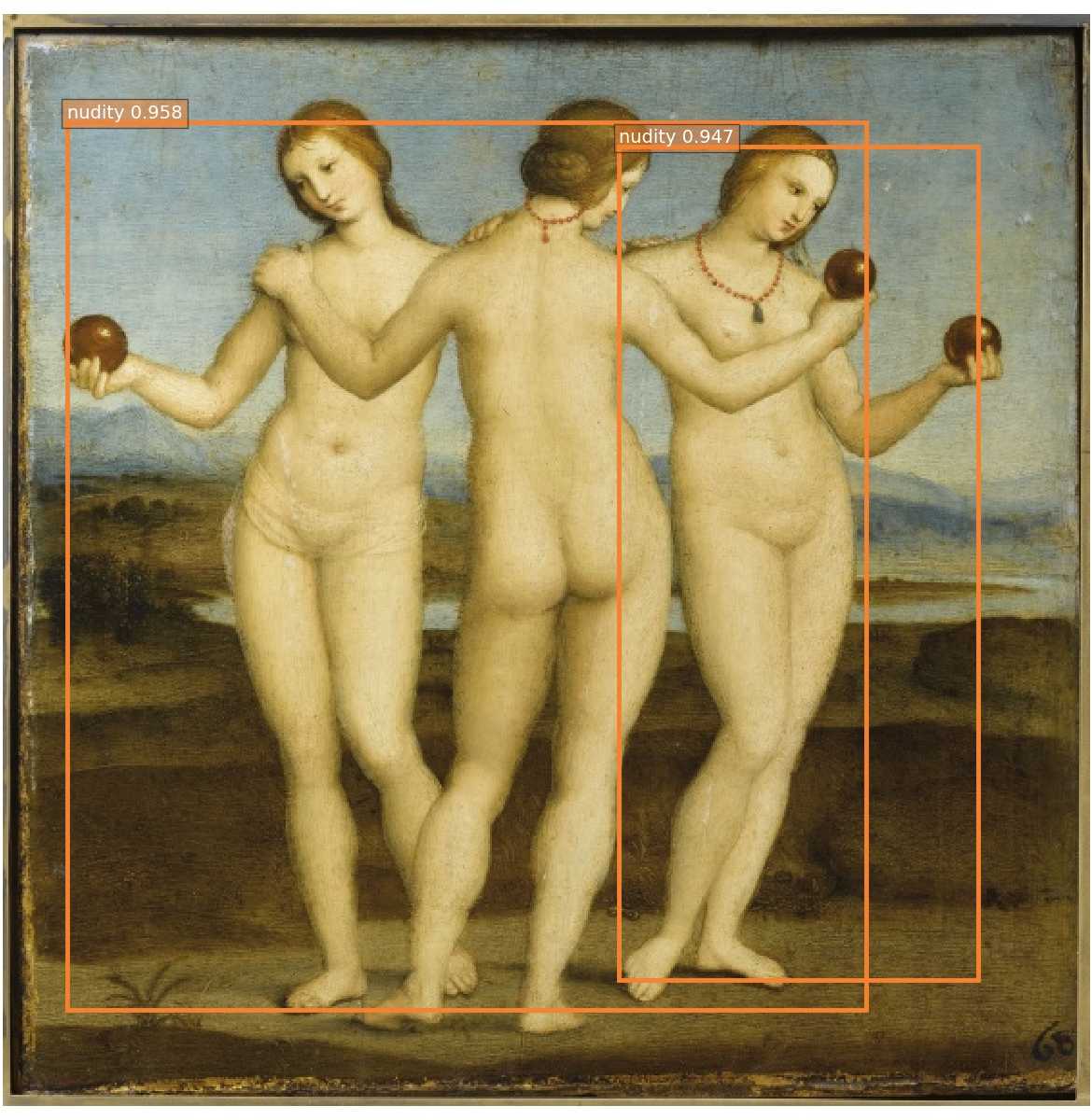}\\
     {\color{red} \footnotesize{Angel 0.959}} & {\color{carrotorange} \footnotesize{Nudity 0.958 0.947}} \\
      \multicolumn{2}{c}{\includegraphics[height=\heightimageCASPAzeroSept]{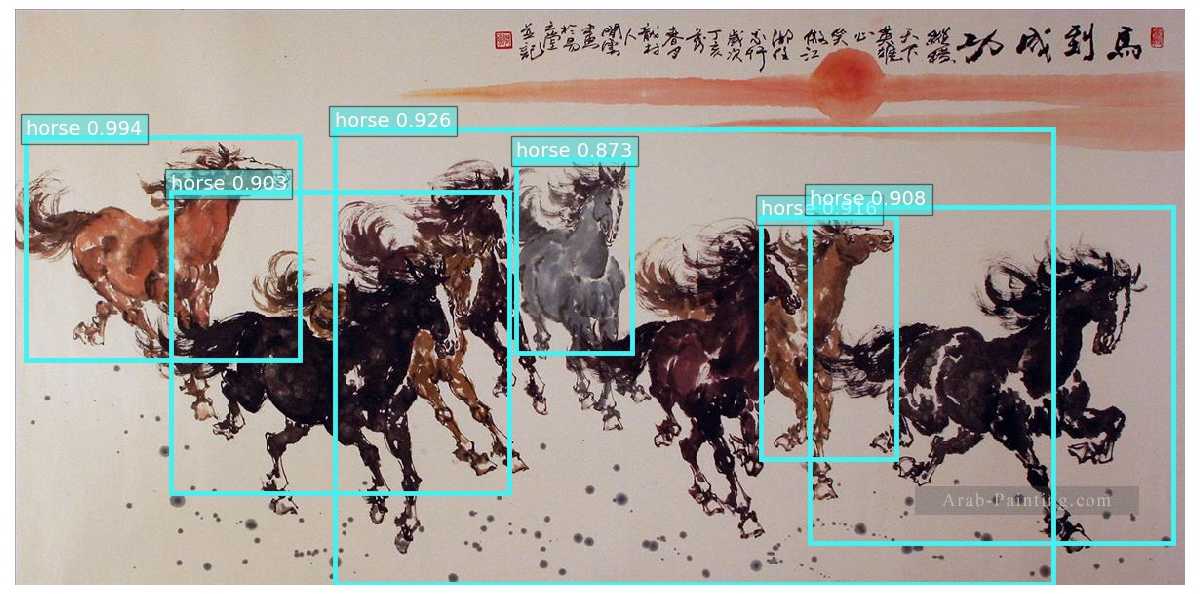}} \\
          \multicolumn{2}{c}{{\color{cyan} \footnotesize{Horse 0.994 0.903 0.926 0.873 0.916 0.908} }} \\
\end{tabular}
    \caption{Failure examples using our our \MaxOfMaxS{} detection scheme on different datasets. We only show boxes whose scores are over 0.75. Whole groups are detected instead of the instances. Figure must be seen in color.}
    \label{fig:detectionWholeGroup}
\end{figure}

\begin{figure}
\centering
\setlength\tabcolsep{1pt}
\renewcommand{\arraystretch}{0.5}
\begin{tabular}{cc}
\includegraphics[height=\heightimageCASPA]{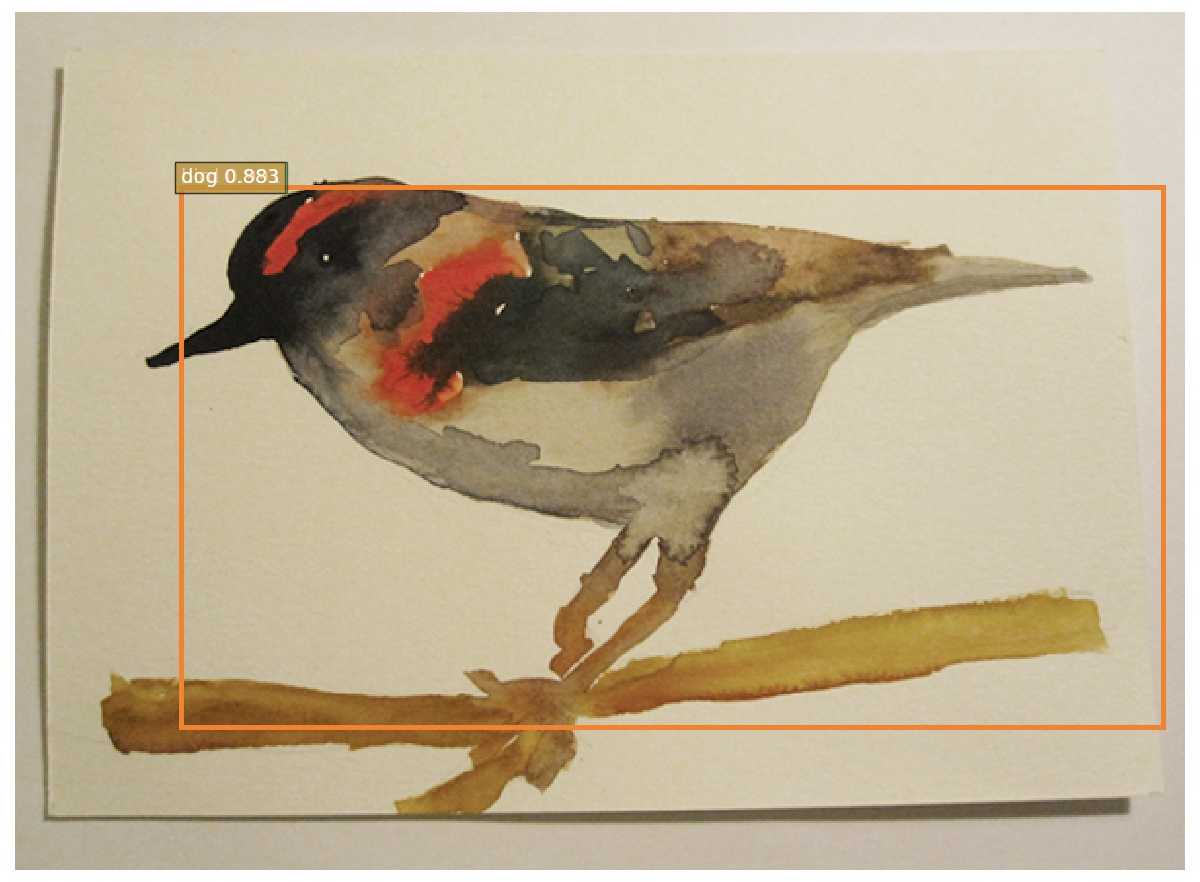} & 
     \includegraphics[height=\heightimageCASPA]{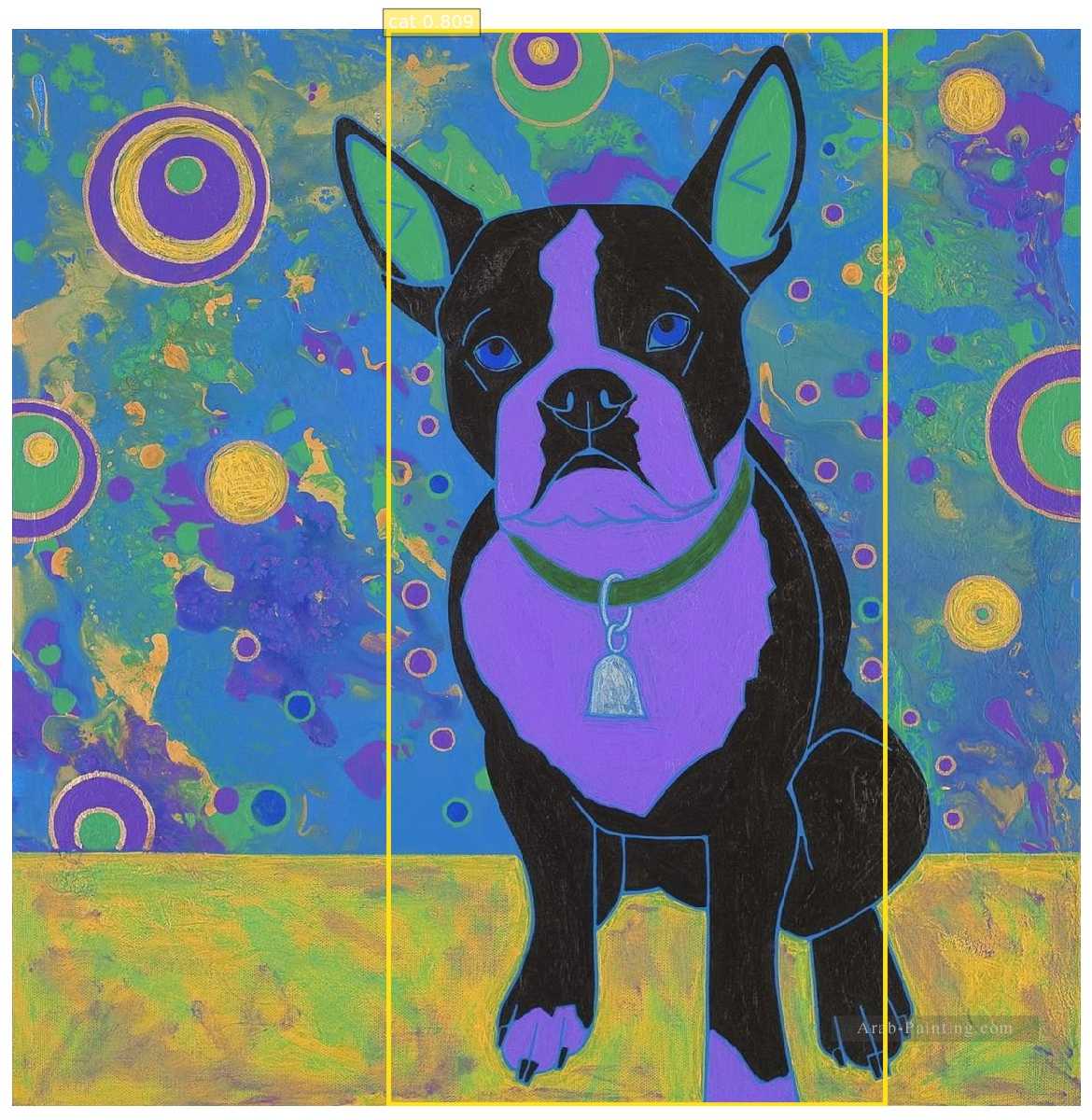}\\
     {\color{carrotorange} \footnotesize{Dog 0.883}} &  {\color{electricyellow} \footnotesize{Cat 0.809 }}\\
\end{tabular}
    \caption{Failure examples using our our \MaxOfMaxS{} detection scheme on different datasets. We only show boxes whose scores are over 0.75. Mis-classified boxes: on the first image the bird is classified as a dog and on the second one the dog is detected as a cat. Figure must be seen in color.}
    \label{fig:MisClassified}
\end{figure}

\begin{figure}
\centering
\setlength\tabcolsep{1pt}
\renewcommand{\arraystretch}{0.5}
\begin{tabular}{cc}
\includegraphics[height=\heightimageCASPA]{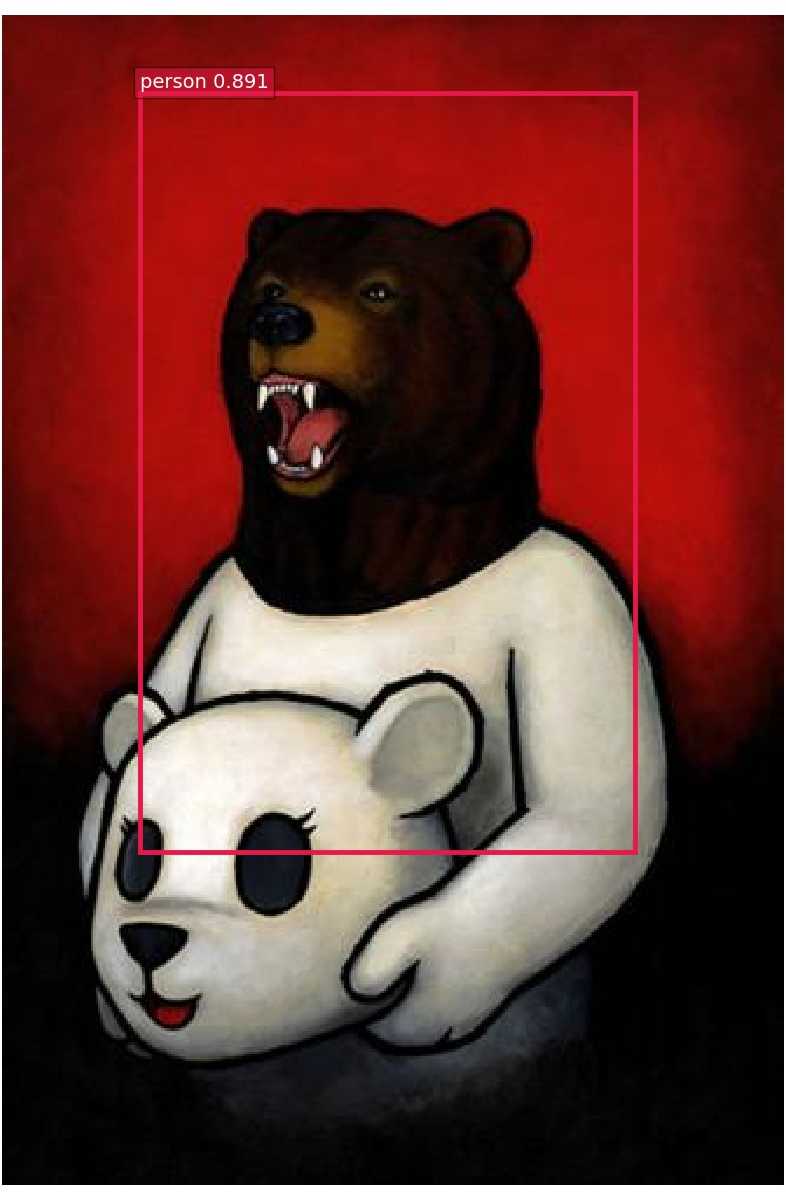} &
             \includegraphics[height=\heightimageCASPA]{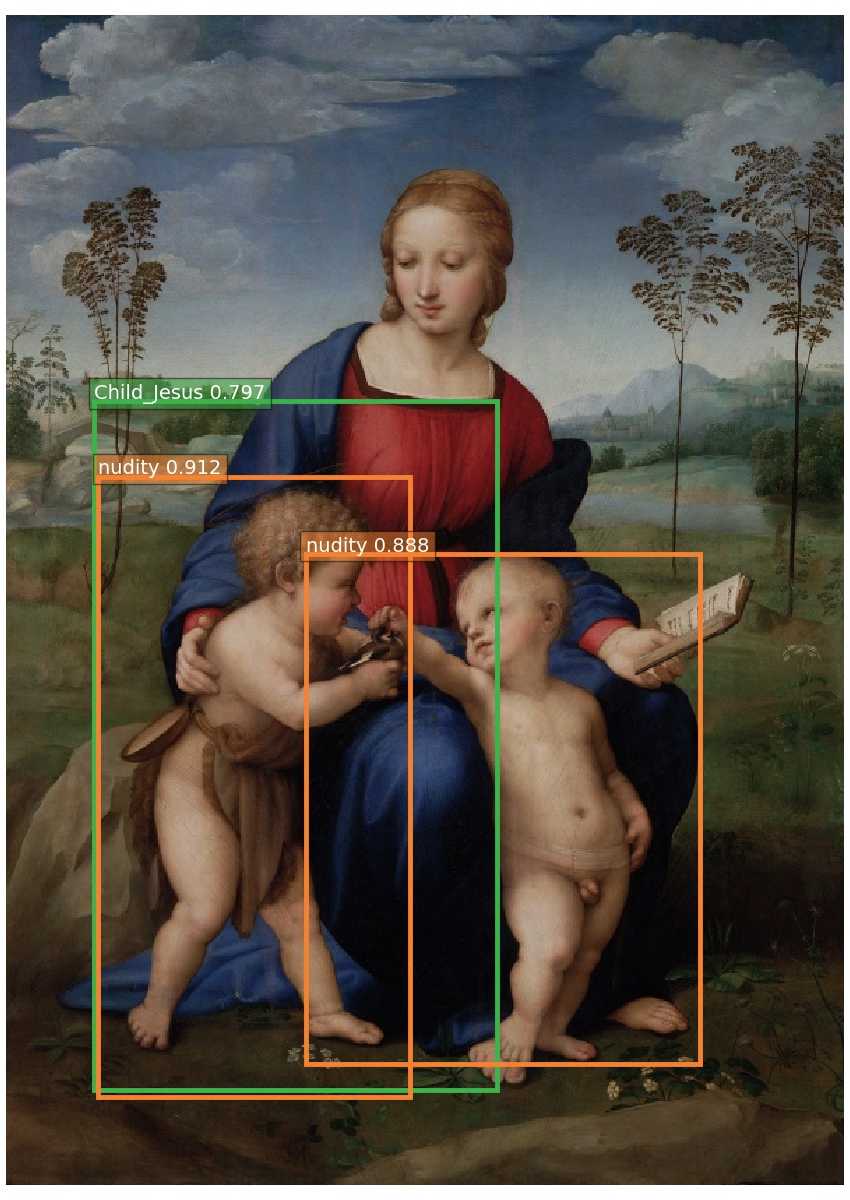}\\
 {\color{red} \footnotesize{Person 0.891}} &    {\color{darkpastelgreen} \footnotesize{Jesus Child 0.797}} {\color{carrotorange} \footnotesize{Nudity  0.912 0.888}}  \\
 \multicolumn{2}{c}{\includegraphics[height=\heightimageCASPA]{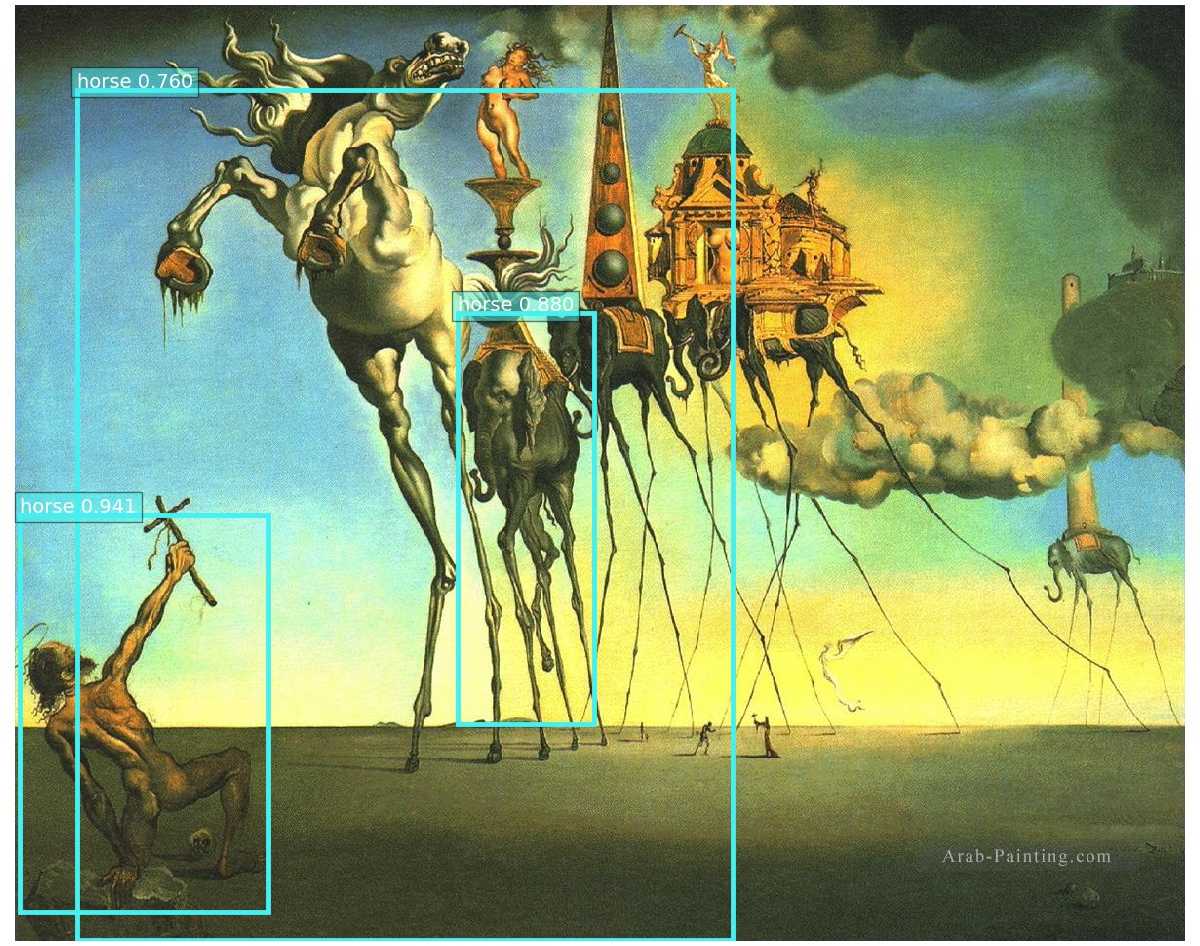}}\\ 
  \multicolumn{2}{c}{{\color{cyan} Horse 0.981 0.76 0.98}}\\
\end{tabular}
    \caption{Failure examples using our our \MaxOfMaxS{} detection scheme on different datasets. We only show boxes whose scores are over 0.75. Those are confusing images. In the first one a bear in an human posture is detected as a person. In the middle, the horse, the man and other animals are deformed. The last one is a confusing case between Saint John the Baptist and Jesus children who are visually similar. Figure must be seen in color.}
    \label{fig:ConfusingImages}
\end{figure}

{\bf Failures examples:} We can categorize the failures cases into five main categories: 
\begin{enumerate}
\item Discriminative elements are detected instead of the whole object: the hand for instance in Figure \ref{fig:DetectionOfDiscriminativeElements} for the \MaxOfMax{} model or the arrows instead of Saint Sebastian in Figure \ref{fig:Bowes_without_and_withScore}) for the \MILS{} model without score.
\item  Detection of a whole group instead of individual instances (Figure \ref{fig:detectionWholeGroup}).
\item Misclassification of correct bounding box, as in Figure \ref{fig:MisClassified}.
\item Confusing images (Figure \ref{fig:ConfusingImages}, relatively advanced knowledge in art history is needed to know that the child on the left is Saint John the Baptist).
\end{enumerate}

\FloatBarrier
\section{Conclusion}
\label{sec:ccl}
  In this paper, we confirm that transfer learning of pretrained CNN can provide good model to automatically analyze non photo-realistic images databases. %
This was previously shown for classification and fully supervised detection tasks, and was here investigated in the case of weakly supervised object detection. We proposed a simple and quick model to solve the multiple instance problem we are facing. 
  In future works, we plan to add some constraint in the polyhedral case to force the hyperplanes to be as distinct as possible to get better boundaries, to develop on piece-wise linear model. It might be beneficial to take in more than one instance per bag to learn better detector and catch multi-modal visual category. 
  A more extensive investigation of the different possible features extractor and boxes proposals algorithms could show the flexibility of our model.
  Another exciting direction is to investigate the potential of weakly supervised learning on large databases with only image-level annotations. For instance, this framework could be used to develop versatile search engine for diverse modalities of images, avoiding the time consuming annotation task.
 Moreover, we plan to supervise the training of weak detector with a fully-trained classifier in order to remove some obvious mis-classified box candidate as it can be done in classical WSOD method \citep{wan_minentropy_2018}. This could help to provide better detection performances.

\vspace{1cm}
{\noindent \bf Acknowledgements.}
 This work is supported by the "IDI 2017" project funded by the IDEX Paris-Saclay, ANR-11-IDEX-0003-02 and by T\'el\'ecom Paris.

{\small
\bibliography{Zotero}
}

\end{document}